\documentclass[journal]{IEEEtran}
\usepackage{times}

\usepackage[numbers]{natbib}
\usepackage{multicol}
\usepackage[bookmarks=true]{hyperref}
\usepackage{mathtools}
\usepackage{wrapfig}

\usepackage{bm}
\usepackage{amsmath, amsfonts, amssymb}
\DeclareMathOperator*{\argmin}{argmin}
\DeclareMathOperator*{\argmax}{argmax} 

\newcommand{\bom}{\mbox{\boldmath\(\omega\)}}

\newcommand{\bSi}{\mbox{\boldmath\(\Sigma\)}}
\newcommand{\bmu}{\mbox{\boldmath\(\mu\)}}
\newcommand{\bL}{\mbox{\boldmath\(L\)}}

\newcommand{\Loss}{\mathcal{L}}
\newcommand{\logexp}{\mathrm{logexp}}
\renewcommand{\vec}{\bm}
\def \u{{\vec{u}}}
\def \x{{\vec{x}}}
\def \X{{\vec{X}}}
\def \U{{\vec{U}}}

\def \g{\vec{g}}
\def \O{\mathcal{O}}
\def \E{\mathbb{E}}
\def \p{{\vec{p}}}
\def \mean{\vec{\mu}}
\def \pg{p_{\g}} 
\usepackage{booktabs}

\usepackage{amsthm}
\newtheorem{claim}{Claim}
\newtheorem{remark}{Remark}

\makeatletter
\newcommand{\subalign}[1]{%
  \vcenter{%
    \Let@ \restore@math@cr \default@tag
    \baselineskip\fontdimen10 \scriptfont\tw@
    \advance\baselineskip\fontdimen12 \scriptfont\tw@
    \lineskip\thr@@\fontdimen8 \scriptfont\thr@@
    \lineskiplimit\lineskip
    \ialign{\hfil$\m@th\scriptstyle##$&$\m@th\scriptstyle{}##$\hfil\crcr
      #1\crcr
    }%
  }%
  }

  \renewcommand\paragraph{\@startsection{paragraph}{4}{\z@}%
  {1.25ex \@plus1ex \@minus.2ex}%
  {-0em}%
  {\normalfont\normalsize\bfseries}}
\makeatother

\newcommand{\kl}[2]{\text{D}_{\text{KL}}\Bigl(#1 \hspace{4pt}\Bigl|\Bigr|\hspace{4pt}#2\Bigr)}
\newcommand{\inlinekl}[2]{\text{D}_{\text{KL}}\left(#1\hspace{4pt}\parallel\hspace{4pt}#2\right)}
\newcommand{\inlinece}[2]{\text{D}_{\text{CE}}\left(#1\hspace{4pt}\parallel\hspace{4pt}#2\right)}
\newcommand{\approxce}[2]{\text{D}_{\text{CE-app}}\Bigl(#1 \hspace{4pt}\Bigl|\Bigr|\hspace{4pt}#2\Bigr)}

\newcommand{\vmin}[1]{#1_{\text{min}}}
\newcommand{\vmax}[1]{#1_{\text{max}}}
\newcommand{\allt}{\quad\forall t \in [0, T]}
\newcommand{\alltm}{\quad\forall t \in [0, T-1]}

\usepackage{graphicx}
\usepackage[skip=10pt,font=small,labelfont=bf]{caption}
\usepackage{subcaption}
\usepackage{xargs}
\usepackage[pdftex,dvipsnames]{xcolor}
\usepackage[colorinlistoftodos,prependcaption,textsize=footnotesize]{todonotes}
\definecolor{international_orange}{RGB}{240, 74, 0}
\def \adamColor {MidnightBlue}
\def \tuckerColor {international_orange}
\newcommandx{\adam}[1]{\todo[linecolor=\adamColor,backgroundcolor=\adamColor!25,bordercolor=\adamColor,inline]{\textbf{AC:} #1}}
\newcommandx{\tucker}[1]{\todo[linecolor=\tuckerColor,backgroundcolor=\tuckerColor!25,bordercolor=\tuckerColor,inline]{\textbf{TH:} #1}}

\begin{document}

\title{\LARGE \bf Planning under Uncertainty to Goal Distributions}

\author{Adam Conkey$^{1}$ and Tucker Hermans$^{1,2}$ %
  \thanks{$^1$~University of Utah Robotics Center and School of Computing,
  University of Utah, Salt Lake City, UT, USA.}
  \thanks{$^2$ NVIDIA Corporation, USA}
  \thanks{Email: \texttt{adam.conkey@utah.edu, thermans@cs.utah.edu}}}

\maketitle

\begin{abstract}

  Goals for planning problems are typically conceived of as subsets of the state
  space. However, for many practical planning problems in robotics, we expect
  the robot to predict goals, e.g. from noisy sensors or by generalizing learned
  models to novel contexts. In these cases, sets with uncertainty naturally
  extend to probability distributions. While a few works have used probability
  distributions as goals for planning, surprisingly no systematic treatment of
  planning to goal distributions exists in the literature. This article serves
  to fill that gap. We argue that goal distributions are a more appropriate goal
  representation than deterministic sets for many robotics applications. We
  present a novel approach to planning under uncertainty to goal distributions,
  which we use to highlight several advantages of the goal distribution
  formulation. We build on previous results in the literature by formally
  framing our approach as an instance of planning as inference. We additionally
  derive reductions of several common planning objectives as special cases of
  our probabilistic planning framework. Our experiments demonstrate the
  flexibility of probability distributions as a goal representation on a variety
  of problems including planar navigation among obstacles, intercepting a moving
  target, and a 7-DOF robot arm reaching to grasp an object. We additionally
  demonstrate the applicability of goal distributions in the domain of
  manipulation skill planning.
  
\end{abstract}

\IEEEpeerreviewmaketitle

\section{Introduction}
\label{sec:intro}

Goals enable a robot to act with intention in its environment and provide an
interface for humans to specify the desired behavior of the robot. Defining and
representing goals is therefore a fundamental step in formalizing robotics
problems~\cite{lavalle2006planning,sutton2018reinforcement}. Goals are most
commonly represented as elements of a (possibly infinite) subset
\(\mathcal{G} \subseteq \mathcal{X}\) of the robot's state space
\(\mathcal{X}\). In practice, it is common to select a particular goal state
\(\g \in \mathcal{G}\) to pursue, or to simply define the goal set as the
singleton \(\mathcal{G} = \{\g\}\).

The goal state \(\g\) is typically incorporated into a goal-parameterized cost
function \(C_{\g}: \mathcal{X} \rightarrow \mathbb{R}\) over the robot's state
space \(\mathcal{X}\). For example, \(C_{\g}\) may be defined as a
goal-parameterized distance function
\(d_{\g}:\mathcal{X} \rightarrow \mathbb{R}\) (e.g. Euclidean distance) between
the goal state \(\g\) and the current state \(\x_t \in \mathcal{X}\). The cost
function \(C_{\g}\) is often used to monitor progress to a point-based
goal~\cite{baranes2013active}, bias graph creation in sampling-based motion
planners~\cite{lavalle2006planning}, or used directly as an objective for
trajectory optimization~\cite{lambert2020stein}. In the latter case, an action
cost is often further incorporated so that
\(C_{\g}: \mathcal{X} \times \mathcal{U} \rightarrow \mathbb{R}\) can encode,
for example, distance to goal while enforcing smoothness constraints on the
actions \(u_t \in \mathcal{U}\) from the robot's action space \(\mathcal{U}\).

The point-based goal formulation just described provides formal elegance but
overlooks many common challenges of robotic systems, including stochastic
dynamics, numerical imprecision, continuous state and action spaces, and
uncertainty in the robot's current state and goal. Each of these phenomena make
it difficult for a robot to arrive in a desired state in a verifiable manner. As
a result, it is common to utilize a catch-all fuzzy notion of goal achievement:
the robot achieves its goal \(\g\) if it arrives in a terminal state
\(\x_T \in \mathcal{X}\) such that \(d_{\g}(\x_T) < \varepsilon\) where
\(\varepsilon \in \mathbb{R}^+\) is a small tolerance governing how much error
is permissible. This common relaxation fails to differentiate the various
sources of uncertainty and imprecision a robot faces in pursuing a goal.

\begin{figure}[t!]
  \centering
  \includegraphics[width=0.485\textwidth]{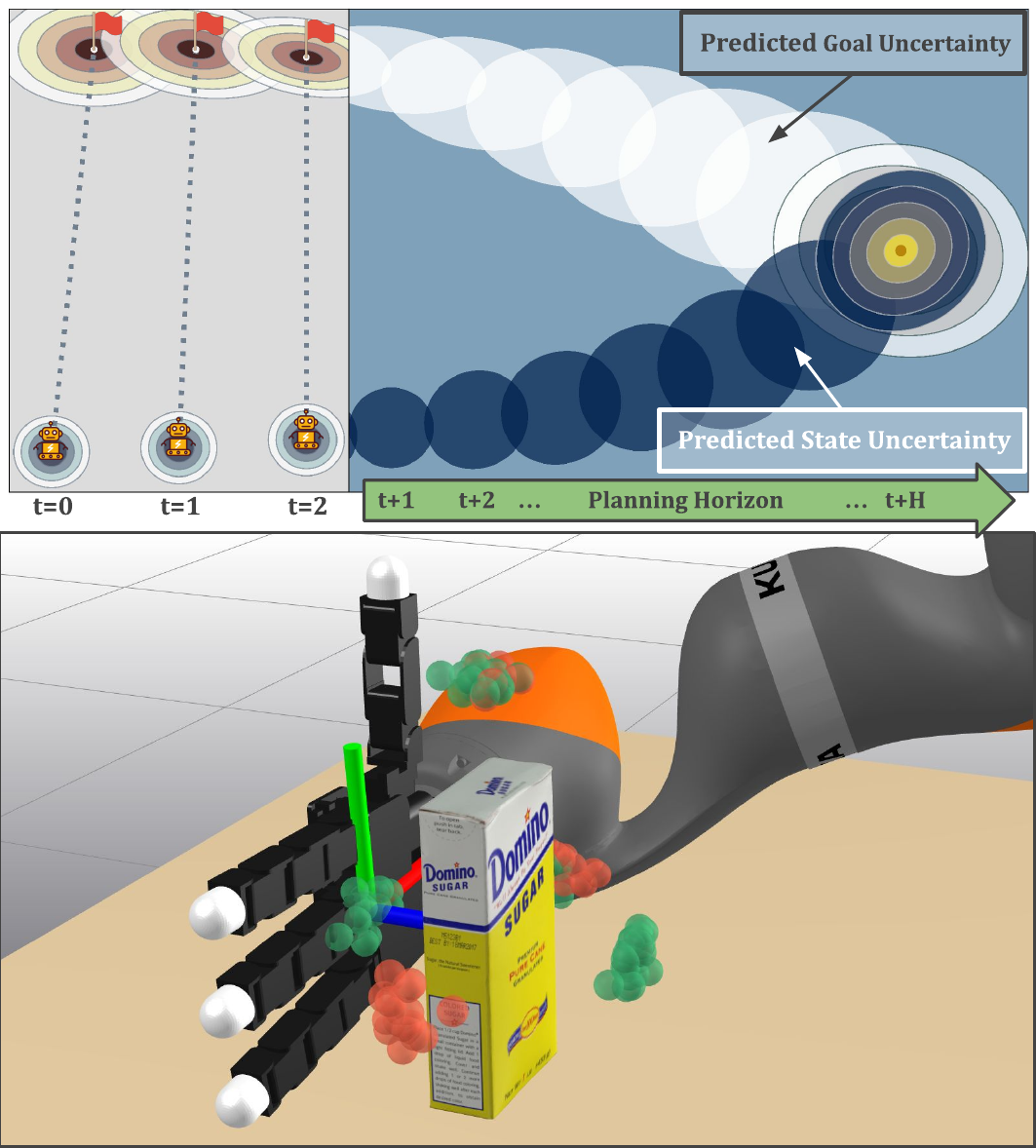}
  \caption{Examples of goals with associated models of uncertainty. \\
  \textbf{(Upper Row)} A robot navigates to intercept a moving target (red flag)
  while updating a Gaussian belief about the target's location as it acquires
  new observations (timesteps \(t=0\) to \(t=2\) in gray region). The robot
  predicts both the future goal uncertainty (white ellipses) and its own state
  uncertainty (blue ellipses) over the planning horizon (light blue region) to
  plan a path that will intercept the target (associated experiments in
  Sec.~\ref{sec:moving_target}). \\ \textbf{(Lower Row)} A robot arm reaches to
  grasp an object while inferring a reachable target pose. The goal distribution
  is modeled as a pose mixture model where some samples are reachable (green
  spheres) and others are unreachable (red spheres) based on the kinematics of
  the arm (associated experiments in Sec.~\ref{sec:arm_reaching}).}
  \label{fig:cover}
\end{figure}


We advocate for probability distributions as a more suitable goal
representation. Goals as probability distributions extend the traditional notion
of set-based goals to include a measure of belief (i.e. uncertainty) that a
point in the state space satisfies the goal condition. Goal uncertainty arises
frequently in robotics applications whenever the robot must predict its goal,
e.g. estimating a goal from noisy sensors~\cite{kuffner2003online}, forecasting
a dynamic goal~\cite{kim2014catching, maeda2017probabilistic}, or in
generalizing a learned goal representation~\cite{akgun2016simultaneously,
conkey2019active, koert2020incremental, paraschos2018using}. See
Fig.~\ref{fig:cover} for a couple of examples we examine experimentally in this
article. Goal distributions are also a natural representation in many domains,
e.g. mixture models represent targets for distributed multi-agent
systems~\cite{foderaro2012decentralized, rudd2017generalized} and encode
heuristic targets for grasping objects~\cite{conkey2019active,
lu-isrr2017-grasp-inference, miller2003automatic}. Goal distributions formally
subsume traditional point-based and set-based goals as Dirac-delta and uniform
distributions, respectively. In spite of the myriad use cases and benefits of
goal distributions in various domains, planning and reinforcement learning
frameworks continue to primarily rely on point-based goal representations, and
no formal treatment of planning to goal distributions exists.

To address this gap, we formalize the use of probability distributions as a goal
representation for planning (Sec.~\ref{sec:problem_statement}) and discuss
encoding common goal representations as distributions
(Sec.~\ref{sec:goal_distributions}).  As our main theoretical contribution in
this article, we derive a framework (Sec.~\ref{sec:goal_distribution_planning})
for planning to goal distributions under state uncertainty as an instance of
\textit{planning as inference}~\cite{levine2018reinforcement,
rawlik2012stochastic, toussaint2010bayesian}. Planning as inference seeks to
infer an optimal sequence of actions by conditioning on achieving optimality and
inferring the posterior distribution of optimal trajectories. We derive a
planning as inference objective that enables the robot to plan under state
uncertainty to a goal state distribution by minimizing an information-theoretic
loss between its predicted state distribution and the goal distribution. We show
that several common planning objectives are special cases of our probabilistic
planning framework with particular choices of goal distributions and loss
function (Sec.~\ref{sec:cost_reductions}). We provide a practical and novel
algorithm (Sec.~\ref{sec:practical_algorithm}) that utilizes the
\textit{unscented transform}~\cite{uhlmann1995dynamic} for state uncertainty
propagation to efficiently compute the robot's terminal state distribution and
expected running cost.

We apply our approach to a variety of different problems, goal distributions,
and planning objectives (Sec.~\ref{sec:applications}). We showcase the
flexibility of probability distributions as a goal representation on the problem
of planar navigation among obstacles (Sec.~\ref{sec:dubins}). These results also
exhibit the ease with which our planning framework accommodates different models
of goal uncertainty simply by swapping in different goal distributions and
information-theoretic losses. We provide an example of how our planning approach
can leverage sources of uncertainty in the environment to achieve a target state
distribution (Sec.~\ref{sec:amplifier}) in a ball-rolling task. We also apply
our approach to the problem of intercepting a moving target in which the agent
updates its belief of the goal as it acquires noisy observations of the target
(Sec.~\ref{sec:moving_target}). We investigate a higher-dimensional problem in
which a 7-DOF robot arm reaches to grasp an object
(Sec.~\ref{sec:arm_reaching}), where we model target end-effector poses as a
mixture of pose distributions about the object. We show that we are able to plan
to reachable poses using this distribution directly as our goal representation
without checking for reachability, while a point-based goal requires computing
inverse kinematics to check reachability prior to planning. Finally, we apply
our approach to the more challenging domain of manipulation skill planning
(Sec.~\ref{sec:skill_planning}), which requires learning a dynamics model to
support planning over the contact dynamics involved in dynamic manipulations. We
show we are able to accommodate more dynamic skills than are typically used in
skill planning while also explicitly representing the uncertainty inherent to
common manipulation goals.

We conclude the article in Sec.~\ref{sec:conclusion}, where we discuss the
opportunities to expand the use of goal distributions in other existing planning
as inference frameworks (e.g. reinforcement and imitation learning).


\section{Related Work}
\label{sec:related_work}

We first review works on planning as inference, since we formally situate the
problem of planning to goal distributions as an instance of planning as
inference. Planning as inference has proven to be a powerful framework to
formalize problems in robotics~\cite{attias2003planning, lambert2020stein,
levine2018reinforcement, mukadam2018continuous, rawlik2012stochastic,
toussaint2010bayesian, watson2021advancing}. It constitutes a Bayesian view of
stochastic optimal control~\cite{lambert2020stein, rawlik2012stochastic,
toussaint2010bayesian} from which popular decision-making algorithms can be
derived such as max-entropy reinforcement
learning~\cite{levine2018reinforcement} and sampling-based solvers like
model-predictive path integral control (MPPI)~\cite{bhardwaj2021storm,
lambert2020stein, williams2017model} and the cross-entropy method
(CEM)~\cite{kobilarov2012cross, lambert2020stein}. The probabilistic perspective
of planning as inference enables elegant problem definitions as factor graphs
that can be solved by message-passing
algorithms~\cite{toussaint2010bayesian,watson2021advancing} and non-linear least
squares optimizers~\cite{mukadam2018continuous}. Goal distributions have not
been explicitly considered in the planning as inference framework. Part of our
contribution in this article is formulating the problem of planning to goal
distributions as an instance of planning as inference, thereby connecting to the
rich literature on planning as inference and enabling access to a variety of
existing algorithms to solve the problem.

As mentioned in Sec.~\ref{sec:intro}, goal distributions have cropped up in
various sub-domains of robotics, but the topic has not yet received systematic
attention. The closest work we are aware of is~\cite{nasiriany2021disco} which
advocates for goal-distribution-conditioned policies as an extension to
goal-conditioned policies in reinforcement learning. A limited class of goal
distributions are considered in~\cite{nasiriany2021disco} and the formulation is
specific to reinforcement learning. We view the present article as complementary
to~\cite{nasiriany2021disco}, where we consider more general classes of goal
distributions and develop an approach for planning to goal distributions in a
unified way.

Target distributions have been utilized in reinforcement learning to encourage
exploration~\cite{lee2019efficient} and expedite policy
learning~\cite{andrychowicz2017hindsight}. \textit{State marginal
matching}~\cite{lee2019efficient} learns policies for which the state marginal
distribution matches a target distribution (typically uniform) to encourage
targeted exploration of the environment. Goal distributions have also been used
as sample generators for goal-conditioned reinforcement
learning~\cite{nair2018visual, pong2019skew}. Recent improvements on
\textit{hindsight experience replay} (HER)~\cite{andrychowicz2017hindsight} have
sought to estimate goal distributions that generate samples from low density
areas of the replay buffer state distribution~\cite{kuang2020goal,
pitis2020maximum, zhao2019maximum} or that form a curriculum of achievable
goals~\cite{ren2019exploration, pitis2020maximum}. We note that HER-based
methods are typically used in multi-goal reinforcement learning where states are
interpreted as goals irrespective of their semantic significance as a task goal.

A closely related but distinct area of research is \textit{covariance
steering}~\cite{hotz1987covariance} which optimizes feedback controller
parameters and an open-loop control sequence to move a Gaussian state
distribution to a target Gaussian distribution. Recent work on covariance
steering has focused on satisfying chance constraints~\cite{okamoto2018optimal,
pilipovsky2021chance, ridderhof2020chance}, including generating
constraint-satisfying samples for MPPI~\cite{yin2021improving}. Covariance
steering typically assumes linear dynamics and is limited to Gaussian state
distributions, where the objective is often to ensure the terminal state
covariance is fully contained within the goal state covariance. Covariance
steering also decouples control of the distribution into \textit{mean steering}
and \textit{covariance steering}, which assumes homogeneous uncertainty and
accuracy of control over the state space. These assumptions do not hold for most
robotics domains. For example, visual odometry estimates degrade when entering a
dark room~\cite{thrun2005probabilistic}, and it is harder to maneuver over
varied terrain~\cite{dahlkamp2006self}. In contrast, our approach easily
accommodates non-linear dynamics, non-homogeneous state uncertainty, and
non-Gaussian state distributions.

We utilize information-theoretic costs in our approach as they are fitting
objectives for optimizing plans to goal distributions. A variety of information
theoretic costs have been utilized in planning and policy optimization which we
briefly review here. Probabilistic control design~\cite{karny1996towards}
minimizes Kullback-Leibler (KL) divergence between controlled and desired
state-action distributions and bears some similarity to planning as inference
previously described. KL divergence is also commonly used to constrain
optimization iterations in policy search~\cite{chebotar2017path,
peters2010relative, schulman2015trust} and
planning~\cite{abdulsamad2017state}. Broader classes of divergences including
\(f\)-divergence~\cite{belousov2017fdivergence, ghasemipour2019divergence,
ke2020imitation} and Tsallis divergence~\cite{wang2021variational} have also
been utilized for policy improvement~\cite{belousov2017fdivergence}, imitation
learning~\cite{ghasemipour2019divergence, ke2020imitation}, and stochastic
optimal control~\cite{wang2021variational}. Stochastic optimal control and
planning techniques often seek to minimize the expected cost of the
trajectory~\cite{deisenroth2011pilco, williams2017information} or maximize the
probability of reaching a goal set~\cite{blackmore2010probabilistic,
lew2020safe}. Entropy of a stochastic policy is utilized in maximum-entropy
reinforcement learning~\cite{haarnoja2017reinforcement, haarnoja2018soft,
abdulsamad2017state} and inverse reinforcement
learning~\cite{ziebart2008maximum} to prevent unnecessarily biasing the policy
class. We demonstrate the use of cross-entropy and KL divergence in our
formulation of planning to goal distributions. However, our approach is general
enough to admit other information-theoretic losses between the robot's predicted
state distribution and a goal distribution. We discuss this point further in
Sec.~\ref{sec:conclusion}.


\section{Problem Statement}
\label{sec:problem_statement}

We focus our work on planning problems with continuous state and action spaces
in which the robot must reach a desired goal while acting under stochastic
dynamics. These problems fall in the domain of stochastic optimal control and
planning~\cite{stengel1994optimal, rawlik2012stochastic}. We consider a robot
with continuous state space \(\mathcal{X} \subseteq \mathbb{R}^{N_x}\) and
continuous action space \(\mathcal{U} \subseteq \mathbb{R}^{N_u}\) and focus on
a discrete time setting, although extending to continuous time would be
straightforward. The stochastic dynamics function
\(f:\mathcal{X} \times \mathcal{U} \times \Omega \rightarrow \mathcal{X}\)
determines the resulting state \(\x_{t+1} = f(\x_t, \u_t, \bom_t)\) from
applying action \(\u_t \in \mathcal{U}\) in state \(\x_t \in \mathcal{X}\) at
time \(t\) subject to noise \(\bom_t \sim \Omega\).

Planning to a goal state \(\g \in \mathcal{X}\) in this setting requires the
agent to utilize the dynamics function \(f\) to find a sequence of actions
\(\u_0, \dots \u_{T-1} \in \mathcal{U}^T\) from its initial state\footnote{We
use \(\x_0\) to indicate the state from which the agent is planning, but note
the timestep is arbitrary and replanning from any timestep is permissible, as is
common in model-predictive control schemes~\cite{lambert2020stein,
williams2017model}.}  \(\x_0 \in \mathcal{X}\) to the goal state \(\g\).  Due to
sensor noise and partial observability, the robot rarely knows its current state
precisely and must therefore plan from an initial estimated distribution of
states \(\hat{p}(\x_0)\), e.g. as the output of a state estimator like a Kalman
filter~\cite{thrun2005probabilistic}.  Typically the robot must minimize the
expected cost under state uncertainty and stochastic dynamics defined by a cost
function \(C: \mathcal{X} \times \mathcal{U} \rightarrow \mathbb{R}\).  As noted
in Sec.~\ref{sec:intro}, the cost function typically includes a distance
function parameterized by \(\g\) to induce goal-seeking behavior.

Instead of planning to particular states, we consider the more general problem
of planning to a \textit{goal state distribution}. A goal distribution
\(p(\x | \g=1)\) encodes the robot's belief (i.e. uncertainty) that a particular
state \(\x\) belongs to the goal set \(\mathcal{G}\). Using Bayes rule, we see
for any particular \(\x\) the goal density is proportional to the goal
likelihood \(p(\x | \g=1) \propto p(\g=1|\x)\). We abbreviate the goal
distribution as \(p_{\g}(\x)\). In this article, we assume the goal distribution
is given in order to focus our efforts on formalizing the problem of planning to
goal distributions. We note that goal distributions can be set as desired if
known in parametric form, which can be as simple as adding uncertainty bounds to
a point-based goal, e.g. adding a Gaussian covariance or setting uniform bounds
on a region centered about a target point. Goal distributions can also be
estimated from data~\cite{akgun2016simultaneously, conkey2019active,
koert2020incremental, paraschos2018using}. We discuss this point further in
Sec.~\ref{sec:conclusion}.

Given a goal distribution, we require two main ingredients to generate a plan to
it. First, we need a means of predicting the robot's terminal state distribution
after following a planned sequence of actions. This is a form of state
uncertainty propagation which we discuss further in
Sec.~\ref{sec:uncertainty_propagation}. Second, we require a loss function
\(\mathcal{L}:\mathcal{Q} \times \mathcal{P} \rightarrow \mathbb{R}\) that
quantifies the difference between distributions \(q\in\mathcal{Q}\) and
\(p \in \mathcal{P}\), where \(\mathcal{Q}\) and \(\mathcal{P}\) are arbitrary
families of distributions. In particular, we are interested in quantifying the
difference between the robot's terminal state distribution and the goal state
distribution. Typically \(\mathcal{L}\) will take the form of a statistical
divergence (e.g. KL divergence, total variation, etc.), but we leave open the
possibility for other losses which may not meet the formal definition of a
divergence (e.g. cross-entropy).

We can now formally state our problem as minimizing the information-theoretic
loss \(\mathcal{L}\) between the terminal state distribution
\(q(\x_T \mid \X_{T-1}, \U_{T-1})\) and \(p_{\g}(\x_T)\), where we abbreviate
\(\U_t \doteq (\u_0, \dots, \u_t)\), and \(\X_t \doteq (\x_0, \dots, \x_t)\). We
formulate this as the following constrained optimization problem
\begin{subequations}
  \label{opt:problem}
\begin{alignat}{3}
  &\argmin_{\pi} &\quad
  & \mathcal{L}\left(q(\x_T \mid \X_{T-1}, \U_{T-1}), p_{\g}(\x_T)\right) \label{opt:goal_measure}\\
  &&&+ \mathbb{E}_{q(\X_{T-1}, \U_{T-1})}\left[ \sum_{t=0}^{T-1} c_t(\x_t, \u_t) \right] \label{opt:running_cost} \\
  &\text{s.t.} && \X_T \in \mathcal{X}^T, \U_{T-1} \in
  \mathcal{U}^T \label{opt:state_act_const} \\
  &&& q(\x_0) = \hat{p}(\x_0) \label{opt:init_state_dist}
\end{alignat}
\end{subequations}
where \(\pi\) defines the policy being optimized. In its simplest form, the
policy can just be a sequence of actions \(\pi = \U_{T-1}\), but we note that
more general policy parameterizations can also be utilized. The sequence of
states \(\X_t\) is induced by the dynamics function \(f\) together with the
sequence of actions \(\U_t\). Eq.~\ref{opt:goal_measure} is the loss between the
terminal state distribution under the policy and the goal
distribution. Eq.~\ref{opt:running_cost} is the expected cost accumulated over
the trajectory, where
\(c_t:\mathcal{X}\times\mathcal{U}:\rightarrow \mathbb{R}\) encapsulates
arbitrary running costs. Eq.~\ref{opt:state_act_const} ensures states and
actions are from the robot's state-action space. Eq.~\ref{opt:init_state_dist}
is a constraint that ensures planning initiates from the robot's belief about
its initial state.

We present a more concrete instantiation of this optimization problem in
Sec.~\ref{sec:uncertainty_propagation} where we present a tractable method for
computing the terminal state distribution utilized in the information-theoretic
loss in Eq.~\ref{opt:goal_measure}. Before diving into the details of our
planning formulation in Sec.~\ref{sec:goal_distribution_planning}, we first
define some useful goal distributions in Sec.~\ref{sec:goal_distributions}.


\section{Goal Distributions}
\label{sec:goal_distributions}

We present a selection of goal distributions we explore in this article. Our list
is by no means exhaustive and we emphasize there are likely domains that benefit
from less standard distributions~\cite{list_of_distributions}.

\subsection{Dirac-delta}
\label{sec:dirac_delta}
A Dirac-delta distribution has a density function with infinite density at its
origin point \(\p\) and zero at all other points:
\begin{equation}
  \label{eq:dirac_delta}
  \delta_\p(\x) =
  \begin{cases}
    \infty & \text{if } \x = \p \\
    0 & \text{otherwise}
  \end{cases}
\end{equation}
such that \(\int_\mathcal{X}\delta_{\p}(\x)d\x = 1\) over its domain
\(\mathcal{X}\). The Dirac-delta is a probabilistic representation of a
point-based goal~\cite{candido2011minimum}. Point-based goals are the most
common goal representation in both planning~\cite{lavalle2006planning} and
reinforcement learning~\cite{sutton2018reinforcement}.

\subsection{Uniform}
\label{sec:uniform}
A uniform distribution is a probabilistic representation of a set-based goal and
has the density function
\begin{equation}
  \label{eq:uniform}
  \mathbb{U}_A(\x) =
  \begin{cases} \frac{1}{\texttt{vol}(A)} & \text{if } \x \in A \\ 0 &
    \text{otherwise}
  \end{cases}
\end{equation}
where \(\texttt{vol}(A)\) defines the volume of the set \(A\). Uniform
distributions are useful for encoding a bounded region of acceptable goal states
without any preference to any particular state within that region. Examples
include navigating to be in a particular room~\cite{xia2021relmogen}, or placing
an object on a desired region of a table surface~\cite{curtis2021long}. Uniform
goal distributions have also proven useful as goal sample generators,
particularly when learned to bias goal-conditioned policies to reach desirable
states~\cite{pong2019skew}.

\subsection{Gaussian}
\label{sec:gaussian}
A Gaussian distribution has the density function
\begin{equation}
  \label{eq:gaussian}
  \mathcal{N}(\x \mid \mean, \bSi) =
  \frac{1}{\sqrt{(2\pi)^d |\bSi|}} \exp \left( -\frac{1}{2} d_M(\x; \bmu, \bSi)^2 \right)
\end{equation}
where \(\mean\) and \(\bSi\) define the mean and covariance of the distribution,
respectively, and
\(d_M(\x; \bmu, \bSi) = \sqrt{(\x - \mean)^T \bSi^{-1} (\x - \mean)}\) is the
\textit{Mahalanobis distance}~\cite{mahalanobis1936generalized}. Gaussian
distributions have been utilized in learning from demonstration to encode goals
learned in a data-driven fashion from sub-optimal
experts~\cite{akgun2016simultaneously, conkey2019active, koert2020incremental,
paraschos2018using}. Gaussians also naturally encode uncertainty the agent might
have about its goal, e.g. in dynamic tasks like object
handovers~\cite{maeda2017probabilistic} and catching moving
objects~\cite{kim2014catching}, or estimating a goal online from noisy
observations, e.g. footstep planning~\cite{kuffner2003online}.

We also consider a truncated Gaussian~\cite{tallis1961moment} distribution with
density function
\begin{equation}
  \label{eq:truncated_gaussian}
  \mathcal{N}(\x \mid \mean, \bSi, \mathcal{A}) =
  \begin{cases}
    \frac{\exp\left( -\frac{1}{2} d_M(\x; \bm{\mu}, \bm{\Sigma})^2
    \right)}{\int_{\mathcal{A}} \exp\left( -\frac{1}{2} d_M(\x; \bm{\mu},
      \bm{\Sigma})^2 \right) \mathrm{d}\x} & \text{if } \x \in \mathcal{A} \\
    0 & \text{otherwise}
  \end{cases}
\end{equation}
which is a common model for bounded Gaussian
uncertainty~\cite{cozman1994truncated}.

\subsection{Bingham}
\label{sec:bingham}
The Bingham distribution~\cite{bingham1974antipodally} is an antipodally
symmetric distribution on the unit hypersphere \(\mathbb{S}^d\) with density
function
\begin{equation}
  \label{eq:bingham_pdf}
  \mathcal{B}(x \mid \Lambda, V) = \frac{1}{F(\Lambda)} \exp \left( \sum_{i=1}^d
  \lambda_i \left( \bm{v}_i^T \x \right)^2 \right)
\end{equation}
where \(\x \in \mathbb{S}^d \subset \mathbb{R}^{d+1}\) is constrained to the
unit hypersphere \(\mathbb{S}^d\), \(\Lambda\) is a diagonal matrix of
concentration parameters, the colums of \(V\) are orthogonal unit vectors, and
\(F(\Lambda)\) is a normalization constant. The Bingham density function bears
resemblance to the multivariate Gaussian density function because it is derived
from a zero-mean Gaussian conditioned to lie on the unit
hypersphere~\(\mathbb{S}^d\).

The Bingham distribution on \(\mathbb{S}^3\) is of particular interest to
robotics as it encodes a Gaussian distribution over orientations in
3D~\cite{glover2013tracking}. The key difficulty in utilizing the Bingham
distribution is computing the normalization constant \(F(\Lambda)\) since no
closed-form solution exists. In spite of this, efficient and accurate
approximations have enabled use of the Bingham distributions for 6-DOF object
pose estimation~\cite{glover2012monte}, tracking of moving
objects~\cite{glover2013tracking}, and pose uncertainty quantification in deep
neural networks~\cite{peretroukhin2020smooth}.

\subsection{Mixture Models}
\label{sec:mixture_models}
Mixture models comprise a weighted combination of multiple probability
distrutions. A Gaussian mixture model (GMM) is the most commonly used mixture
model with density function
\begin{equation}
  \label{eq:mixture_model}
  \mathcal{M}\left(\x \mid \{\alpha_i, \bmu_i,
    \bSi_i\}_{i=1}^M\right) = \sum_{i=1}^M \alpha_i \mathcal{N}(\x \mid \bmu_i, \bSi_i)
\end{equation}
where \(\{\alpha\}_{i=1}^M\) are mixture weights associated with each component
such that \(\sum_{i=1}^M \alpha_i = 1\). GMMs have been used to encode goals
learned in a data-driven fashion where a single mode does not suffice, such as
the desired pre-grasp pose to pick up an obect~\cite{conkey2019active,
lu-isrr2017-grasp-inference, miller2003automatic}. Mixture models also provide a
natural goal representation for distributed multi-agent
systems~\cite{foderaro2012decentralized, rudd2017generalized}. We highlight that
mixture models are not limited to mixtures of Gaussians and that other
distribution families are possible. For example, a mixture of uniform
distributions probabilistically encodes a union of disjoint goal
sets~\cite{bhatia2011goals}.

\subsection{Goal Likelihood Classifier}
\label{sec:goal_classifier}
We can model the likelihood that a state achieves goal
\(p(\g=1|\x) = f(\x; \theta)\) using a discriminative classifier, such as
logistic regression, parameterized by \(\theta\). This model offers great
flexibility~\cite{fu2018variational} and can be used to model complex goal
relations such as planning to achieve grasps for multi-fingered
hands~\cite{lu-ram2020-grasp-inference} and deformable object manipulation from
image observations~\cite{singh2019end}.


\section{Planning to Goal Distributions}
\label{sec:goal_distribution_planning}

Our proposed use of goal distributions discussed in
Sec.~\ref{sec:problem_statement} fits naturally within the methods of
\textit{planning and control as probabilistic
inference}~\cite{levine2018reinforcement, rawlik2012stochastic}. In this
section, we first present a background on traditional planning as inference
frameworks in Sec~\ref{sec:planning_as_inference}. We then provide our novel derivation and theoretical analysis of planning to goal distributions within the planning as inference framework Sec.~\ref{sec:distribution_pai}.

\subsection{Planning as Inference}
\label{sec:planning_as_inference}

Planning as inference leverages a duality between optimization and probabilistic
inference for motion control and planning
problems~\cite{levine2018reinforcement, rawlik2012stochastic,
toussaint2010bayesian}. In the planning as inference framework, we consider
distributions of state-action trajectories \(p(\tau)\) where
\(\tau = \left(\X_{T}, \U_{T-1}\right)\). We introduce a binary random variable
\(\O_\tau \in \{0, 1\}\) where values of \(\O_\tau=1\) and \(\O_\tau=0\) denote
whether a trajectory \(\tau\) is optimal or not, respectively. We use the
notation \(\O_\tau\) to represent \(\O_\tau = 1\) for brevity, and similarly we
use \(\O_t\) to represent \(\O_t = 1\) to denote optimality at a particular
timestep in the trajectory \(\tau\). We treat optimality as an observed quantity
and seek to infer the posterior distribution of optimal trajectories
\(p(\tau \mid \O_\tau)\). Algorithms based on variational inference are commonly
used to optimize a proposal distribution \(q(\tau)\) from a known
family~\(\mathcal{Q}\) (e.g. exponential) by solving the following minimization:
\begin{equation}
  \label{eq:pai_objective}
  q^* = \argmin_{q\in \mathcal{Q}} \kl{q(\tau)}{p(\tau \mid \O_\tau)}
\end{equation}
This minimization is equivalently solved by (details in
Appendix~\ref{app:variational_inference})

\begin{equation}
  \label{eq:pai_min_neg_elbo}
  q^* = \argmin_{q\in \mathcal{Q}} \underbrace{-\E_q \left[ \log p(\mathcal{O}_\tau \mid \tau) \right]}_{\mathrm{T}_1}
  + \underbrace{\kl{q(\tau)}{p_0(\tau)}}_{\mathrm{T}_2}
\end{equation}
where we have labeled the first and second terms in
Eq.~\ref{eq:pai_min_neg_elbo} as \(\mathrm{T}_1\) and \(\mathrm{T}_2\),
respectively, for ease of reference in
Sec.~\ref{sec:distribution_pai}.

The objective in Eq.~\ref{eq:pai_min_neg_elbo} seeks to maximize the
log-likelihood of being optimal in expectation under the trajectory while being
regularized by a state-action trajectory prior \(p_0(\tau)\). A salient example
of a trajectory prior from the literature is a Gaussian process prior to ensure
trajectory smoothness~\cite{mukadam2018continuous}. The state-action trajectory
distribution \(p_0(\tau)\) is induced by the stochastic policy prior
\(\pi_0(\u_t|\x_t)\). We elaborate further on different prior policies in
Sec.~\ref{sec:distribution_pai}.

The likelihood \(p(\O_\tau \mid \tau)\) in Eq.~\ref{eq:pai_min_neg_elbo} is key
to planning as inference, as it connects the optimization to task-specific
objectives. The likelihood may be set to any density function, but it is most
commonly set as the exponentiated negative cost~\cite{rawlik2012stochastic}:
\begin{equation}
  \label{eq:pai_likelihood}
  p(\O_\tau \mid \tau) = \exp(-\alpha C(\tau))
\end{equation}
for \(C(\tau) = c_\mathrm{term}(\x_T) + \sum_{t=0}^{T-1} c_t(\x_t, \u_t) \)
where \(c_\mathrm{term}(\cdot)\) is a terminal cost function defined for the
final timestep in the planning horizon and \(c_t(\cdot, \cdot)\) is the cost
function for all other timesteps.

We have so far presented planning as inference in its standard formulation akin
to~\cite{rawlik2012stochastic} and~\cite{lambert2020stein}. We now turn to our
novel contributions to incorporate goal distributions in planning as inference.

\subsection{Goal Distributions in Planning as Inference}
\label{sec:distribution_pai}
We examine incorporating goal distributions into the planning as inference
framework just described and derive its relation to the optimization problem
from Eq.~\ref{opt:problem}.

We first address the optimality likelihood \(p(\O_\tau \mid \tau)\). We define
optimality at the terminal state to mean reaching the goal, i.e.
\(p(\O_T) = \pg(\x_T)\).  We then define the trajectory optimality likelihood as
\begin{equation}
  \label{eq:likelihood_ours}
  p(\O_\tau \mid \tau) = \pg(\x_T)\exp\left(-\alpha\sum_{t=0}^{T-1}c_t(\x_t, \u_t)\right)
\end{equation}
which captures our proposed notion of optimality, namely satisfying high
probability under the goal distribution density function \(\pg(\x_T)\) while
accounting for arbitrary running costs over the rest of the trajectory. We note
this is equivalent to defining
\(c_{\mathrm{term}} = -\frac{1}{\alpha}\ln \pg(\x_T)\) in
Eq.~\ref{eq:pai_likelihood}. However, since \(\pg(\x_T)\) is a density function,
we find it more appropriate to directly incorporate it as a factor in the
optimality likelihood.

We now consider the implications of using the optimality likelihood from
Eq.~\ref{eq:likelihood_ours} by plugging it into
Eq.~\ref{eq:pai_min_neg_elbo}. Plugging Eq.~\ref{eq:likelihood_ours} into
\(\mathrm{T}_1\) we get
\begin{align}
  \mathrm{T}_1 &= -\E_q \left[ \log
    \left(\pg(\x_T)\exp\left(-\alpha\sum_{t=0}^{T-1}c_t(\x_t, \u_t)\right)
    \right) \right] \\
  &= -\E_q \left[ \log \pg(\x_T) +
    \logexp\left(-\alpha\sum_{t=0}^{T-1}c_t(\x_t, \u_t)\right) \right] \\
  &= \underbrace{\E_q \left[ -\log \pg(\x_T)\right]}_{\mathrm{T}_3} +
  \underbrace{\E_q \left[\alpha\sum_{t=0}^{T-1}c_t(\x_t, \u_t)
    \right]}_{\mathrm{T}_4} \label{eq:reduced_expected_optimality}
\end{align}
where \(\mathrm{T}_4\) is simply the expected running cost accumulated over
non-terminal timesteps of the trajectory, which we inherit from the standard
planning as inference formulation. Our formulation differs for the final
timestep with term \(\mathrm{T}_3\) which we further expand as
\(\mathrm{T}_3 = \E_q \left[ -\log \pg(\x_T)\right] = \E_{q(\x_T \mid \tau)}
\left[ -\log \pg(\x_T)\right] \). This quantity is the \textit{cross-entropy} of
\(\pg(\x_T)\) with respect to the terminal state distribution
\(q(\x_T \mid \tau)\) which we denote by
\(\mathcal{H}(q(\x_T \mid \tau), \pg(\x_T))\).

Recombining the terms above, we restate the planning as inference objective from
Eq~\ref{eq:pai_objective} for the case of planning to goal distributions as
\begin{align}
  \label{eq:pai-goal-dist}
  q^* = \argmin_{q\in \mathcal{Q}} &\mathcal{H}(q(\x_T \mid \tau), \pg(\x_T)) +
                                     \E_q \left[\alpha\sum_{t=0}^{T-1}c_t(\x_t,
                                     \u_t) \right] \nonumber \\
  &+ \kl{q(\tau)}{p_0(\tau)}
\end{align}
\begin{remark}
  Planning as inference for planning to goal distributions is equivalent to
  minimizing the cross-entropy between the terminal state distribution and the
  goal distribution while minimizing expected running costs and regularizing to
  a prior state-action distribution.
\end{remark}

We now discuss some special cases of this result and relate the objective from
Eq.~\ref{eq:pai-goal-dist} to the optimization problem we defined in
Eq.~\ref{opt:problem}.

\subsubsection{Deterministic Policy}
If we assume a deterministic policy
\(\pi(\u_t \mid \x_t) = \delta_{\u_t = \phi(\x_t)}\), then the KL regularization
term in Eq.~\ref{eq:pai-goal-dist} reduces to
\(\kl{q(\tau)}{p_0(\tau)} = \E_{q} \left[-\sum_{t=0}^{T-1} \log \pi_0(\u_t|\x_t)
\right]\) (see Appendix~\ref{app:pai_alternative_derivation} for details) and we
simplify Eq.~\ref{eq:pai-goal-dist} to
\begin{align}
  \label{eq:pai-goal-dist-det}
  q^* = \argmin_{q\in \mathcal{Q}} &\mathcal{H}(q(\x_T \mid \tau), \pg(\x_T)) +
                                     \E_q \left[\alpha\sum_{t=0}^{T-1}c_t(\x_t,
                                     \u_t) \right] \nonumber \\
  &+ \E_{q} \left[-\sum_{t=0}^{T-1} \log \pi_0(\u_t|\x_t)\right]
\end{align}
We assume a deterministic policy in the rest of this article. We discuss
extensions for reinforcement learning later in Sec.~\ref{sec:conclusion}.

\subsubsection{Uniform Prior Policy}
If we specify a uniform distribution for the policy prior
\(\pi_0(\u_t \mid \x_t) = \mathcal{U}(\u_t)\), its value becomes constant with
respect to the optimization~\cite{rawlik2012stochastic}, reducing the problem to
exactly that defined in Eq.~\ref{opt:goal_measure} with the cross entropy as
loss
\(\mathcal{L}(q(\x_T \mid \tau), \pg(x)) = \mathcal{H}(q(\x_T \mid \tau),
\pg(x))\).
\begin{remark}
  Planning as inference for planning to goal distributions with a uniform prior
  policy is equivalent to solving the planning to goal distribution problems
  with a cross entropy loss.
\end{remark}

\subsubsection{Maximum Entropy Terminal State}
\label{sec:max_ent_terminal}
We consider setting the loss in Eq.~\ref{opt:goal_measure} to be the
KL-divergence, \(\kl{q(\x_T|\tau)}{\pg(x)}\). Then we have the following
objective for the planning to goal distribution problem
\begin{align}
   q^* =& \argmin_{q \in \mathcal{Q}} \kl{q(\x_T|\tau)}{\pg(\x_T)} + \E_q \left[\alpha\sum_{t=0}^{T-1}c_t(\x_t, \u_t) \right]\\
          =& \argmin_{q \in \mathcal{Q}} \mathcal{H}(q(\x_T|\tau),\pg(\x_T)) + \E_q \left[\alpha\sum_{t=0}^{T-1}c_t(\x_t, \u_t) \right] - \notag\\
           &\mathcal{H}(q(\x_T|\tau)) \label{eq:max-ent-planning-cost}
\end{align}
following from the relation
\(\mathcal{H}(p_1, p_2) = \text{D}_{\text{KL}}(p_1 \parallel p_2) +
\mathcal{H}(p_1)\).  If we set Eq.~\ref{eq:max-ent-planning-cost} equal to
Eq.~\ref{eq:pai-goal-dist-det} we see the first two terms cancel and we are left
with the equality
\begin{equation}
  \label{eq:max-ent-prior}
  \mathcal{H}(q(\x_T|\tau)) = \E_q \left[\sum_{t=0}^{T-1} \log \pi_0(\u_t|\x_t)\right]
\end{equation}
\begin{remark}
  Planning as inference for planning to goal distributions with a prior policy
  that maximizes entropy of the terminal state is equivalent to solving the
  planning to goal distributions problem with a KL divergence loss.
\end{remark}

\subsubsection{M-Projections for Planning to Finite Support Goals}
\label{sec:projections}

The KL divergence in the variational inference objective for planning as
inference in Eq.~\ref{eq:pai_objective} is formulated as an information
projection. Since KL divergence is an asymmetric loss between distributions,
there are two possible projections an optimizer can solve to minimize KL
divergence:
\begin{align}
  q^* &= \argmin_q \kl{q(x)}{p(x)} \label{eq:i_projection} \tag{\textbf{I-projection}} \\
  q^* &= \argmin_q \kl{p(x)}{q(x)} \label{eq:m_projection} \tag{\textbf{M-projection}}
\end{align}
The information projection (I-projection) exhibits mode-seeking behavior while
the moment projection (M-projection) seeks coverage of all regions where
\(p(x) > 0\) and thus exhibits moment-matching
behavior~\cite{murphy2012machine}. Importantly, when \(p(x)\) has finite support
(e.g. uniform, Dirac-delta, truncated Gaussian), it is necessary to use an
M-projection to avoid the division by zero that would occur in the I-projection
over regions outside the support of \(p(x)\).

Our formulation has so far only considered an I-projection. In general, the
M-projection is intractable to solve for arbitrary planning as inference
problems. This is due to the fact that one would require access already to the
full distribution of optimal trajectories \(p(\tau \mid \O_\tau)\) in order to
compute the optimization. However, since we do assume access to the goal
distribution, we \textit{can} compute either the I-projection or M-projection
for the KL divergence at the terminal timestep. Thus as a final objective for
investigation, we examine setting the distributional loss in
Eq.~\ref{opt:goal_measure} to be the M-projection KL divergence, instead of the
I-projection as previously examined. We get the following objective:
\begin{align}
  \pi^* =& \argmin_{\pi} \kl{\pg(\x_T)}{q(\x_T|\tau)} \notag \\
  &\hspace{1.5cm}+ \E_q \left[\alpha\sum_{t=0}^{T-1}c_t(\x_t, \u_t) \right]
  \label{eq:kl-m-proj-loss}
\end{align}
This is nearly equivalent to the planning as inference with a maximimum-entropy
prior result from Sec.~\ref{sec:max_ent_terminal}, however the KL divergence
term is now an M-projection. A key advantage this affords us is our method
naturally accommodates goal distributions with finite support, which we explore
in our experiments in Sec.~\ref{sec:dubins}. We thus have a unified framework
for planning to arbitrary goal distributions under uncertain dynamics utilizing
information-theoretic loss functions.

Note that for a fixed goal distribution, the M-projection of cross-entropy is
equivalent to the objective in Eq.~\ref{eq:kl-m-proj-loss} since the entropy of
the goal distribution is constant with respect to the decision
variables. However, this is not necessarily the case for goal distributions that
may change over time based on the agent's observations. We discuss this point
further in Sec.~\ref{sec:conclusion}.


\section{Cost Reductions}
\label{sec:cost_reductions}

We present an additional theoretical contribution to illustrate how our
probabilistic planning framework encompasses common planning objectives in the
literature. We examine the cross-entropy cost term (term \(T_3\) in
Eq.~\ref{eq:reduced_expected_optimality}) between the predicted terminal state
distribution \(q(\x_T \mid \tau)\) and the goal distribution \(p_{\g}(\x_T)\)
for several common distribution choices. Looking at both the I-projection
\(\inlinece{q(\x_T \mid \tau)}{p_{\g}(\x_T)}\) and the M-projection
\(\inlinece{p_{\g}(\x_T)}{q(\x_T \mid \tau)}\), we reduce the costs to commonly
used cost functions from the literature.

\subsection{(Weighted) Euclidean Distance}
For a Gaussian goal distribution
\(p_{\g}(\x_T) = \mathcal{N}(\x_T|\mean_{\g}, \bSi_{\g})\) and minimizing
the I-projection of cross-entropy we have:
\begin{align}
  \pi^* &= \argmin_{\pi} \E_{q(\tau)} \left[ - \log \mathcal{N}(\x_T|\mean_{\g}, \bSi_{\g}) \right ] \\
        &= \argmin_{\pi} \E_{q(\tau)} \left[ - \log \exp\{-(\x_T - \mean_{\g})^T \bSi_{\g}^{-1} (\x_T - \mean_{\g})\} \right ] \\
        &= \argmin_{\pi} \E_{q(\tau)} \left[ (\x_T - \mean_{\g})^T \bSi_{\g}^{-1} (\x_T - \mean_{\g}) \right ] \\
        &= \argmin_{\pi} \E_{q(\tau)} \left[ \parallel \x_T - \mean_{\g}\parallel^2_{\bSi_{\g}^{-1}} \right ] \\
        &= \argmin_{\pi} \E_{q(\tau)} \left[ \parallel \x_T - \mean_{\g}\parallel^2_{\bm{\Lambda}_{\g}} \right ]
\end{align}
For the case of deterministic dynamics, that is
\(q(\x_T|\tau) = \delta_{x_T | \tau}(\x)\), the expectation simplifies to a
single point evaluation and we recover the common weighted Euclidean
distance. Since it is weighted by the precision (i.e. inverse covariance) of the
goal distribution, it is equivalent to the Mahalanobis
distance~\cite{mahalanobis1936generalized}. The same cost arises for the case of
a goal point (i.e. Dirac delta distribution) and Gaussian state uncertainty if
we consider the M-projection of cross-entropy. In this case, the distance is
weighted by the precision of the terminal state distribution instead of the goal
precision.

\subsection{Goal Set Indicator}
For a uniform goal distribution
\(p_{\g}(\x) = \mathbb{U}_{\mathcal{G}}(\x)\) and deterministic dynamics
\(q(\x_T|\tau) = \delta_{x_T | \tau}(\x)\), minimizing the I-projection of
cross-entropy amounts to
\begin{align}
  \pi^* &= \argmin_{\pi} \E_{q(\tau)} \left[ - \log \mathbb{U}_{\mathcal{G}}(\x) \right ] \\
        &= \argmin_\pi \int_{\x \in \{\x_T\}} -\delta_{\x_T|\tau}(\x) \log \mathbb{U}_{\mathcal{G}}(\x) \mathrm{d}\x \\
        &= \argmin_\pi
          \begin{cases}
            -\log u_\mathcal{G} & \textit{if } \x_{T} \in \mathcal{G}\\
            \infty & \textit{otherwise}
          \end{cases}
\end{align}
where \(u_\mathcal{G} = 1/\texttt{vol}(\mathcal{G})\) as described in
Sec.~\ref{sec:uniform}. Hence the minimum is obtained with a constant cost if
the terminal state $\x_T$ from executing trajectory $\tau$ reaches any point in
the goal set, while any state outside the goal set receives infinite cost. Note
the function is non-differentiable as expected from the set-based goal
definition. We can treat a single goal state naturally as a special case of this
function. The non-differentiable nature of this purely set-based formulation
motivates using search-based and sampling-based planners over optimization-based
approaches in these deterministic settings.

\subsection{Chance-Constrained Goal Set}
\label{sec:chance_constrained}
For a uniform goal distribution
\(p_{\g}(\x) = \mathbb{U}_{\mathcal{G}}(\x)\), minimizing the M-projection
of cross-entropy amounts to
\begin{align}
  \pi^* &= \argmin_\pi \E_{\mathbb{U}_{\mathcal{G}}(\x)} \left[ - \log q(\x_T \mid \tau) \right] \\
        &= \argmin_\pi \int_{\x \in \mathcal{G}} -\mathbb{U}_{\mathcal{G}}(\x) \log q(\x_T \mid \tau) \mathrm{d}\x \\
        &= \argmin_\pi -u_{\mathcal{G}} \int_{\x \in \mathcal{G}} \log q(\x_T \mid \tau) \mathrm{d}\x \label{eq:neg-log-chance-constrained} \\
        &= \argmax_\pi \int_{\x \in \mathcal{G}} q(\x_T \mid \tau) \mathrm{d}\x  \label{eq:chance-constrained}
\end{align}
which defines the probability of reaching any state in the goal set
\(\mathcal{G}\), a commonly used term for reaching a goal set in
chance-constrained control (e.g. Equation~(6) in
\cite{blackmore2010probabilistic}).

\subsection{Maximize Probability of Reaching Goal Point}
\label{sec:max_prob_goal}
A special case of the previous result in Sec.~\ref{sec:chance_constrained} is a
Dirac-delta goal distribution \(p_{\g}(\x) = \delta_{\g}(\x)\) instead of a
uniform goal distribution. We get
\begin{equation}
  \pi^* = \argmax_\pi q(\x_T = \g \mid \tau)
\end{equation}
which maximizes the probability of reaching a point-based goal \(\g\) following
trajectory \(\tau\).


\section{Practical Algorithm}
\label{sec:practical_algorithm}

In this section, we formulate planning to goal distributions as a practical
instantiation of our method described in
Sec.~\ref{sec:goal_distribution_planning}. We first describe how we compute the
agent's predicted terminal state distribution with the unscented transform in
Sec.~\ref{sec:uncertainty_propagation}. In
Sec.~\ref{sec:ut_optimization_problem} we incorporate the unscented transform
uncertainty propagation into a more concrete formulation of the constrained
optimization problem defined in Eq.~\ref{opt:problem} from
Sec.~\ref{sec:problem_statement}. We then discuss tractable methods for
computing our information-theoretic losses in Sec.~\ref{sec:approximate_loss}.
We use this formulation in our experiments in Sec~\ref{sec:applications}.

\subsection{State Uncertainty Propagation}
\label{sec:uncertainty_propagation}

As noted in Sec.~\ref{sec:problem_statement}, a robot typically maintains a
probabilistic estimate of its state \(\hat{p}(\x_0)\) to cope with alleotoric
uncertainty from its sensors and the environment~\cite{kendall2017uncertainties,
thrun2005probabilistic}. In order to compute the terminal state distribution
\(q(\x_T \mid \tau)\) and in turn compute the information-theoretic losses for
Eq.~\ref{opt:goal_measure}, we require a means of propagating the state
uncertainty over the planning horizon given the initial state distribution
\(\hat{p}(\x_0)\), a sequence of actions \(\U_{T-1}\), and the robot's
stochastic (possibly nonlinear) dynamics function
\(f:\mathcal{X} \times \mathcal{U} \times \Omega \rightarrow \mathcal{X}\).

Our choice of uncertainty propagation method is informed by the evolution of
nonlinear Bayesian filters~\cite{thrun2005probabilistic,
watson2021advancing}. Perhaps the simplest method is Monte Carlo sampling,
i.e. sampling initial states \(\x_0 \sim \hat{p}(\x_0)\) and sequentially
applying the dynamics function to acquire a collection of samples from which the
terminal state distribution can be approximated. However, this approach can
require thousands of samples to estimate the distribution well, which imposes a
computational burden when estimating auxiliary costs and constraints
(e.g. collision constraints). Other options include Taylor series approximations
(akin to the extended Kalman filter) and numerical quadrature
methods~\cite{watson2021advancing}. When the marginal state distribution is
well-estimated by a Gaussian, the \textit{unscented transform} provides an
accurate estimate of the state distribution propagated under nonlinear
dynamics~\cite{howell2021direct} (akin to the unscented Kalman
filter~\cite{uhlmann1995dynamic, julier1997new}). We leverage the unscented
transform in our approach since we can achieve an accurate estimate of the
propagated state marginal using only a small set of deterministically computed
query points without having to compute Jacobians or Hessians of the dynamics
function~\cite{thrun2005probabilistic}.

Given a state distribution \(\mathcal{N}(\x_t \mid \bmu_t, \bSi_t)\), the
unscented transform computes a small set of \textit{sigma points}
\(\mathcal{P}_t = \{\bmu_t \pm \beta \bL_t[i]\}_{i=1}^N\) for state size \(N\)
where \(\bL_t[i]\) is the i-th row of the Cholesky decomposition of the
covariance \(\bSi_t\) and \(\beta\) is a hyperparameter governing how spread out
the sigma points are from the mean. Given an action \(\u_t \in \mathcal{U}\), we
estimate how the distribution will transform under the robot's dynamics by
evaluating the noise-free dynamics function, \(f(\x_t, \u_t, 0)\) at each of the
sigma points \(\p_t \in \mathcal{P}_t\) together with the action \(\u_t\) to get
a new set of points
\(\mathcal{P}_{t}^\prime = \{f(\p_t, \u_t, 0) \mid \p_t \in
\mathcal{P}_t\}\). Taking the sample mean and covariance of the points in
\(\mathcal{P}_{t}^\prime\) provides an estimate of the distribution at the next
timestep \(\mathcal{N}(\x_{t+1} \mid \bmu_{t+1}, \bSi_{t+1})\). This procedure
is similar to the unscented dynamics described in~\cite{howell2021direct} as
well as its use in unscented model predictive
control~\cite{farrokhsiar2014unscented, volz2015stochastic}.

We assume Gaussian state distributions and use the unscented transform for
uncertainty propagation in our experiments in Sec.~\ref{sec:applications} to
illustrate our approach. Most popular state estimation techniques assume a
Gaussian state distribution~\cite{thrun2005probabilistic, uhlmann1995dynamic},
so this is not a particularly limiting assumption. However, we emphasize that
other uncertainty propagation techniques are possible in our framework. We
discuss this point further in Sec~\ref{sec:conclusion}.

\subsection{Planning to Goal Distributions with Unscented Transform}
\label{sec:ut_optimization_problem}

We now present a more concrete instantiation of the abstract optimization
problem from Sec.~\ref{sec:problem_statement} which we will subsequently use in
our experiments in Sec.~\ref{sec:applications}. We formulate the optimization
problem as a direct transcription of a trajectory optimization for planning to a
goal distribution \(\pg(\x_T)\) under Gaussian state uncertainty propagated by
the unscented transform as described in Sec.~\ref{sec:uncertainty_propagation}:

\begin{subequations}
\begin{alignat}{3}
  &\min_{\Theta} &\hspace{2pt} 
  &\Loss\big( \mathcal{N}(\x_T \mid \bmu_T, \bSi_T), \pg(\x_T) \big) \label{opt:goal_loss}\\
  &&& \hspace{10pt} + \sum_{t=0}^{t-1} \sum_{\p_t \in \mathcal{P}_t}
  \mathcal{N}(\p_t \mid \bmu_t, \bSi_t) \cdot c_t\left(\p_t^i, \u_t\right) \label{opt:cost} \\
  &\text{s.t.} && \vmin{\x} \le \bmu_t \le \vmax{\x}, \allt \label{opt:bb_mu} \\
  &&& \vmin{\u} \le \u_t \le \vmax{\u}, \alltm \label{opt:bb_u} \\
  &&& \bmu_0 = \x_0, \hspace{5pt} \bSi_0 = \bSi_{\x_0} \label{opt:init} \\
  &&& \bmu_{t} = \frac{1}{|\mathcal{P}_t^\prime|} \sum_{\p_t^\prime \in
  \mathcal{P}_t^\prime} \p_t^\prime, \hspace{5pt} \forall t \in [1,T] \label{opt:empirical_mean} \\
  &&& \bSi_t = \frac{1}{|\mathcal{P}_t^\prime|} \sum_{\p_t^\prime \in
  \mathcal{P}_t^\prime} (\p_t^\prime - \bmu_t) (\p_t^\prime - \bmu_t)^T + \bm{R}_t,
  \hspace{5pt} \forall t \in [1,T] \label{opt:empirical_cov} \\ 
  &&& \bSi_t \in S^{N_x}_{++}, \allt \label{opt:pd_sigma} \\
  &&& g_i(\Theta) \leq 0, \hspace{10pt} \forall i \in [1, N_g] \label{opt:ineq} \\
  &&& h_j(\Theta) = 0, \hspace{10pt} \forall j \in [1, N_h] \label{opt:eq}
\end{alignat}
\end{subequations}
where the decision variables \(\Theta\) include the sequence of actions
\(\U_{T-1}\) and, due to the direct transcription formulation, the sequence of
means \(\bmu_0, \dots, \bmu_T\) and covariances \(\bSi_0, \dots, \bSi_T\)
parameterizing the state distributions over the trajectory.

Eq.~\ref{opt:goal_loss} is an information-theoretic loss between the terminal
state and goal distributions. In our experiments in Sec.~\ref{sec:applications}
we focus on the cross-entropy and KL divergence losses that resulted from our
derivations in Sec.~\ref{sec:distribution_pai}. Eq.~\ref{opt:cost} encapsulates
the arbitrary running costs from Eq.~\ref{opt:running_cost}, adapted here to be
applied to the sigma points resulting from uncertainty propagation with the
unscented transform. Note the costs are weighted by the probability of the sigma
points under the predicted state distribution for the associated timestep.

Regarding constraints, Eqs.~\ref{opt:bb_mu} and~\ref{opt:bb_u} are bound
constraints on the decision variables. Eq.~\ref{opt:init} ensures the initial
state distribution matches the observed initial state
distribution. Eqs.~\ref{opt:empirical_mean} and~\ref{opt:empirical_cov}
constrain the distribution parameters at each timestep to match the empirical
mean and covariance, respectively, estimated from the transformed sigma points
at each timestep \(\mathcal{P}_t^\prime\) as described in
Sec.~\ref{sec:uncertainty_propagation}. As a reminder,
\(\mathcal{P}_{t}^\prime = \{f(\p_t, \u_t, 0) \mid \p_t \in \mathcal{P}_t\}\) is
the set of transformed sigma points under the noise-free dynamics function. The
stochasticity of the dynamics is accounted for by the term \(\bm{R}_t\) in
Eq.~\ref{opt:empirical_cov}.

Eq.~\ref{opt:pd_sigma} ensures the covariance matrices remain symmetric positive
semi-definite. We note that in practice we use \(\bL_t = \texttt{chol}(\bSi_t)\)
for the covariance optimization such that \(\bL_t\bL_t^T = \bSi_t\), which makes
it easier for optimizers to satisfy this constraint. Finally,
Eqs.~\ref{opt:ineq} and~\ref{opt:eq} enable us to incorporate arbitrary
inequality and equality constraints on the decision variables. For example, we
add collision-avoidance constraints on the sigma points computed from the
distribution parameters at each timestep in our experiments in
Sec.~\ref{sec:applications} similar to~\cite{howell2021direct}.

\subsection{Approximate Information-Theoretic Losses}
\label{sec:approximate_loss}

Cross-entropy and KL divergence are two salient information theoretic losses
that came out of our planning as inference derivation in
Sec.~\ref{sec:distribution_pai}, and we primarily utilize these losses in our
experiments in Sec.~\ref{sec:applications}. There are often closed form solutions
for computing these losses (e.g. between two Gaussian
distributions~\cite{iyer_gaussian_ce}). However, there are instances where no
closed form solutions exist and approximations are necessary (e.g. between two
Gaussian mixture models~\cite{hershey2007approximating}, truncated
Gaussians~\cite{cozman1994truncated}). Since we assume Gaussian state
uncertainty in this article, we utilize a simple and accurate approximation
where needed based on the sigma point computation discussed in
Sec.~\ref{sec:uncertainty_propagation}. For a Gaussian distribution
\(p_1(\x) = \mathcal{N}(\x \mid \bmu, \bSi)\) and an arbitrary distribution
\(p_2(\x)\), we compute approximate cross-entropy as
\begin{equation}
  \label{eq:qpprox_ce}
  \approxce{p_1(\x)}{p_2(\x)} = \frac{1}{|\mathcal{P}|} \sum_{\p \in
  \mathcal{P}} -p_1(\p) \log p_2(\p)
\end{equation}
where \(\mathcal{P}\) is the set of sigma points as discussed in
Sec.~\ref{sec:uncertainty_propagation}. A similar approximation follows for KL
divergence. These approximations are based on the unscented approximation
described in~\cite{hershey2007approximating} which demonstrated its accuracy and
efficiency over Monte Carlo estimates. We utilize this approximation in our
experiments when closed form solutions do not exist.


\section{Applications}
\label{sec:applications}

\newcommand\figexamplesW{0.329}  

\begin{figure*}[t!]
  \centering
  \begin{subfigure}{\figexamplesW\textwidth}
    \centering
    \includegraphics[width=\textwidth]{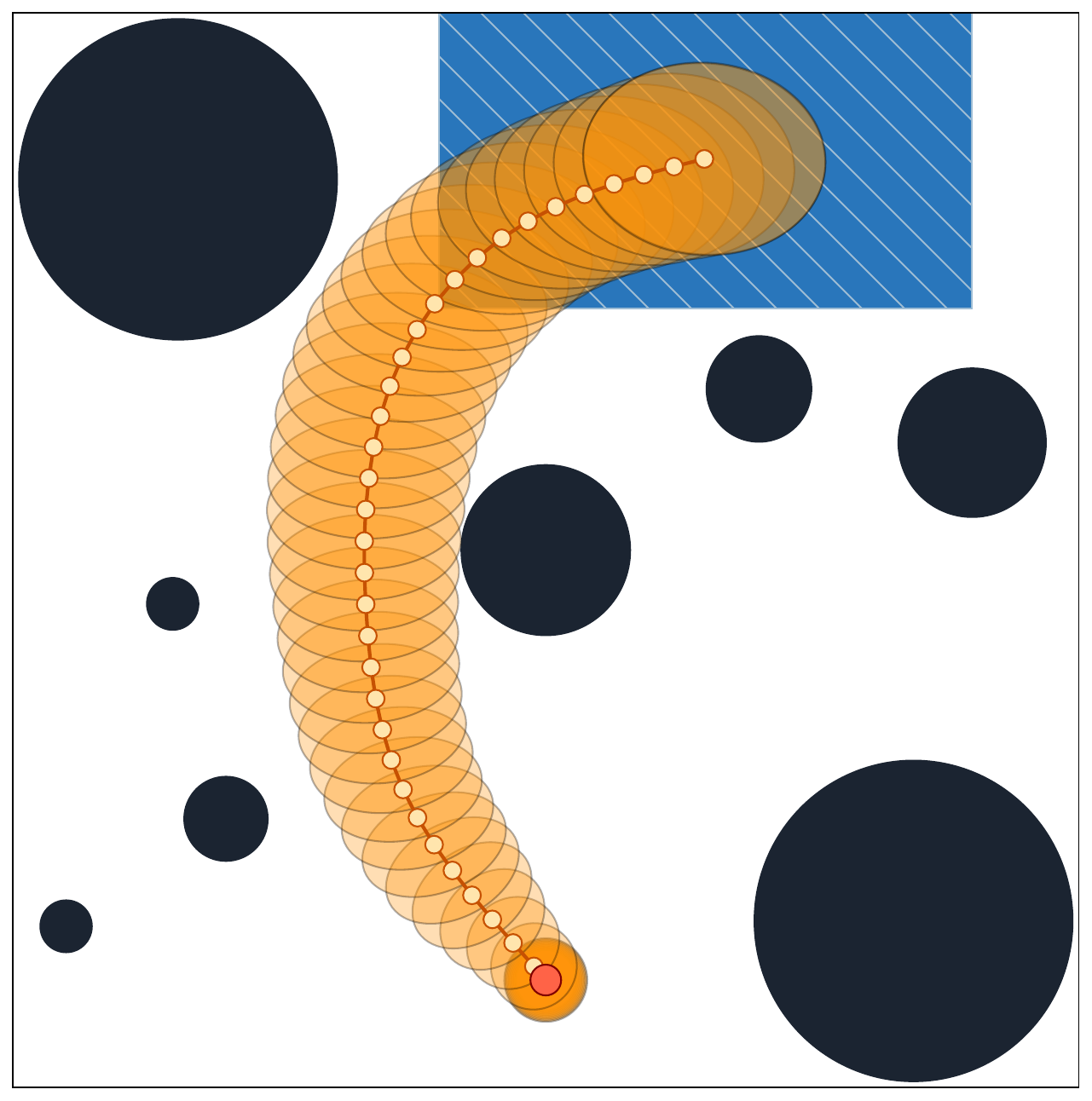}
    \caption{Uniform}
    \label{fig:uniform}
  \end{subfigure}
  \begin{subfigure}{\figexamplesW\textwidth}
    \centering
    \includegraphics[width=\textwidth]{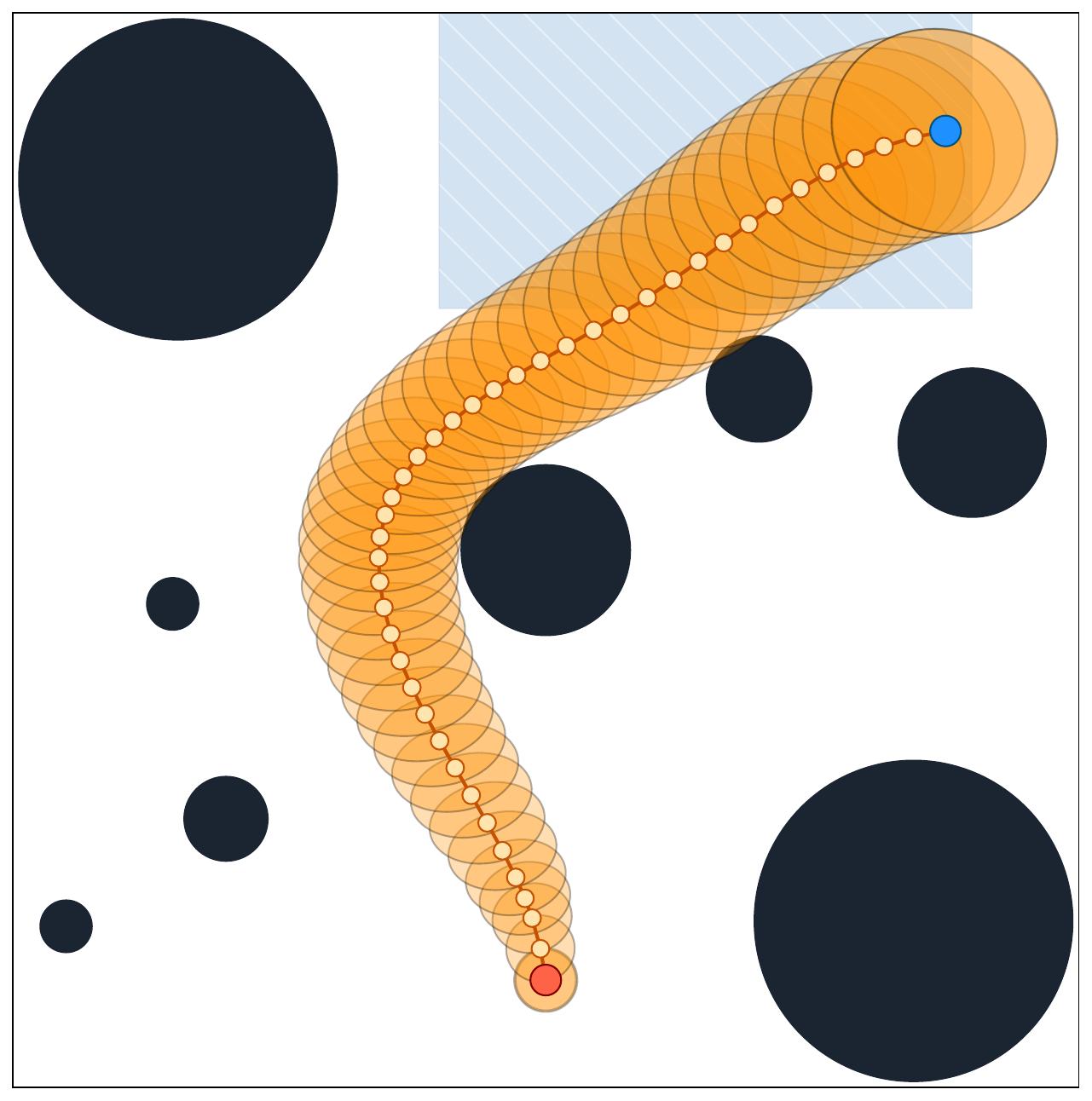}
    \caption{Dirac Delta}
    \label{fig:dirac_delta}
  \end{subfigure}
  \begin{subfigure}{\figexamplesW\textwidth}
    \centering
    \includegraphics[width=\textwidth]{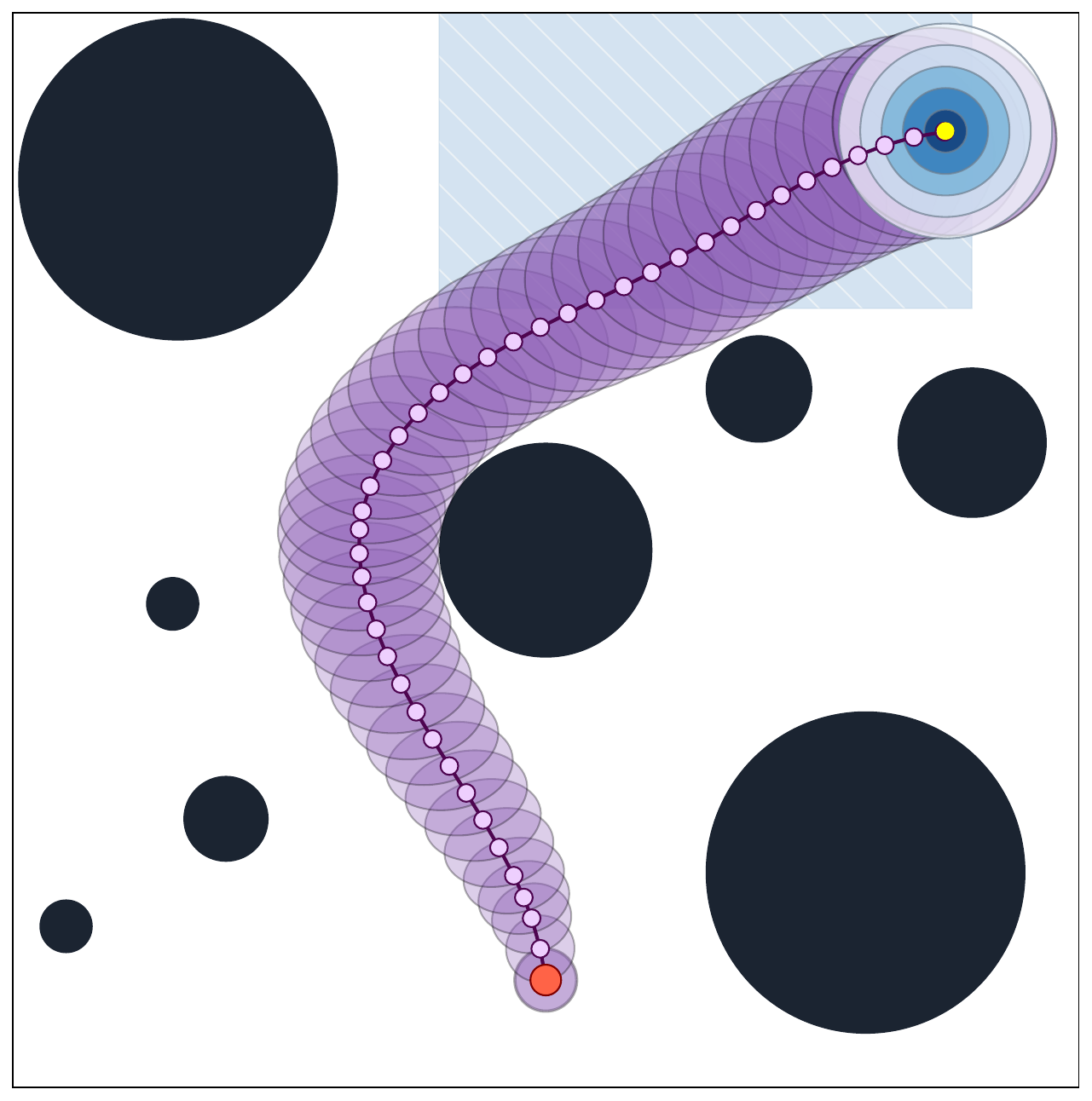}
    \caption{Gaussian}
    \label{fig:gaussian}
  \end{subfigure} 
 
  \begin{subfigure}{\figexamplesW\textwidth}
    \centering
    \includegraphics[width=\textwidth]{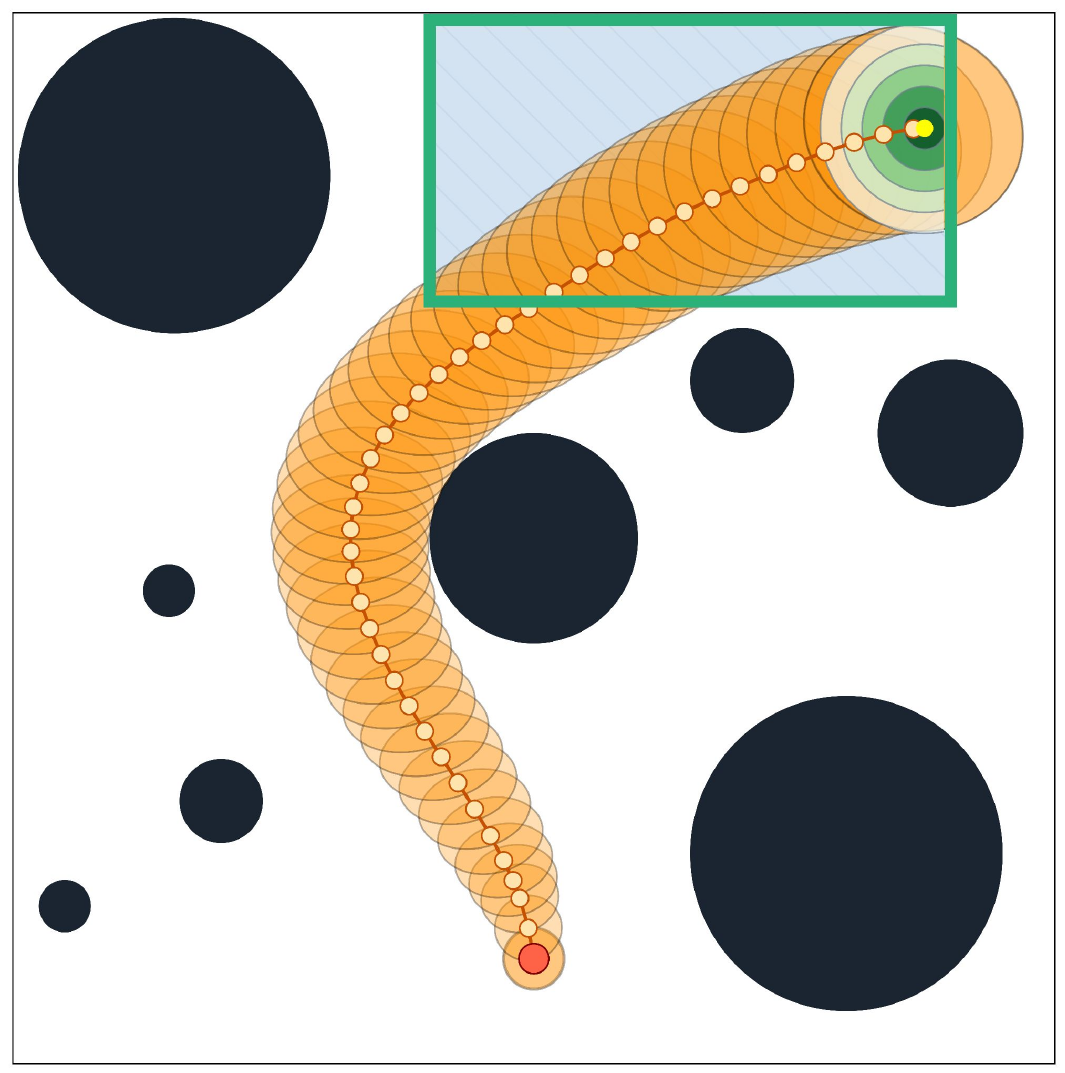}
    \caption{Truncated Gaussian}
    \label{fig:truncated_gaussian}
  \end{subfigure}
  \begin{subfigure}{\figexamplesW\textwidth}
    \centering 
    \includegraphics[width=\textwidth]{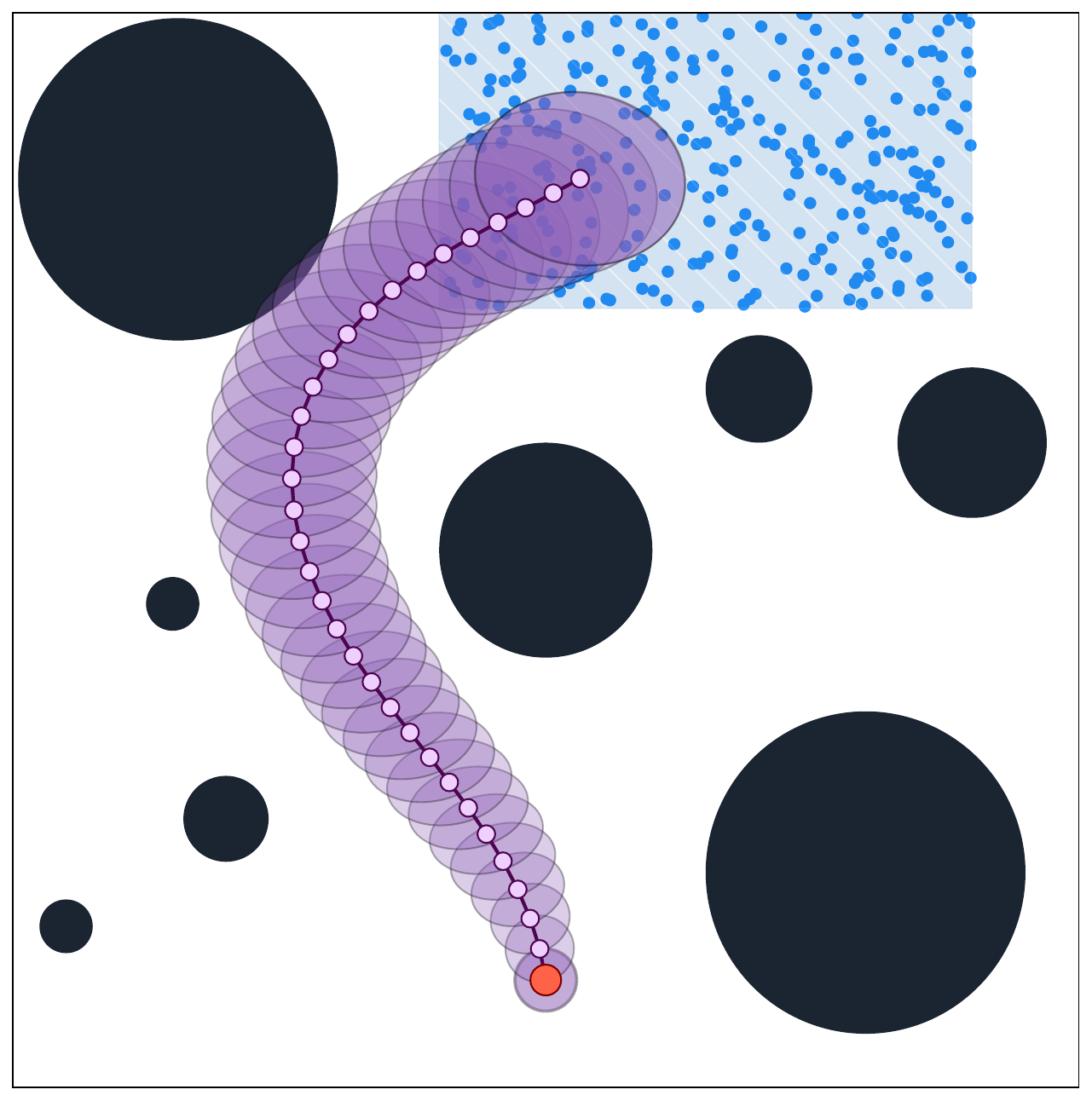}
    \caption{Learned Goal Classifier}
    \label{fig:goal_classifier}
  \end{subfigure}
  \begin{subfigure}{\figexamplesW\textwidth}
    \centering
    \includegraphics[width=\textwidth]{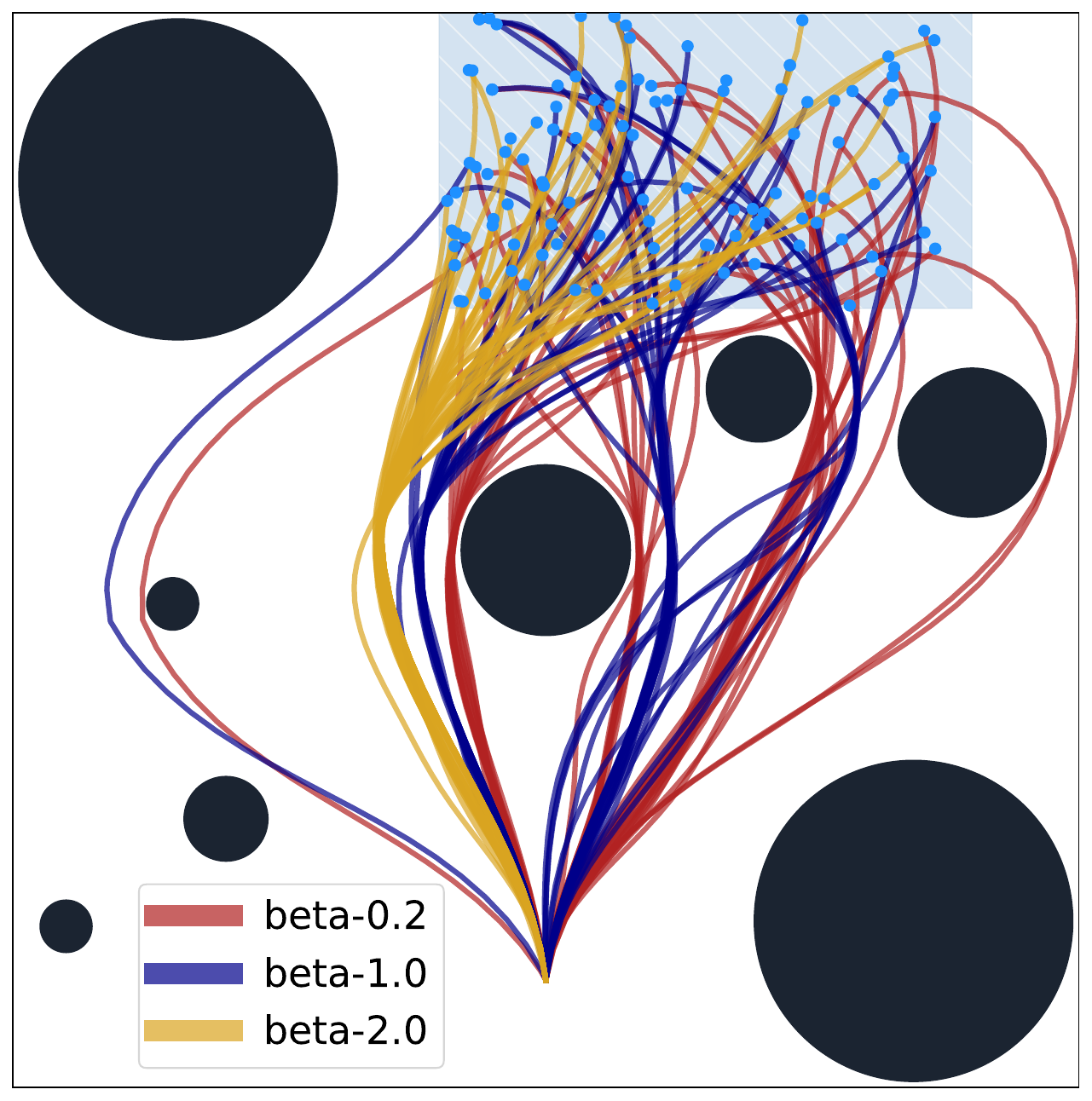}
    \caption{Plans to Goal Samples}
    \label{fig:sample_plans}
  \end{subfigure}
  \caption{ Examples of planned paths with our probabilistic planning framework
  on a 2D navigation environment. In each figure, the robot starts at the red
  dot and must navigate to a goal while avoiding obstacles (black
  spheres). Planned mean paths (dotted lines) and the associated predicted
  covariance (ellipses) are shown. Orange paths were solved with the
  M-projection of cross-entropy and purple paths were solved with the
  I-projection of cross-entropy. \textbf{(a)} A goal set modeled as a uniform
  distribution (blue, hatched rectangle). \textbf{(b)} A goal point modeled as a
  Dirac-delta distribution (blue dot). \textbf{(c)} A Gaussian goal distribution
  (yellow dot is mean, blue ellipses are standard deviations of covariance).
  \textbf{(d)} A truncated Gaussian goal where the PDF is Gaussian inside the
  green rectangle and zero outside of it. \textbf{(e)} A goal classifier learned
  from goal samples (blue dots). \textbf{(f)} Plans to particular goal points
  (Dirac-delta distributions) using values \(\beta=0.2\) (red lines),
  \(\beta=1.0\) (blue lines), and \(\beta=2.0\) (yellow lines).}
  \label{fig:examples}
\end{figure*}


We present several applications in which we demonstrate the utility of goal
distributions as a goal representation in a planning as inference framework. In
Sec.~\ref{sec:dubins} we demonstrate some of the basic behaviors and flexibility
of our approach on a simple planar navigation among obstacles problem. In
Sec.~\ref{sec:amplifier} we further explore the behavior of different
information-theoretic losses on an underactuated planar ball-rolling
environment. We explore a more realistic scenario in
Sec.~\ref{sec:moving_target} in which the agent must intercept a moving target
and update its belief of the target's location with noisy sensor readings. In
Sec.~\ref{sec:arm_reaching}, we apply our approach to a higher dimensional
problem of a 7-DOF robot arm reaching to grasp an object and uncover some
intriguing benefits of a distribution-based goal in a manipulation
setting. Finally, in Sec.~\ref{sec:skill_planning}, we show how goal
distributions naturally encode semantic goals for skill planning, and enable the
use of more dynamic skills than are typically utilized in traditional skill
planning approaches.

Additional details about the different solvers and environment parameters we use
in this section can be found in Appendix~\ref{app:solver_details}. All
associated
code\footnote{A preliminary release of our code can be found here: \\
\texttt{\scriptsize\url{https://bitbucket.org/robot-learning/distribution_planning}}}
and data will be released upon acceptance.

\subsection{Goal distributions are a flexible goal representation}
\label{sec:dubins}

We first apply our approach to a simple 2D navigation problem to highlight the
flexibility of modeling task goals as goal distributions in our probabilistic
planning framework. The task objective is to navigate from a start configuration
to a goal while avoiding obstacles.

We emphasize that in all examples in this section, we use the same planning
algorithm that we formalized in Sec.~\ref{sec:practical_algorithm}. Merely by
changing the family and/or parameterization of the goal distribution in each
example (and selecting an appropriate information-theoretic loss), we are able
to plan to point-based and set-based goals as well as goals with varied models
of uncertainty associated with them.

\subsubsection{Environment Description}

We use the Dubins car model from~\cite{kobilarov2012cross},
a simple vehicle model with non-holonomic constraints in the state space
\(\mathcal{X}=SE(2)\). The state \(\x=(p_x, p_y, \phi)\) denotes the car's
planar position \((p_x, p_y)\) and orientation \(\phi\). The dynamics obey
\begin{equation}
    \dot{p}_x = v\cos\phi, \hspace{5pt}\dot{p}_y = v\sin\phi,
    \hspace{5pt}\dot{\phi} = r
\end{equation}
where \(v \in [0, v_{max}]\) is a linear speed and
\(r \in [-\tan\psi_{max}, \tan\psi_{max}]\) is the turn rate for
\(\psi_{max} \in \left(0, \frac{\pi}{2}\right)\). We use an arc primitive
parameterization similar to~\cite{kobilarov2012cross} where actions
\(\u = (v, r)\) are applied at each timestep for duration \(\Delta t\) such that
the robot moves in a straight line with velocity \(v\) if \(r=0\) and arcs with
radius \(\frac{v}{r}\) otherwise. An action sequence has the form
\(\U_{T-1} = (v_1, r_1, \dots, v_{T-1}, r_{T-1})\). We extend the model
in~\cite{kobilarov2012cross} to have stochastic dynamics by adding Gaussian
noise \(\bm{\omega}_t \sim \mathcal{N}(\bm{x} \mid \bm{0}, \alpha\bm{I})\) to
the state updates at each timestep.

We manually define a Gaussian estimate \(\mathcal{N}(\x_0 \mid \bmu_0, \bSi_0)\)
for the robot's initial state. We compute the marginal terminal state
distribution \(\mathcal{N}(\x_T \mid \bmu_T, \bSi_T)\) using the unscented
transform as described in Sec.~\ref{sec:uncertainty_propagation}. We use
spherical obstacles and utilize the negative signed distance function (SDF) in
an inequality constraint \(-d_{\mathrm{SDF}}(\x_t, \bm{o}_i, r_i) \leq 0\) for
all timesteps \(t \in [0,T]\) where \(T=45\) is the planning horizon and
\((\bm{o}_i, r_i)\) are the origin point \(\bm{o}_i \in \mathbb{R}^2\) and
radius \(r_i\in \mathbb{R}\) of a spherical obstacle. These constraints apply to
the mean trajectory as well as the computed sigma points at each timestep as
described in Sec.~\ref{sec:ut_optimization_problem}.

\subsubsection{Navigating to a goal region}

We define the goal region \(\mathcal{G}\) as the blue, hatched rectangle shown
in Fig.~\ref{fig:uniform}. We are able to plan directly to this region by
modeling it as a uniform goal distribution
\(p_{\g}(\x_T) = \mathbb{U}_\mathcal{G}(\x_T)\). Here we utilize the
M-projection of the cross-entropy loss
\(\inlinece{\mathbb{U}_{\mathcal{G}}(\x_T)}{\mathcal{N}(\x_T \mid \bmu_T,
\bSi_T)}\) since the uniform distribution has finite support as discussed in
Sec.~\ref{sec:uniform} and Sec.~\ref{sec:projections}. As a reminder, this
objective is equivalent to the chance-constrained goal set
objective~\cite{blackmore2010probabilistic} as discussed in
Sec.~\ref{sec:chance_constrained}. The resulting plan is shown in
Fig~\ref{fig:uniform}, where the orange dotted line shows the planned mean path
of the robot and the orange ellipses show two standard deviations of the robot's
predicted Gaussian state uncertainty over the trajectory. This result
demonstrates our probabilistic planning framework accommodates traditional
set-based goal representations. In contrast to traditional planning approaches
that sample a particular point \(\g \in \mathcal{G}\) from the set
\(\mathcal{G}\) to plan to~\cite{lavalle2006planning}, we are able to plan
directly to the region.

While we are able to plan directly to the goal region \(\mathcal{G}\) as just
described, we can also plan to any particular point \(\g \in \mathcal{G}\) (blue
dot in Fig~\ref{fig:dirac_delta}) by modeling the goal point as a Dirac-delta
distribution \(p_{\g}(\x_T) = \delta_{\g}(\x_T)\) as discussed in
Sec.~\ref{sec:dirac_delta}. We again use the M-projection cross-entropy loss
\(\inlinece{\delta_{\g}(\x_T)}{\mathcal{N}(\x_T \mid \bmu_T, \bSi_T)}\) due to
the finite support of the Dirac-delta distribution. As a reminder, this
objective amounts to maximizing the probability of reaching a goal point
(Sec.~\ref{sec:max_prob_goal}). We show the resulting plan for reaching this
arbitrarily, sampled point in Fig~\ref{fig:dirac_delta}. This results shows that
we retain point-based and set-based goal representations in our probabilistic
framework.

We depart from traditional set-based and point-based goal representations in
Fig.~\ref{fig:gaussian} with a Gaussian goal distribution
\(p_{\g}(\x_T) = \mathcal{N}(\x_T \mid \bmu_{\g}, \bSi_{\g})\) as described in
Sec.~\ref{sec:gaussian}. The Gaussian distribution has infinite support and
therefore we use the I-projection of cross-entropy
\(\inlinece{\mathcal{N}(\x_T \mid \bmu_T, \bSi_T)}{\mathcal{N}(\x_T \mid
\bmu_{\g}, \bSi_{\g})}\). The resulting plan is shown in
Fig.~\ref{fig:gaussian}. We set the mean of the Gaussian to the same point
sampled from the previous example for comparison.  We note that in this
particular unimodal example, the M-projection plan looks similar. This result
demonstrates we are able to generate plans to a goal location while explicitly
accounting for the robot's Gaussian belief about which goal point it should
navigate to.

We additionally consider a truncated Gaussian goal distribution
\(p_{\g}(\x_T) = \mathcal{N}(\x_T \mid \mean_{\g}, \bSi_{\g}, \mathcal{G})\)
shown in Fig.~\ref{fig:truncated_gaussian}. We choose the same Gaussian
distribution as in the previous example, but bound the PDF to the goal region
\(\mathcal{G}\) as illustrated by the green rectangle in
Fig.~\ref{fig:truncated_gaussian} such that the PDF is equal to zero outside the
goal region. As discussed in Sec.~\ref{sec:gaussian}, this serves as a model of
bounded uncertainty in which the agent has Gaussian uncertainty about a
point-based goal in the goal region, but the well-defined goal region limits the
agent's belief to be contained within the goal region. Note we again use the
M-projection of cross-entropy
\(\inlinece{\mathcal{N}(\x_T \mid \bmu_{\g}, \bSi_{\g},
\mathcal{G})}{\mathcal{N}(\x_T \mid \bmu_T, \bSi_T)}\) since the truncated
Gaussian has finite support. We also note we use the approximate cross-entropy
discussed in Sec.~\ref{sec:approximate_loss} since there is no closed form
solution for truncated Gaussians.

We now consider a goal classifier learned from data as shown in
Fig.~\ref{fig:goal_classifier}. The blue dots indicate samples that are
considered goals. We acquire negative samples by sampling uniformly from the
environment region outside the goal region. We train a simple neural network
model \(f(\x;\bm{\theta})\) consisting of fully connected layers with ReLU
activations. We utilized two hidden layers of width 32 and trained the model
with binary cross-entropy loss. We again utilize the approximate cross-entropy
loss from Sec.~\ref{sec:approximate_loss}. As discussed in
Sec.~\ref{sec:goal_classifier}, this is a very general goal representation that
can model complex goals, and we can use it directly as a goal in our planning
framework.

\begin{figure}[t!]
  \centering
  \includegraphics[width=0.45\textwidth]{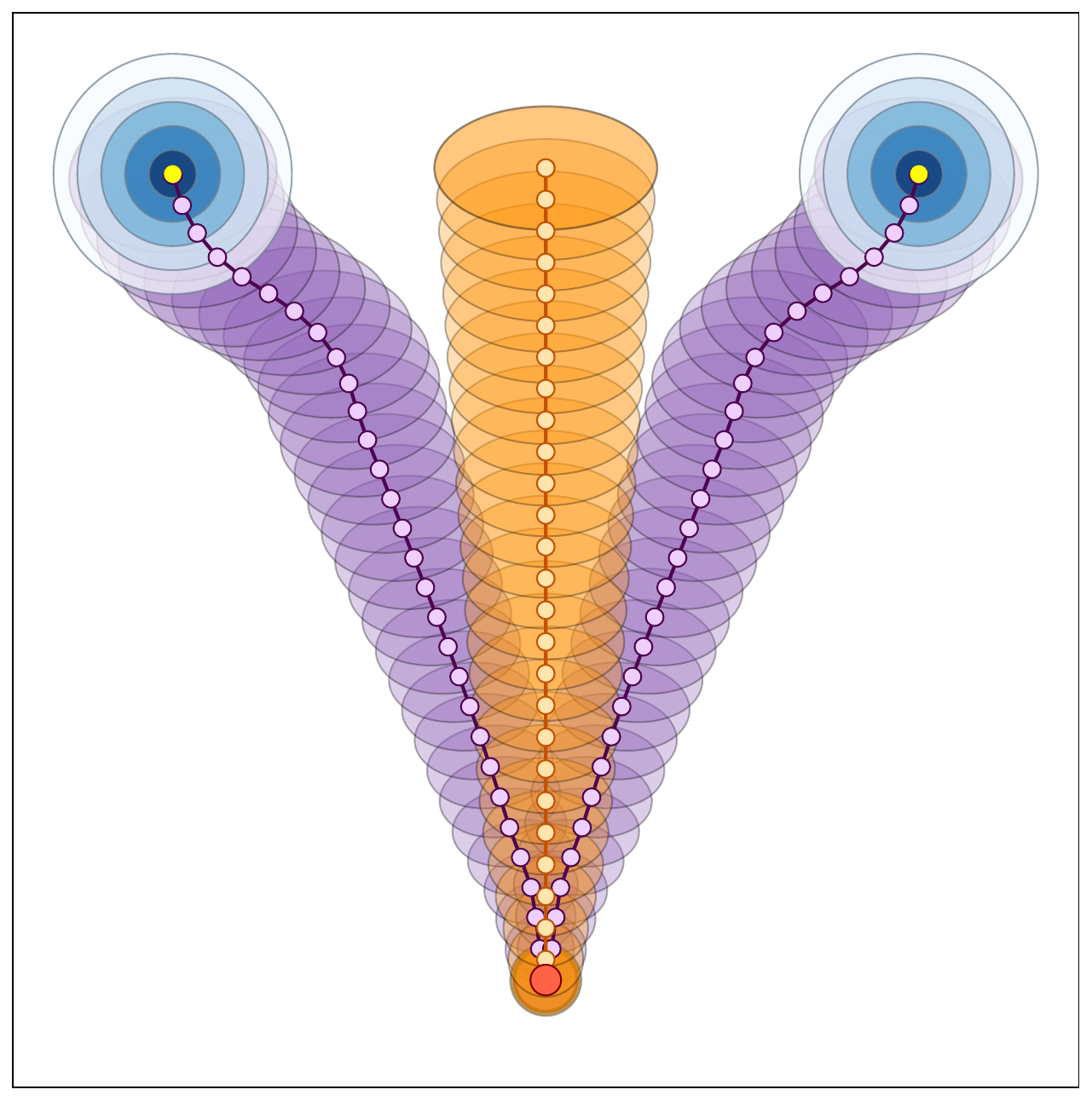}
  \caption{A two-component Gaussian mixture model goal distribution. Yellow dots
  indicate the two modes of goal distribution and blue ellipses show standard
  deviations of the goal covariance. We show two I-projection paths (purple
  paths, one to each component) and one M-projection path (orange path). Purple
  and orange ellipses show the associated state uncertainty for the respective
  paths.}
  \label{fig:gmm_projections}
\end{figure}


For our last example in this section, we show several plans to different
point-based goals (i.e. Dirac-delta distributions) with varying values of the
\(\beta\) parameter discussed in Sec.~\ref{sec:uncertainty_propagation}. The
\(\beta\) parameter governs how spread out the sigma points get in propagating
state uncertainty with the unscented transform. Since we apply collision costs
to the sigma points, the beta parameter in this environment determines how
closely the agent will navigate to obstacles. As shown in
Fig.~\ref{fig:sample_plans}, a value of \(\beta=0.2\) results in more aggressive
plans that navigate closer to obstacles while a value of \(\beta=2.0\) generates
plans that keep a wider berth away from obstacles. Plans for \(\beta=2.0\) all
take a similar route since that is the only homotopy solution class for this
environment that does not result in sigma point collisions. Simply by lowering
the value of beta, we start getting paths in different homotopy classes that can
safely maneuver through more narrow gaps between obstacles. We note that the
variation in solutions largely comes from the initial solution provided to the
solver. Thus, seeding the solver differently can result in paths from different
homotopy classes even when planning to the same goal.

\subsubsection{I-projection vs. M-projection for multimodal goal}
\label{sec:multimodal_goal}

We now look at a Gaussian mixture model (GMM) goal distribution
\(p_{\g}(\x_T) = \mathcal{M}\left(\x \mid \{\alpha_i, \bmu_i,
  \bSi_i\}_{i=1}^M\right)\) as defined in Sec.~\ref{sec:mixture_models}. In
contrast to the unimodal example from Fig.~\ref{fig:gaussian}, the I-projection
and M-projection of cross-entropy for a multimodal GMM exhibit notably different
behavior as shown in Fig.~\ref{fig:gmm_projections}. As discussed in
Sec.~\ref{sec:projections}, the I-projection is mode-seeking and thus is capable
of generating plans to either mode depending on the parameter initialization
provided to the solver. We show one instance of a plan to each GMM mode in
Fig.~\ref{fig:gmm_projections} (purple paths). The M-projection exhibits
moment-matching behavior and will strive to concentrate the mass of the terminal
state distribution to cover all modes of the goal distribution. The M-projection
results in a plan that terminates in between the two modes of the goal
distribution (orange path in Fig.~\ref{fig:gmm_projections}). This is a
well-known feature of the information and moment projections for multimodal
distributions~\cite{murphy2012machine}. However, we find it insightful to
demonstrate this behavior in our probabilistic planning framework. In most
planning problems it is desirable to optimize to a particular mode, and thus the
I-projection will be preferred. However, there are instances where the
M-projection may be desirable for a multimodal distribution, e.g. surveillance
or allocation problems that require an agent to be in proximity to several
targets simultaneously~\cite{batalin2004coverage}.

\subsection{Goal distributions enable leveraging sources of uncertainty for
planning}
\label{sec:amplifier}

In Sec.~\ref{sec:planning_as_inference}, we showed that both cross-entropy and
KL divergence are meaningful information theoretic losses when planning to goal
distributions from the perspective of planning as inference. The examples in
Sec.~\ref{sec:dubins} did not have notable differences between cross-entropy and
KL divergence. This is because in the M-projection cases, the entropy of the
goal distribution is constant in the optimization and thus minimizing
cross-entropy and KL divergence are equivalent. For the I-projections, the
terminal covariance is largely determined by the horizon of the plan, since we
only considered homogeneous dynamics noise over the environment. We now consider
an example with heteroscedastic noise that shows a clear difference between
these two objectives, and compare also to point-based planners.

We consider a 2D ball-rolling environment (pictured in Fig.~\ref{fig:amplifier})
in which the agent must select the initial position and velocity of a ball such
that the ball will end up as close to a target location as possible. Note that
in contrast to the previous environment, the agent only applies control input at
the initial timestep and must leverage the passive dynamics of the environment
to get the ball to the desired target location.

\newcommand\figamplifierW{0.241}

\begin{figure}[t!]
  \centering  
  \begin{subfigure}{\figamplifierW\textwidth}
    \centering
    \includegraphics[width=\textwidth]{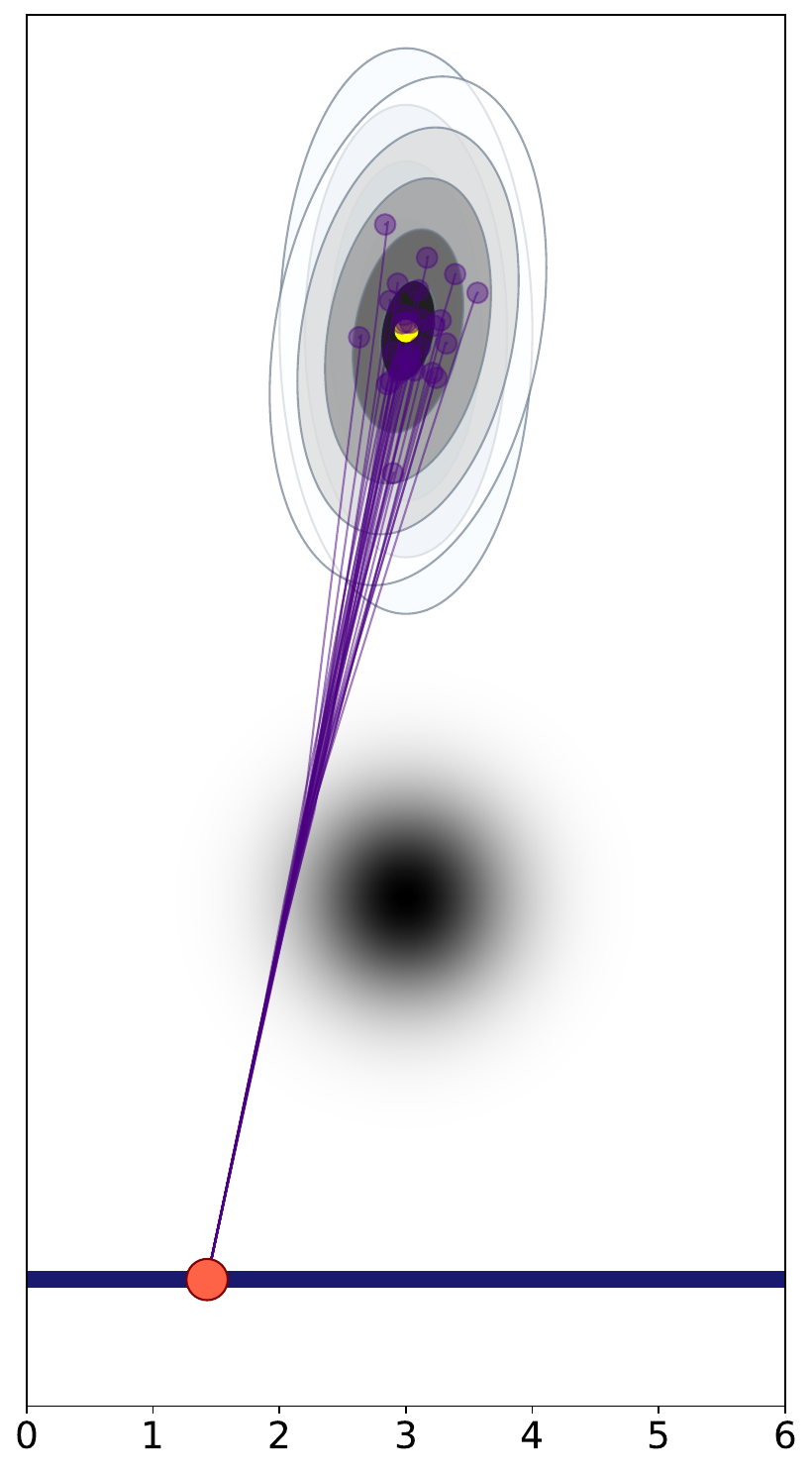}
    \caption{KL Divergence}
    \label{fig:amplifier_kl}
  \end{subfigure}
  \begin{subfigure}{\figamplifierW\textwidth}
    \centering
    \includegraphics[width=\textwidth]{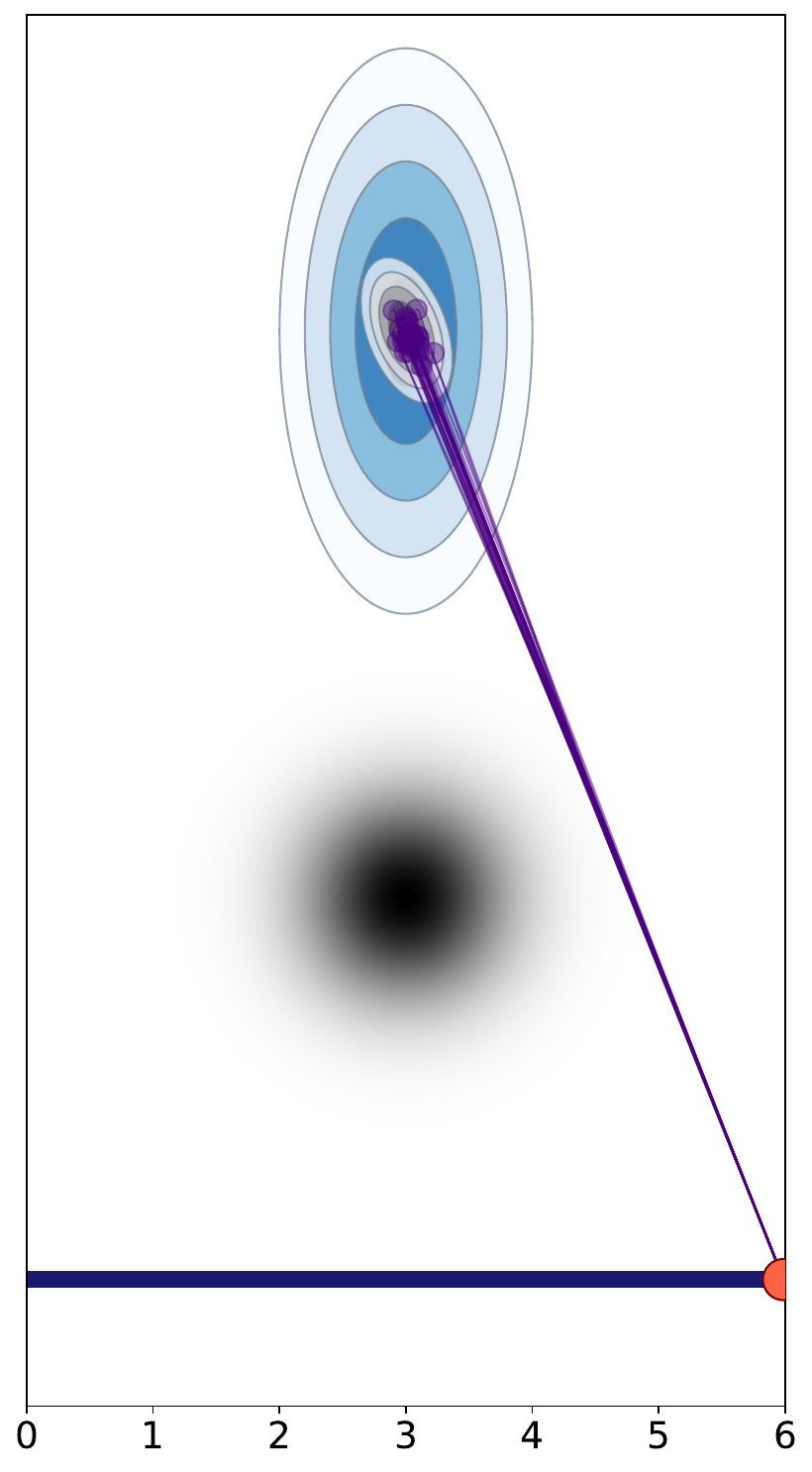}
    \caption{Cross-entropy}
    \label{fig:amplifier_ce}
  \end{subfigure}
 
  \begin{subfigure}{\figamplifierW\textwidth}
    \centering
    \includegraphics[width=\textwidth]{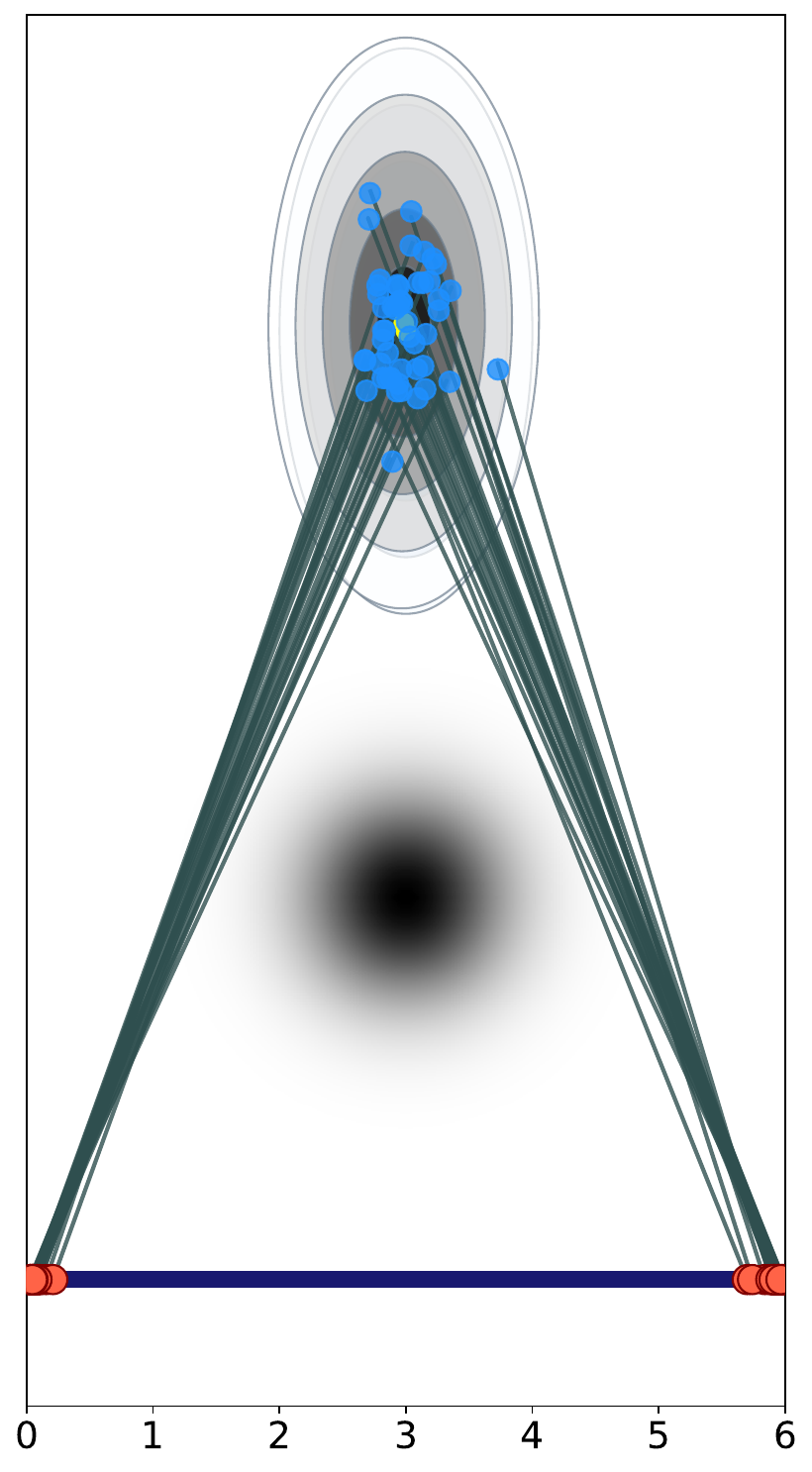}
    \caption{Log PDF of \(q(\x_T \mid \tau)\)}
    \label{fig:amplifier_logpdf}
  \end{subfigure}
  \begin{subfigure}{\figamplifierW\textwidth}
    \centering
    \includegraphics[width=\textwidth]{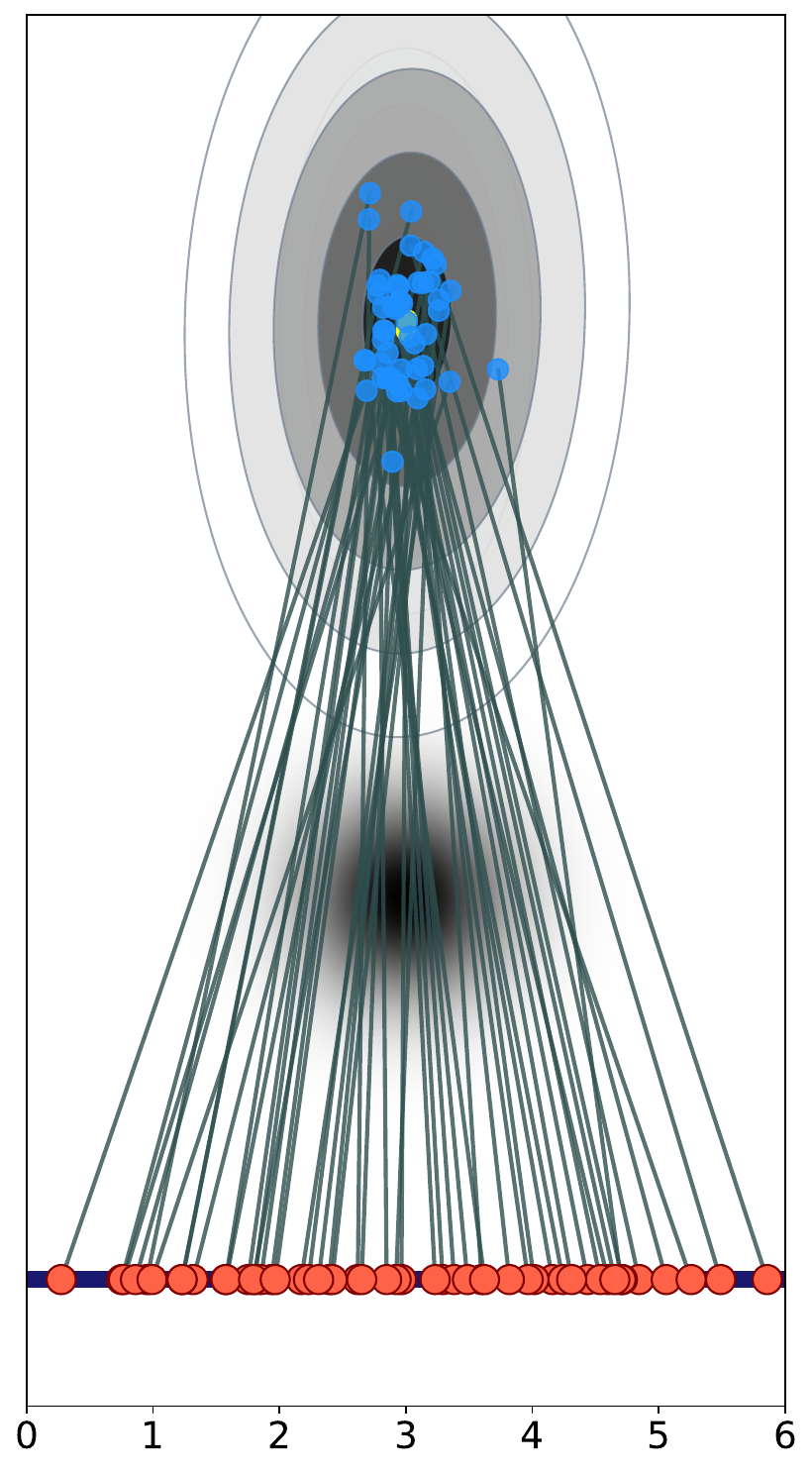}
    \caption{Deterministic Planner}
    \label{fig:amplifier_det}
  \end{subfigure}
  \caption{Comparison of different planning objectives for a 2D ball-rolling
  environment. The initial ball position (red dot) is constrained to the dark
  blue line. Blue ellipses denote the Gaussian goal distribution. Grey ellipses
  illustrate the terminal state distribution fitted from rollouts. The black
  gradient is a noise amplifier that increases dynamics noise the closer the
  ball gets to darker-colored regions. \textbf{(a)} KL divergence plans leverage
  the additional noise uncertainty from the amplifier to more closely match the
  terminal covariance of the goal distribution. Purple lines show paths from 20
  rollouts of the plan. \textbf{(b)} Cross-entropy plans avoid the noise
  amplifier and achieve a tight covariance about the goal mean. \textbf{(c)}
  Using the negative log PDF of the terminal state distribution as the cost, we
  plan to 50 samples from the goal distribution (blue dots) and avoid the noise
  amplifier. Lines indicate the planned path for the 50 different goal
  samples. \textbf{(d)} A deterministic planner ignores the noise amplifier in
  planning to the goal samples, resulting in a terminal distribution that does
  not match the goal distribution.}
  \label{fig:amplifier}
\end{figure}


We use the state and dynamics model of~\cite{vanderbei2001case}. The state space
\(\mathcal{X} \subseteq \mathbb{R}^6\) consists of the planar position
\(\bm{p}\), velocity \(\bm{v} = \dot{\bm{p}}\), and acceleration
\(\bm{a} = \ddot{\bm{p}}\) of the ball. The acceleration of the ball is computed
by \(\bm{a} = \bm{F} / m\) where \(m = 0.045kg\) is the mass of the ball in
kilograms and the force
\begin{equation}
  \bm{F} = -\mu_f ||\bm{N}|| \frac{\bm{v}}{||\bm{v}||}
\end{equation}
is the frictional force applied to the ball from the ground where \(\mu_f\) is
the coefficient of friction and \(\bm{N}\) is the normal force applied to the
ball from the ground. We assume a flat surface so that the normal force is
simply counteracting gravity, i.e. \(\bm{N} = (0, 0, mg)\) for
\(g=9.8m/s^2\). We use a friction coefficient of \(\mu_f = 0.04\). We compute
velocity and position using Euler integration from the computed acceleration
with a timestep of \(\mathrm{d}t=0.3\). We constrain the position to the blue
line shown in Fig.~\ref{fig:amplifier}.

We apply a small additive isotropic Gaussian noise
\(\bm{\omega} \sim \mathcal{N}(\bm{a} \mid \bm{0}, \bm{\alpha}\bm{I})\) to the
acceleration for \(\bm{\alpha} = 0.0001\). However, this environment has
heterogeneous noise. The black, circular gradient shown in
Fig.~\ref{fig:amplifier} is a noise amplifier defined by a Gaussian distribution
that adds an additional
\(\bm{\omega}^\prime \sim \mathcal{N}(\bm{a} \mid \bm{0},
\bm{\alpha}^\prime\bm{I})\) where \(\bm{\alpha}^\prime\) is proportional to the
Mahalanobis distance between the agent and the center of the noise
amplifier. Intuitively, the closer the ball gets to the center of the noise
amplifier, the more dynamics noise is added to the acceleration. This behavior
mimics exacerbated state uncertainty induced by the environment, e.g. a patch of
ice.

We consider a Gaussian goal distribution
\(p_{\g}(\x) = \mathcal{N}(\x \mid \mean_{\g}, \bSi_{\g})\) illustrated by the
blue ellipses in Fig.~\ref{fig:amplifier}. We again use the unscented transform
to compute the marginal terminal state distribution
\(\mathcal{N}(\x_T \mid \mean_T, \bSi_T)\). In order to compare the terminal
distributions resulting from each plan, we run a total of 500 rollouts per trial
and fit a Gaussian distribution to the points at the last timestep using maximum
likelihood estimation. Terminal distributions are visualized by grey ellipses in
Fig.~\ref{fig:amplifier} showing standard deviations. Note we estimate the
terminal distributions using rollouts insteads of the unscented transform
prediction in order to put the comparison on an even footing with the
deterministic point-based planner we will discuss shortly, which has no
associated method of uncertainty propagation.

As shown in Fig.~\ref{fig:amplifier_kl}, minimizing the I-projection of KL
divergence
\(\inlinekl{\mathcal{N}(\x_T \mid \bmu_T, \bSi_T)}{\mathcal{N}(\x_T \mid
\bmu_{\g}, \bSi_{\g})}\) results in a plan that has the ball pass close by the
noise amplifier. The terminal distribution matches the goal distribution with a
KL divergence value of 0.796. However, minimizing the I-projection of
cross-entropy
\(\inlinece{\mathcal{N}(\x_T \mid \bmu_T, \bSi_T)}{\mathcal{N}(\x_T \mid
\bmu_{\g}, \bSi_{\g})}\) results in a plan that avoids the noise amplifier and
keeps a tight covariance for the terminal state distribution as shown in
Fig.~\ref{fig:amplifier_ce}. The resulting distribution has a KL divergence
value of 2.272 with respect to the goal distribution. This higher KL divergence
value is due to cross-entropy seeking to maximize the expected (log) probability
of reaching the goal, which incentivizes a lower entropy terminal state
distribution. This is evident when we consider that
\(\inlinece{p1}{p2} = \mathcal{H}(p1) + \inlinekl{p1}{p2}\) as discussed
previously in Sec.~\ref{sec:max_ent_terminal}.

\begin{figure*}[t!]
  \centering  
  \includegraphics[width=\textwidth]{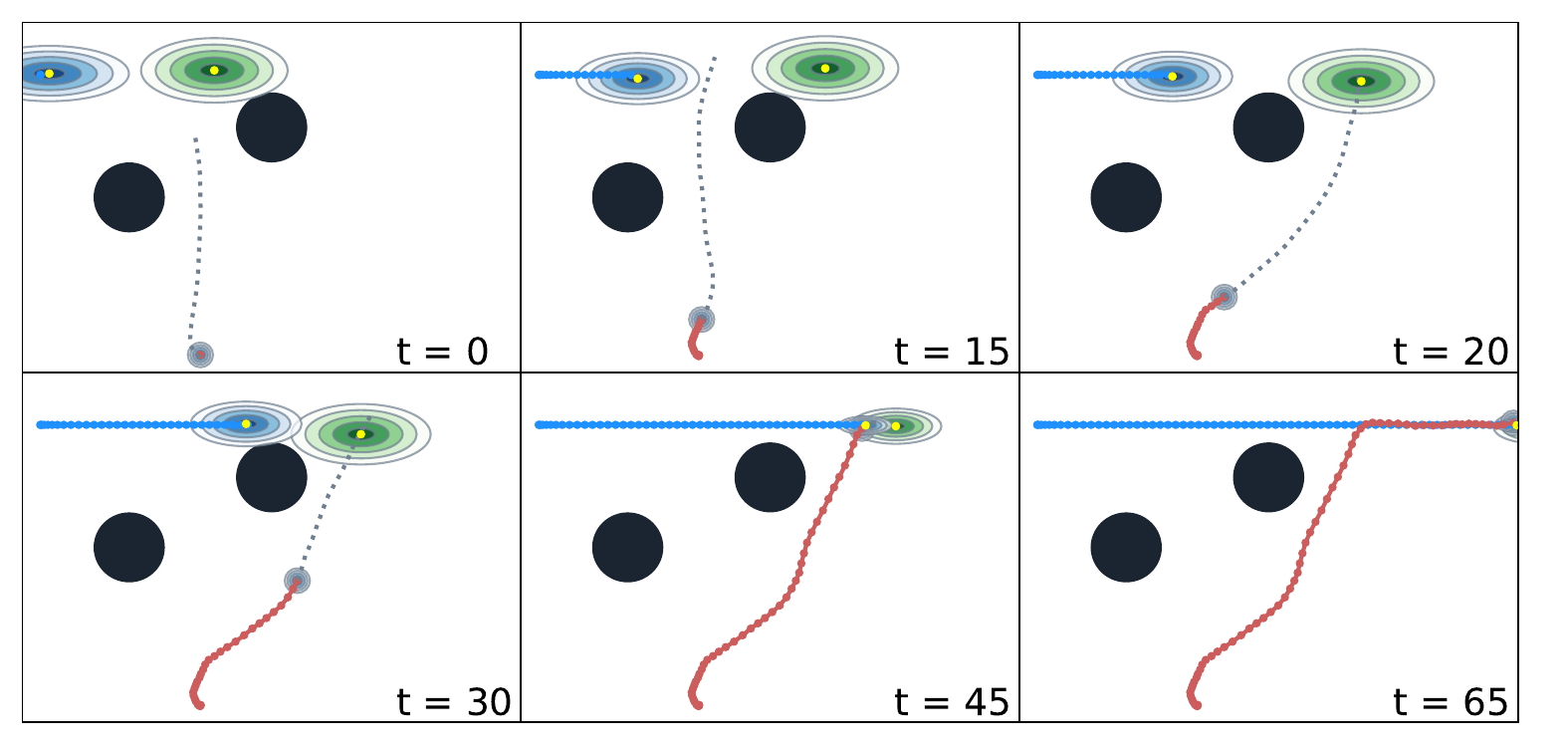}
  \caption{A selection of frames from an MPC execution of a planar navigation
  agent intercepting a moving target. The actual goal path is denoted by the
  blue line, the blue ellipses display the agent's Gaussian belief about where
  the goal is at each timestep, and the green ellipses display the agent's
  projected Gaussian belief of where the goal will be at the end of the planning
  horizon. The agent's path is denoted by the red line in each frame, and the
  gray dotted line displays its current plan. Dark spheres are obstacles to
  avoid. See Sec.~\ref{sec:moving_target} for details.}
  \label{fig:moving_target}
\end{figure*}


We further consider two point-based planners. First, we use the negative log
probability of the computed terminal state distribution
\(\mathcal{N}(\x_T \mid \mean_T, \bSi_T)\) as the cost. We sample 50 goal points
from the Gaussian goal distribution (blue dots in
Fig.~\ref{fig:amplifier_logpdf}) and generate a plan for each goal point. We
perform 10 rollouts for each of the 50 plans for a total of 500 rollouts and fit
the Gaussian distribution displayed with grey ellipses in
Fig.~\ref{fig:amplifier_logpdf}. We see that the terminal distribution matches
very closely to the goal distribution with a KL divergence of 0.770. This value
is only slightly lower than the distribution we achieved with 500 rollouts of
the single plan resulting from optimizing KL divergence. Second, we consider a
deterministic planner that plans to each of the goal samples and assumes there
is no noise in the dynamics. As shown in Fig.~\ref{fig:amplifier_det}, we get
solutions where the initial position of the ball is spread out over the range of
the start region. Some of the plans then have the ball pass directly over the
noise amplifier, resulting in rollouts that may disperse the ball far away from
its target location. We again perform 10 rollouts for each of the 50 plans and
find the terminal Gaussian distribution to have a KL divergence of 1.388, nearly
double the value of the KL divergence plan and log probability plans. It is
therefore not sufficient to deterministically plan to sample points from the
goal distribution, and we achieve better results utilizing the predicted
terminal state uncertainty. Note that with the point-based planners, we must
compute a separate plan for every point. In contrast, we computed a single plan
optimizing KL divergence that resulted in a terminal state distribution closely
matching the goal distribution.

These examples illustrate that our framework is able to advantageously leverage
sources of uncertainty in the environment to achieve a target state
distribution. We note similar results can also be achieved in environments with
homogeneous noise if the agent optimizes stochastic policies, e.g. with
reinforcement learning.

\subsection{Goal distributions enable belief updates for moving targets}
\label{sec:moving_target}

We have so far only considered examples with static goal distributions. We now
consider a more realistic scenario in which the goal distribution changes over
time and is updated based on the agent's observations. The robot's task is
to intercept a moving target while maintaining a Gaussian belief of the target's
location from noisy observations.

We use the widely utilized double integrator system~\cite{rao2001naive,
lambert2020stein} for both the agent and the moving target in a planar
environment. The agent's state space \(\mathcal{X} \subseteq \mathbb{R}^4\)
consists of a planar position \(\p \in \mathbb{R}^2\) and velocity
\(\bm{v} \in \mathbb{R}^2\) and the action space
\(\mathcal{U} \subseteq \mathbb{R}^2\) is the agent's acceleration
\(\bm{a} \in \mathbb{R}^2\). The dynamics obey \(\bm{v} = \bm{\dot{p}}\) and
\(\bm{a} = \bm{\dot{v}}\) and we compute the position and velocity through
double Euler integration of the applied acceleration (hence the name ``double
integrator'').

The agent maintains a Gaussian belief of the target's location (blue ellipses in
Fig.~\ref{fig:moving_target}) that is updated at every timestep using a Kalman
filter~\cite{kalman1960filter}. The agent also uses the Kalman filter to project
its Gaussian belief of the target's location over the planning horizon to
represent the agent's Gaussian belief of the targets future location (green
ellipses in Fig.~\ref{fig:moving_target}). Note that the target moves
deterministically in a straight line with constant velocity from left to
right. The agent uses this motion model in the Kalman updates, but assumes the
agent moves stochastically. The agent receives noisy observations of the
target's location at every timestep, where the observations get more accurate
the closer the agent gets to the target.

We use the I-projection of cross-entropy as our planning objective. We
interleave planning and control with a model predictive control (MPC)
scheme. The agent creates a plan for a small horizon, executes the first action
of the plan in the environment, acquires an observation from the environment,
and then re-plans. This procedure continues until a termination condition is
achieved.  We use \textit{model predictive path integral control (MPPI)} as our
plan update procedure. MPPI is a sampling-based solver that is widely used due
to its computation efficiency and efficacy on complex domains. See
Appendix~\ref{app:moving_target_solver} for details on the solver and MPC scheme
we use.

We see in Fig.~\ref{fig:moving_target} that initially (timesteps \(t=0\) and
\(t=15\)) the agent plans a path between the two obstacles to the projected goal
distribution. However, as the target advances in its trajectory, the agent gets
new observations that push the projected goal distribution far enough along that
the agent starts planning paths that go to the right of all obstacles (starting
at \(t=20\)). The agent thus switched the homotopy class of its planned path to
better intercept the target based on its prediction.

By timestep \(t=45\), the agent has intercepted the target. Note that the
agent's uncertainty at this point is much lower than earlier in the execution
(denoted by the smaller ellipses for the Gaussian belief in
Fig.~\ref{fig:moving_target}). This is due to the observation model we define
that enables the agent to incorporate more accurate observations into the Kalman
filter state estimation as the agent gets closer to the target. Once the agent
intercepts the target (\(t=45\)), it continues to move along in sync with the
target, as shown in the final frame of Fig.~\ref{fig:moving_target} for
\(t=65\). Note that the agent's path does not precisely follow the target but
deviates slightly. This is due to the agent executing with stochastic dynamics,
and there is some irreducible uncertainty about where the target is located due
to the alleotoric uncertainty of the agent's sensors.

This example demonstrates an application of our approach to a more realistic
setting in which the agent must update its belief of its goal online using noisy
observations. Importantly, we were able to use the output of a Kalman filter
directly as our goal representation for planning, as opposed to only using the
mean or a sample from the belief distribution. We only explored the I-projection
of cross-entropy in this simple example. However, we believe other objectives
(e.g. the M-projection) discussed in our planning framework deserve more
detailed attention, a point we disucss further in Sec.~\ref{sec:conclusion}.

\subsection{Goal distributions are a better goal representation for robot arm
reaching}
\label{sec:arm_reaching}

\begin{figure*}[t!]
  \centering
  \includegraphics[width=\textwidth]{imgs/arm_reaching.pdf}
  \caption{Examples of a 7-DOF arm reaching its end-effector to a goal pose
  mixture model distribution about an object to be grasped. Close-up views of
  the object are shown in the first frame in each row, where green and red
  spheres indicate whether a pose is reachable or not, respectively. Each row
  shows the reaching motion where time increases from left to right.}
  \label{fig:arm_reaching}
\end{figure*}


We now turn to a more complex domain than the previous experiments to
demonstrate an advantage of a distribution-based goal over a point-based
goal. We address the problem of a 7-DOF robot arm reaching to grasp an
object~\cite{conkey2019active, maeda2017active} shown in
Fig.~\ref{fig:arm_reaching}. The objective for the robot is to reach its
end-effector to a pre-grasp pose near an object such that closing the fingers of
the hand would result in the robot grasping the object.

A common heuristic for generating pre-grasp pose candidates is to select poses
that align the palm of the hand to a face of an axis-aligned bounding box of the
object~\cite{lu-isrr2017-grasp-inference, lu-ram2020-grasp-inference,
miller2003automatic}. Zero-mean Gaussian noise may also be added to each
candidate to induce further variation in the pose
candidates~\cite{lu-isrr2017-grasp-inference}. We model this explicitly as a
mixture of distributions (Sec.~\ref{sec:mixture_models}) where each component is
a distribution over \(\mathrm{SE}(3)\) poses. We represent each
\(\mathrm{SE}(3)\) pose distribution as a Gaussian distribution
(Sec.~\ref{sec:gaussian}) over the 3D position together with a Bingham
distribution (Sec.~\ref{sec:bingham}) over the \(\mathrm{SO}(3)\) orientations
of the end-effector. We refer to this distribution as a \textit{pose mixture
model}. The distribution is defined relative to the object, and therefore as the
object is placed in different locations on the table shown in
Fig.~\ref{fig:arm_reaching}, the goal distribution is transformed to the
object's reference frame. This distribution is similar to the Gaussian mixture
model used to encode goals fit from data for object-reaching in a learning from
demonstration setting in~\cite{conkey2019active}. However, our pose mixture
model is more correct in the sense we properly model the distribution of
orientations as a Gaussian distribution in \(\mathrm{SO}(3)\).

We use the simulated 7-DOF KUKA iiwa arm shown in Fig.~\ref{fig:arm_reaching}
with state space \(\mathcal{X} \subset \mathbb{R}^7 \times \mathrm{SE}(3)\)
consisting of joint positions \(\bm{q} \in \mathbb{R}^7\) in radians together
with the pose of the end-effector \(\bm{p} \in \mathrm{SE}(3)\). We use
quaternions to represent the end-effector orientation which naturally pairs with
our use of the Bingham distribution as described in Sec.~\ref{sec:bingham}. The
robot's action space \(\mathcal{U} \subset \mathbb{R}^7\) consists of changes in
joint angles \(\Delta \bm{q} \in \mathbb{R}^7\). The state trasition dynamics
are governed by
\begin{align}
  \bm{q}_{t+1} &= \bm{q}_t + \Delta\bm{q}_t \label{eq:arm_transition}\\
  \bm{p}_{t+1} &= FK(\bm{q}_{t+1}) \label{eq:arm_fk}
\end{align}
where \(FK(\cdot)\) is the robot's forward kinematics. An Allegro hand is
mounted on the robot arm. The robot's objective is to orient the palm link of
the hand to a pre-grasp pose. We use a fixed grasp pre-shape joint configuration
for the fingers. We use the ``sugar box'' object from the YCB dataset in our
experiments, shown in Fig.~\ref{fig:arm_reaching}. We defined a goal pose
mixture model with 6 components following the heuristic
from~\cite{lu-isrr2017-grasp-inference}.

This problem is traditionally solved by selecting a particular point-based goal
to create a motion plan for~\cite{lu-isrr2017-grasp-inference}. For example, we
can generate a sample from the pose mixture model we have defined. However, many
of the samples will be unreachable based on the kinematics of the robot arm. A
target pose relative to the object may be reachable when the object is placed in
some poses on the table, and unreachable in others. We quantify this for our
environment by spawning the object in 100 random locations on the table. We then
generate 100 samples from the pose mixture model distribution and use the
robot's inverse kinematics (IK) to compute joint configurations that would
enable the robot to reach to each sample. A sample is considered reachable if an
IK solution can be found for that sample, and unreachable otherwise. We use
Drake~\cite{drake} to compute IK solutions, where we make up to 10 attempts to
find a solution, each time seeding the solver with a different joint
configuration sampled uniformly within the robot's joint limits.

We find that for the 100 random poses of the object, a mean value of 39.53 of
the 100 generated samples are unreachable with a standard deviation of 16.97. We
additionally quantify the number of unreachable components per object pose by
generating 100 samples from each component. We determine a component to be
unreachable if at least 50 of the 100 component samples are unreachable. We find
that on average 2.38 of the 6 components are unreachable with a standard
deviation of 1.08 (see Appendix~\ref{app:arm_reaching} for more visualizations
of reachability). These numbers suggest that simply generating goal samples for
a point-based planner will frequently result in goal points that are not
feasible (i.e. unreachable). The point-based planner will therefore need to
first validate the point is reachable by computing inverse kinematics prior to
planning.

Using our probabilistic planning approach and the pose mixture model as the goal
distribution, we are able to generate reaching plans without having to check for
reachability with inverse kinematics. A selection of executions are shown in
Fig.~\ref{fig:arm_reaching}. We define a pose distribution for the end-effector
with a fixed covariance which we transform over the planning horizon to mimic
uncertainty propagation. We make this simplification since the dynamics noise on
industrial robot arms like the KUKA iiwa we utilize is typically negligible. We
note that proper uncertainty propagation can be performed using unscented
orientation propagation~\cite{gilitschenski2015unscented}. We use the
approximate cross-entropy loss described in Sec.~\ref{sec:approximate_loss}
since there is no closed form solution between a pose distribution and pose
mixture model.

We ran our approach 10 times on all 100 object poses, each time with a different
seed for random number generators, to quantify how often the plans reach to a
reachable component, where component reachability is determined as described
above. Our approach reaches to a reachable component with a mean percentage of
96.4\% and standard deviation of 2.059\% over the 10 trials. The instances where
our approach does not plan to a reachable component are due to the solver
getting stuck in local optima. This effect could be mitigated by a more advanced
solver, e.g. using Stein variational methods~\cite{lambert2020stein}.

In summary, a pose mixture model is a common heuristic representation for target
pose candidates in reaching to grasp an object. Instead of planning to
point-based samples, our approach is capable of using the pose mixture model
distribution directly as the goal representation. In contrast to the point-based
goal representation, we do not have to compute inverse kinematics to first
determine if the target pose is reachable or not.

\subsection{Goal distributions are a better goal representation for planning
with dynamic skills}
\label{sec:skill_planning}

\begin{figure*}[t!]
  \centering
  \begin{subfigure}{0.594\textwidth}
    \centering
    \includegraphics[width=\textwidth]{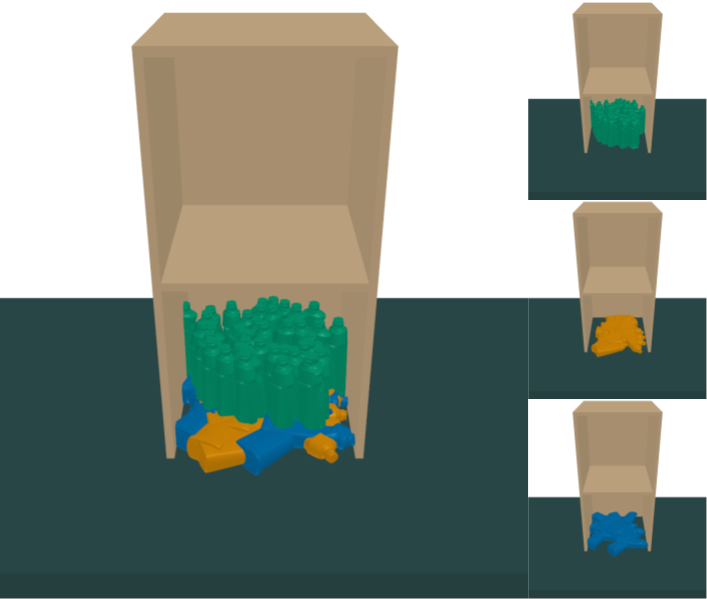}
    \caption{Goal GMM for ``object in cabinet''}
    \label{fig:shelf_goal_gmm}
  \end{subfigure}
  \begin{subfigure}{0.4\textwidth}
    \centering
    \includegraphics[width=\textwidth]{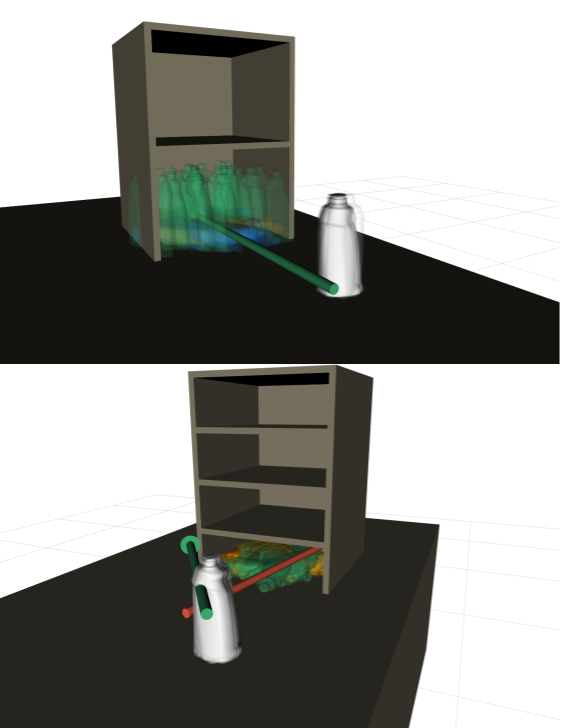}
    \caption{Push plans for shelf environment}
    \label{fig:shelf_push_plans}
  \end{subfigure}
  \caption{Goal distribution and push plans generated by our planner for the
  task of relocating an object (cleaner bottle from YCB~\cite{calli2015ycb}
  dataset) to a nearby cabinet. \textbf{(a)} Goal GMM for the goal of ``object
  in cabinet'' capturing the semantic ambiguity of the goal such that the object
  can be placed in any pose within the cabinet, either upright (green) or on one
  of its flat supporting side surfaces (orange and blue). \textbf{(b)} One-step
  (upper) and two-step (lower) push-plans generated by our planner. The one-step
  plan (green arrow) is a low-push that slides the object into the cabinet,
  where the shelf is high enough that the upright object fits under the
  shelf. The two-step plan first executes a high-push (green arrow) to topple
  the object onto its side, and then executes a low-push (red arrow) to relocate
  the object into the cabinet. The topple-then-relocate approach is necessary
  since the lower shelves prevent the upright object from being directly
  relocated to the cabinet with a single push.}
  \label{fig:skill_planning}
\end{figure*}


All of the applications we have looked at so far have had state dynamics that
were well-captured by analytic functions. In our final application, we
investigate planning to goal distributions in the domain of manipulation skill
planning where a learned dynamics function is necessary for good planning
performance. We show how a learned dynamics function can be utilized in our
framework and demonstrate how goal distributions naturally encode the
uncertainty inherent to goals for manipulation tasks.

Manipulation skill planning requires a robot to determine how to sequence
primitive skills (e.g. push, pick-place) from a set of skills \(\mathcal{S}\) in
order to manipulate its environment into a desired configuration. We will
consider the concrete example shown in Fig.~\ref{fig:skill_planning} in which
the robot is tasked to relocate an object from a kitchen counter to a nearby
shelf. Each skill has an associated set of parameters \(\Theta_k\) that
determine how the skill is executed. For example, a push skill might be
parameterized by start and end poses of the end-effector. Skill planning thus
requires the robot to find a sequence of skills
\(S_1, \dots, S_m \in \mathcal{S}^m\) as well as the associated parameters for
each skill \(\bm{\theta}_1, \dots, \bm{\theta}_m\) such that the desired outcome
is achieved when the sequence is executed.

Skill planning is traditionally formulated with point-based
goals\cite{garrett2021integrated, simeonov2020corl}. However, goals for
manipulation tasks often have inherent uncertainty due to semantic ambiguity and
partial observability which is not captured by a point-based goal
representation. For example, the goal ``object on shelf'' is satisfied
irrespective of the particular pose the object is in on the shelf. If there are
multiple shelves in the scene, the goal may also be satisfied irrespective of
the particular shelf the object is on. The robot may also have uncertainty about
where the object is located, e.g. due to occlusion by other objects in the
scene. These scenarios are nicely encoded by mixture models as discussed in
Sec.~\ref{sec:mixture_models}, e.g. Gaussian mixture models. Goal distributions
are therefore an excellent goal representation to accommodate the various
sources of uncertainty a robot faces in manipulation tasks.

In order to address skill planning in our probabilistic planning framework, we
require a dynamics function to propagate the state uncertainty. Traditionally,
skill planning methods utilize skills that obey quasi-static
dynamics~\cite{simeonov2020corl}. This assumption allows for dynamics models
that assume the skill is executed perfectly and that the outcome of the skill
can be easily predicted as a single point in state space using a learned skill
effect model~\cite{liang2022search}. However, more dynamic skills such as
toppling, shoving, and tossing have aleatoric uncertainty due to the complex
contact dynamics between interacting objects. This makes accurate skill outcome
prediction infeasible for point-based predictors. For example, it is intractable
to predict precisely where a dropped bottle will come to rest. With this
intuition in mind, we learn probabilistic skill effect models, a probabilistic
variant of the skill effect models (SEMs) described in~\cite{liang2022search}
that predict the outcome of a skill. We consider a model for a skill
\(S_k \in \mathcal{S}\) that predicts
\(\bm{\omega}_{t+1} = f_k(\bm{\omega}_t, \bm{\theta}, \bm{\xi}_k)\), where
\(\bm{\omega}_t \in \Omega_\mathcal{X}\) encodes a probability distribution over
states at time \(t\), \(\bm{\omega}_{t+1} \in \Omega^\prime_\mathcal{X}\)
encodes a probability distribution over states at time \(t+1\),
\(\bm{\theta} \in \Theta_k\) are skill parameters, and \(\bm{\xi}_k \in \Xi_k\)
are function parameters (e.g., neural network weights). Intuitively,
probabilistic SEMs predict how a state distribution will transform after
applying skill \(S_k \in \mathcal{S}\) using parameters
\(\bm{\theta}_k \in \Theta_k\). Examples of uncertainty parameters
\(\Omega_\mathcal{X}\) include sufficient statistics of a distribution (e.g.,
mean and covariance of a Gaussian distribution) or Dirac measures over the state
space.

Our probabilistic SEMs predict the state uncertainty induced by a single skill
execution. In order to enable multi-step skill planning, we require a method to
chain our predictions to predict the terminal state distribution, as described
in Sec.~\ref{sec:practical_algorithm}. While the unscented transform was
sufficient for the previous applications we investigated, we require a more
general method of uncertainty propgation for SEMs since the predictions are not
restricted to be Gaussian. We employ a simple strategy that is widely applicable
to arbitrary distribution models based on Markov chain Monte Carlo (MCMC)
sampling. We consider SEMs that have an input space \(\Omega_\mathcal{X}\) of
Dirac-delta measures (i.e., point-based inputs), an arbitrary output space
\(\Omega_\mathcal{X}^\prime\), and we define
\(g:\Omega_\mathcal{X}^\prime \rightarrow \Omega_\mathcal{X}\) to be a function
that generates samples \(\bm{x}_t \sim \bm{\omega}_t\) from the distribution
\(\bm{\omega}_t \in \Omega_\mathcal{X}^\prime\). For each step in the plan, we
sample a state \(\bm{x}_t \sim \bm{\omega}_t\), input that point to the SEM to
predict \(\bm{\omega}_{t+1} = f_k(\bm{x}_t, \bm{\theta}_t, \bm{\xi}_k)\), and
continue in this fashion chaining the model predictions until we acquire a
predicted terminal state \(\bm{x}_T\) for the sample sequence. We then repeat
this procedure for an arbitrary number of samples from the initial state
distribution \(\bm{\omega}_0\) to collect a set of predictions for the terminal
state \(\{\bm{x}_T^i\}\). Finally, we compute the terminal distribution
\(q(\bm{x}_T \mid \bm{\omega}_0, \bm{\theta}_1, \dots, \bm{\theta}_{T-1})\) by
fitting the distribution to the sample set \(\{\bm{x}_T^i\}\) in a maximum
likelihood fashion.

We now have all of the pieces required to apply our probabilistic planning
framework to skill planning under uncertainty with dynamic skills. To
demonstrate the applicability of our method in this context, we consider the
task of relocating a bottle sitting on a table to a nearby cabinet with shelves,
shown in Fig.~\ref{fig:skill_planning}. We equip the robot with a simple pushing
skill parameterized by the start and end poses of the robot's end-effector in
the coordinate frame of the object. Importantly, depending on the height at
which the robot pushes the bottle, the bottle may remain upright and slide or it
may topple over. The terminal pose of the bottle is therefore difficult to
predict with a point-based model since it may rotate and bounce once it is
pushed (see Appendix~\ref{app:skill_planning} for more details).

For our application, we use mixture density networks
(MDNs)~\cite{bishop1994mixture}: given an object pose as input, we predict the
parameters of a Gaussian mixture model (GMM) over the resulting change in object
pose. We found GMMs to best model scenarios with bifurcating dynamics,
e.g. pushing the bottle at half its height might result with the object
remaining upright or toppling over depending on subtle differences between skill
executions (see Appendix~\ref{app:skill_planning} for model comparisons). We
trained the model with 10,000 instances of the robot pushing the bottle with an
engineered, motion-planned behavior. We used the NVIDIA Isaac Gym simulator to
parallelize data collection. Please see Appendix~\ref{app:skill_planning} for
further details on the specific model architecture, data collection, and
training procedure we utilized.

We use Gaussian mixture models (GMMs) to encode goal distributions for this
environment, as illustrated in Fig.~\ref{fig:shelf_goal_gmm}. GMMs adequately
capture the notion that the object is in the cabinet irrespective of the
particular pose it comes to rest in inside the cabinet, including standing
upright or toppled over on its side. We generate goal distributions by
parametrically specifying a nominal GMM and generating samples from it. We then
use the collision geometry of the bottle, table, and shelf and reject samples
where the bottle is in collision with the table or shelf. We re-fit the GMM to
the samples that are free of collision. We thereby achieve a goal distribution
that has collision-free samples and satisfies the goal of ``object in
cabinet''. This procedure generalizes also to different cabinets, e.g. where
shelves are differently spaced as shown in Fig.~\ref{fig:shelf_push_plans}.

A single-step plan for achieving the goal distribution is shown in
Fig.~\ref{fig:shelf_push_plans}. In this case the shelf is high enough that the
robot can execute a push lower on the bottle and slide the object into the
cabinet. When the evenly-spaced shelves are more densely populated, as shown in
Fig.~\ref{fig:shelf_push_plans}, it becomes infeasible to simply slide the
upright bottle into the cabinet because it will collide with the shelf. Our
multi-step skill planning approach finds a suitable plan that first executes a
high-push to topple the bottle over onto its side, and then executes a low-push
to slide the toppled object into the cabinet. We highlight the fact that the
robot does not need to target a specific terminal object pose, nor does it have
to precisely predict the outcome of executing the push skill. The robot
leverages the goal distribution that encodes the semantic ambiguity of the goal,
together with the propagated uncertainty of applying the its skills, to devise a
coarse manipulation plan that achieves the desired outcome.

Our preliminary results suggest goal distributions and probabilistic skill
effect predictions are a promising way forward for manipulation planning with
more dynamic skills, and we are excited to see this idea taken further in future
work. We discuss possible extensions to this application next in
Sec.~\ref{sec:conclusion}.


\section{Discussion and Conclusion}
\label{sec:conclusion}

In this paper, we have argued that goal distributions are a more suitable goal
representation than point-based goals for many problems in robotics. Goal
distributions not only subsume traditional point-based and set-based goals but
enable varied models of uncertainty the agent might have about its goal. We
derived planning to goal distributions as an instance of planning as inference,
thereby connecting our approach to the rich literature on planning as inference
and enabling the use of a variety of solvers for planning to goal
distributions. We additionally derived several cost reductions of our
probabilistic planning formulation to common planning objectives in the
literature. Our experiments showcased the flexibility of probability
distributions as a goal representation, and the ease with which we can
accommodate different models of goal uncertainty in our framework.

We believe there are many exciting avenues for future research in planning to
goal distributions. One interesting direction is incorporating learning into our
probabilistic planning framework. We see two key areas that will benefit from
learning. First, learning conditional generative models of goal distributions
will provide a powerful and flexible goal representation for more complex
environments and behaviors. This is also advocated for
in~\cite{nasiriany2021disco}, albeit for goal-conditioned policies in
reinforcement learning. For example, learning mixture density
networks~\cite{bishop1994mixture} conditioned on a representation of the current
environment configuration would enable learning scene-dependent multi-modal goal
distributions. This could be paired with a multi-modal distribution in the
planner as in Stein-variational methods~\cite{lambert2020stein} to enable
parallelized planning to multiple goal regions simultaneously.

A second area we believe learning will benefit our probabilistic planning
framework is in propagating the robot's state uncertainty. The state spaces and
dynamics functions utilized in our experiments were simple for demonstrative
purposes, but more complex environments may require more advanced uncertainty
propagation techniques to estimate the terminal state distribution well. For
example, if the state space itself is a learned latent representation space,
\textit{normalizing flows}~\cite{kobyzev2020normalizing} may be an interesting
technique to estimate how the latent state distribution transforms over the
planning horizon.

In this paper, we only investigated cross-entropy and KL divergence as
information-theoretic losses in our optimization. As we discussed in
Sec.~\ref{sec:related_work}, other losses are possible such as
\(f\)-divergence~\cite{belousov2017fdivergence, ghasemipour2019divergence,
ke2020imitation} and Tsallis divergence~\cite{wang2021variational}. Given the
use of these broader classes of divergence in imitation
learning~\cite{ghasemipour2019divergence, ke2020imitation}, reinforcement
learning~\cite{belousov2017fdivergence}, and stochastic optimal
control~\cite{wang2021variational}, we believe there are further opportunities
in these related disciplines to utilize goal distributions beyond the planning
framework we presented in this paper. We also foresee the utility of goal
distributions in deterministic planning with uncertain goals, e.g. having
Gaussian goal uncertainty and using Mahalanobis as a heuristic bias in
sampling-based planning algorithms like RRT~\cite{lavalle1998rapidly}.

Our experiments in Sec.~\ref{sec:dubins} demonstrated that minimizing the
M-projection of cross-entropy is often necessary when we wish to model goals as
distributions with finite support. We discussed in Sec.~\ref{sec:projections}
how minimizing the M-projection of KL divergence and cross-entropy are
equivalent for our optimizations since the entropy of the goal distribution is
constant with resepect to the decision variables. However, this is not always
the case. Our experiments in Sec.~\ref{sec:moving_target} required the agent to
maintain a belief of a moving target's location which was updated at every
timestep based on observations from a noisy sensor. The goal distribution was
therefore dependent on the agents observations, which the agent has some control
over. That is, the agent's action determines in part what it will observe next,
and the observation impacts what the goal distribution will be at the next
timestep. Optimizing the M-projection in this case could therefore encourage the
agent to reduce uncertainty about its goal and lead to exploratory
behavior. However, the agent requires a model to predict what it will observe
over its planning horizon, e.g. a maximum likelihood estimate in a POMDP
setting~\cite{platt2010belief}. This scenario is beyond the scope of the current
paper but we believe this is an exciting application of our framework that
deserves further investigation.

We showed in our results in Sec.~\ref{sec:amplifier} that our planning approach
could leverage a source of state uncertainty in the environment to more
accurately match a goal state distribution. By embracing uncertainty both in the
goal representation and how the agent predicts its state to transform over the
planning horizon, our approach opens a wider array of behaviors that would
typically be avoided by deterministic plans to point-based goals. For example,
driving on mixed terrain~\cite{dahlkamp2006self} could be made more robust by
modeling how the robot will behave driving over treacherous terrain like ice and
mud. We believe leveraging sources of uncertainty in a controlled manner will
open up more diverse robot behaviors. We looked at a simple example in the
context of manipulation skill planning in Sec.~\ref{sec:skill_planning} where
predicting the distribution of poses an object might settle in after a
manipulation enables utilizing more dynamic skills. We are excited to see this
idea taken even further. For example, predicting the distribution of object
poses resulting from a tossing robot~\cite{zeng2019tossingbot} would enable
planning to target locations outside the robot's reachable workspace.

We believe goal distributions are a versatile and expressive goal representation
for robots operating under uncertainty. The probabilistic planning framework we
have presented in this article easily accommodates different goal distributions
and probabilistic planning objectives. We foresee these techniques being
particularly applicable to real-world robotics problems where state and goal
uncertainty are inherent. We are excited by the prospect of embracing goal
uncertainty explicitly in planning and control, especially when considering
state or observation uncertainty. We anticipate many fruitful avenues of further
research beyond what we have suggested here.

\section{Acknowledgement}

This work is supported by DARPA under grant \mbox{N66001-19-2-4035} and by NSF
Award 2024778.


\bibliographystyle{plainnat}
{\footnotesize
\bibliography{references}
}

\clearpage
\newpage
\appendices
\onecolumn

\section{Variational Inference for Planning as Inference}
\label{app:variational_inference}

We described at a high-level in Sec.~\ref{sec:planning_as_inference} how
variational inference techniques are often used to solve planning as inference
problems. We now provide a more detailed derivation for the optimization
objective in Eq.~\ref{eq:pai_min_neg_elbo}. This derivation is similar to that
provided in~\cite{lambert2020stein} (cf. Appendix D in~\cite{lambert2020stein}).

\begin{claim}
  Consider the distribution of optimal trajectories \(p(\tau \mid \O_\tau)\) and
  a proposal distribution \(q(\tau) \in \mathcal{Q}\) from a family of
  distributions~\(\mathcal{Q}\). The objective 
  \begin{equation}
    q^* = \argmin_{q\in \mathcal{Q}} \kl{q(\tau)}{p(\tau \mid \O_\tau)}
  \end{equation}
  is equivalently solved by
  \begin{equation}
    q^* = \argmin_{q\in \mathcal{Q}} -\E_q \left[ \log p(\mathcal{O}_\tau \mid \tau) \right]
    + \kl{q(\tau)}{p(\tau)}
  \end{equation}
\end{claim}

\begin{proof}
  We have
  \begin{align}
    q^* &= \argmin_{q\in \mathcal{Q}} \kl{q(\tau)}{p(\tau \mid \O_\tau)} \\
        &= \argmin_{q\in \mathcal{Q}} \int q(\tau) \log\frac{q(\tau)}{p(\tau \mid
          \O_\tau)} \mathrm{d}\tau \\
        &= \argmin_{q\in \mathcal{Q}} \int q(\tau) \log\frac{q(\tau)}{p(\O_\tau
          \mid \tau) p(\tau)} \mathrm{d}\tau \\
        &= \argmin_{q\in \mathcal{Q}} \int q(\tau) \left[ \log q(\tau) - \log p(\O_\tau
          \mid \tau) - \log p(\tau) \right] \mathrm{d}\tau \\
        &= \argmin_{q\in \mathcal{Q}} - \int q(\tau) \log p(\O_\tau \mid \tau)
          \mathrm{d}\tau + \int q(\tau) \left[ \log q(\tau) - \log p(\tau)
          \right] \mathrm{d}\tau \\
        &= \argmin_{q\in \mathcal{Q}} - \int q(\tau) \log p(\O_\tau \mid \tau)
          \mathrm{d}\tau + \int q(\tau) \log \frac{q(\tau)}{p(\tau)} \mathrm{d}\tau \\
        &= \argmin_{q\in \mathcal{Q}} -\mathbb{E}_q \left[\log p(\O_\tau \mid
          \tau) \right] + \kl{q(\tau)}{p(\tau)} \label{proof:vi_result}
  \end{align}
    
\end{proof}

This establishes Eq.~\ref{eq:pai_min_neg_elbo}
in~\ref{sec:planning_as_inference}. Note the KL term in
Eq.~\ref{proof:vi_result} regularizes the trajectory distribution \(q(\tau)\) to
a prior distribution \(p(\tau)\). As such, we denote the prior distribution by
\(p_0(\tau)\) in Eq.~\ref{eq:pai_min_neg_elbo}. This objective is equivalent to
maximizing the \textit{evidence lower bound (ELBO)} expressed by
\begin{equation}
  \label{eq:pai_max_elbo}
  q^* = \argmax_{q\in \mathcal{Q}} \E_q \left[ \log p(\mathcal{O}_\tau \mid \tau) \right]
  - \kl{q(\tau)}{p_0(\tau)}
\end{equation}
which is a common variational objective in machine learning more generally.


\section{Alternative Derivation for Goal Distributions in Planning as Inference}
\label{app:pai_alternative_derivation}

We now present an alternative derivation for our results in
Sec.~\ref{sec:distribution_pai}. We re-derive the planning as inference and
stochastic optimality duality from~\cite{rawlik2012stochastic} taking special
care to introduce the optimal goal distribution for the terminal cost to fit our
needs.

In planning as inference we wish to minimize the following objective
\begin{equation}
  q^* = \argmin_q \kl{q(\tau)}{p(\tau \mid \O=1)}
\end{equation}
where the variational \(q\) has the form:
\begin{equation}
  \label{eq:variational-q-form}
  q(\tau) = q(\x_0)\prod_{t=0}^{T-1}q(\x_{t+1}|\x_{t},\u_{t})\pi_q(\u_t|\x_t)
\end{equation}
and the optimal trajectory distribution has the form
\begin{equation}
  \label{eq:opt-traj-form}
  p(\tau \mid \O=1) \propto p(\tau,\O=1) = p(\x_0)p(\O_{T}=1|\x_T)\prod_{t=0}^{T-1}p(\O=1|\x_t,\u_t)p(\x_{t+1}|\x_t, \u_t)\pi_0(\u_t|\x_t)
\end{equation}
We now plug the trajectory distribution definitions into the KL objective and
expanding terms.
\begin{align}
\kl{q(\tau)}{p(\tau \mid \O=1)} = \E_{q(\tau)}\left[\ln \frac{q(\x_0)\prod_{t=0}^{T-1}q(\x_{t+1}|\x_{t},\u_{t})\pi_q(\u_t|\x_t)}{p(\x_0)p(\O=1|\x_T)\prod_{t=0}^{T-1}p(\O=1|\x_t,\u_t)p(\x_{t+1}|\x_t, \u_t)\pi_0(\u_t|\x_t)} \right] \\
  = \E_{q(\tau)}\left[ \ln \left [ q(\x_0)\prod_{t=0}^{T-1}q(\x_{t+1}|\x_{t},\u_{t})\pi_q(\u_t|\x_t)\right] - \ln \left[ p(\x_0)p(\O=1|\x_T)\prod_{t=0}^{T-1}p(\O=1|\x_t,\u_t)p(\x_{t+1}|\x_t, \u_t)\pi_0(\x_t,\u_t) \right] \right] \\
  = \E_{q(\tau)} [ \ln q(\x_0) - \ln p(\x_0) + \sum_{t=0}^{T-1}\ln q(\x_{t+1}|\x_{t},\u_{t})   - \sum_{t=0}^{T-1} \ln p(\x_{t+1}|\x_t, \u_t) \\
  + \sum_{t=0}^{T-1}\ln\pi_q(\u_t|\x_t) - \ln p(\O=1|\x_T)  -\sum_{t=0}^{T-1} \ln p(\O=1|\x_t,\u_t) - \sum_{t=0}^{T-1} \ln \pi_0(\u_t|\x_t) ]
\end{align}
We make the common assumption~\cite{levine2018reinforcement} that we have the
correct estimate of the initial distribution \(p(\x_0) = q(\x_0)\) from our
state estimation process as well as the correct dynamics model
\(q(\x_{t+1}|\x_t, \u_t) = p(\x_{t+1}|\x_t, \u_t)\). Then the objective
simplifies to
\begin{align}
\kl{q(\tau)}{p(\tau \mid \O=1)} = \E_{q(\tau)} \left[ \sum_{t=0}^{T-1}\ln\pi_q(\u_t|\x_t) - \ln p(\O=1|\x_T)  -\sum_{t=0}^{T-1} \ln p(\O=1|\x_t,\u_t) - \sum_{t=0}^{T-1} \ln \pi_0(\u_t|\x_t) \right ]
\end{align}
If we further assume that the policy is deterministic
\(\pi_q(\u_t|\x_t) = \delta_{\u_t = \phi(\x_t)}\), we are able to integrate the
term out:
\begin{align}
\kl{q(\tau)}{p(\tau \mid \O=1)} = \E_{q(\tau)} \left[ - \ln p(\O=1|\x_T)  -\sum_{t=0}^{T-1} \ln p(\O=1|\x_t,\u_t) - \sum_{t=0}^{T-1} \ln \pi_0(\u_t|\x_t) \right ]
\end{align}
We now assert our optimality distributions, namely that the terminal state
reaches the estimated goal, \(p(\O=1|\x_t) = p_{\g}(\x_T)\) and our non-terminal
optimality conditions are exponentiated negative cost (i.e. reward),
\(p(\O=1|\x_t,\u_t) = \exp\{-\alpha c_t(\x_t, \u_t)\}\). We get the objective of
\begin{align}
\kl{q(\tau)}{p(\tau \mid \O=1)} = \E_{q(\tau)} \left[ - \ln p_{\g}(\x_T)  + \sum_{t=0}^{T-1}-\alpha c_t(\x_t, \u_t) - \sum_{t=0}^{T-1} \ln \pi_0(\u_t|\x_t) \right ]
\end{align}
Which shows us that we wish to minimize the sum of the expected negative log
likelihood of reaching the goal, the expected running costs (scaled by
\(\alpha\)) as well as a term penalizing low entropy in the prior
policy. However, if we assume a uniform prior policy
\(\pi_0(\u_t|\x_t) = \mathbb{U}(\u_t\mid\mathcal{U})\) then we obtain the simpler form
\begin{align}
  \argmin_{\pi} \kl{q(\tau)}{p(\tau \mid \O=1)} = \argmin_{\pi} \E_{q(\tau)} \left[ - \ln p_{\g}(\x_T)  + \sum_{t=0}^{T-1}-\alpha c_t(\x_t, \u_t)\right ]
\end{align}

Which is equivalent to maximizing the expected probability of reaching the goal,
times the Boltzman distribution over costs:
\begin{align}
  \min \E_{q(\tau)} \left[ - \ln p_{\g}(\x_T)  + \sum_{t=0}^{T-1}-\alpha c_t(\x_t, \u_t)\right ] \equiv \max \E_{q(\tau)} \left[ p_{\g}(\x_T) \prod_{t=0}^{T-1}\exp\{-\alpha c_t(\x_t, \u_t)\}\right ]
\end{align}


Let \(\lambda = \frac{1}{\alpha}\), then
\begin{align}
  \pi* &= \argmin_{\pi} \E_{q(\tau)} \left[ - \ln p_{\g}(\x_T)  + \sum_{t=0}^{T-1}-\alpha c_t(\x_t, \u_t)\right ] \\
    &= \argmin_{\pi} \E_{q(\tau)} \left[ - \ln p_{\g}(\x_T)  + \sum_{t=0}^{T-1}-\alpha c_t(\x_t, \u_t)\right ] \cdot \lambda \\
  &= \argmin_{\pi} \E_{q(\tau)} \left[ - \lambda \ln p_{\g}(\x_T)  + \sum_{t=0}^{T-1}-c_t(\x_t, \u_t)\right ]
\end{align}
which we can interpret as a Lagrange multiplier on the terminal cost term,
enforcing an inequality constraint that
\(E_{q(\tau)} [ p_g(\x_T) ] > (1- \epsilon)\) and if this is not met we can
update the dual parameter \(\lambda\) and re-solve the unconstrained
optimization problem.


\section{Arm Reaching -- Additional Details}
\label{app:arm_reaching}

\begin{figure*}[t]
  \centering
  \includegraphics[width=\textwidth]{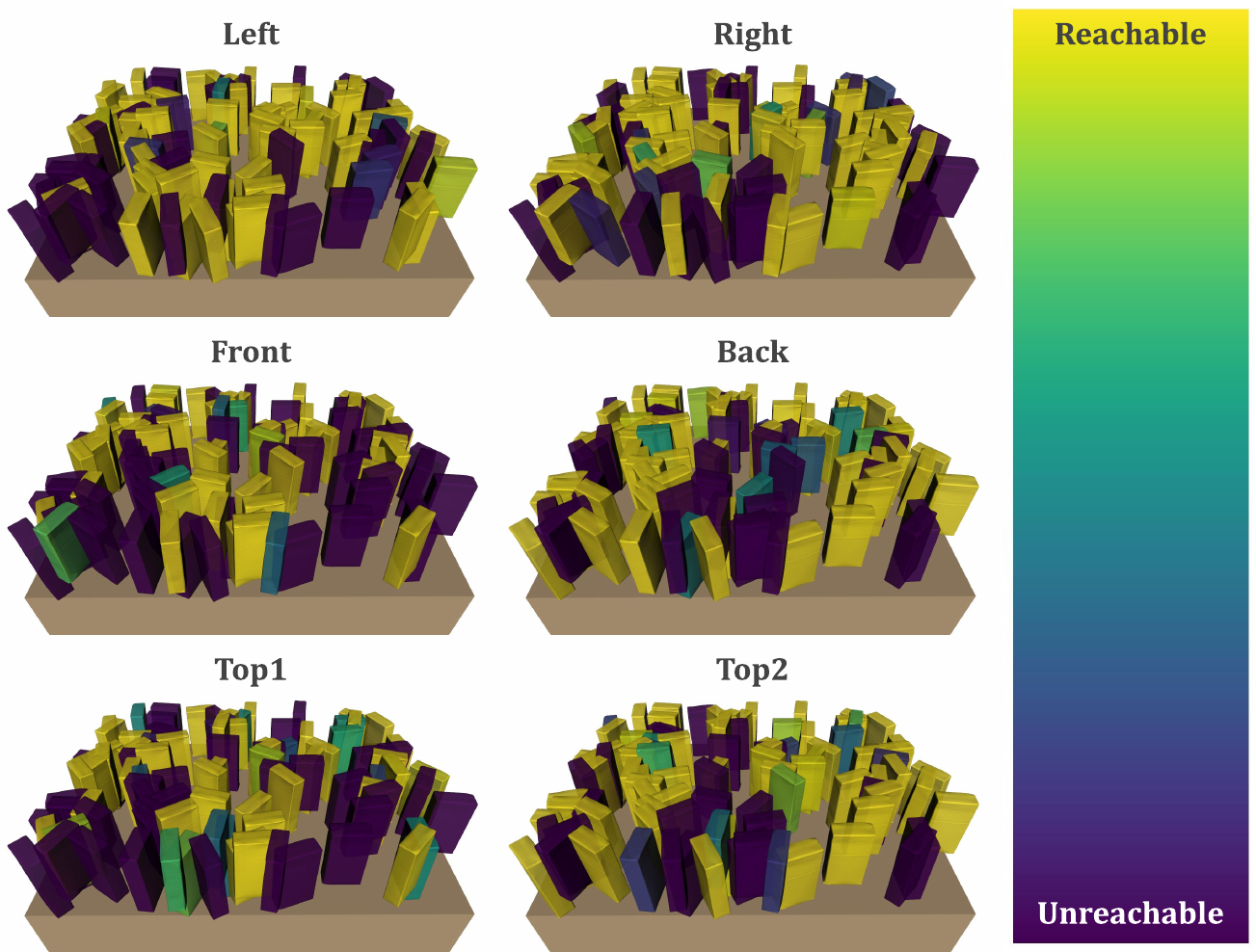}
  \caption{Comparison of the reachability of the different distribution
  components of the pose mixture model used in our arm-reaching experiments in
  Sec.~\ref{sec:arm_reaching}. Each sub-figure visualizes the object mesh in the
  same 100 random poses on the table where the mesh color indicates whether the
  component was reachable (yellow/green) or unreachable (blue/purple) for that
  object pose. Reachability is computed as described in
  Sec.~\ref{sec:arm_reaching}.}
  \label{fig:obj_reachability}
\end{figure*}


We provide some additional visualizations (Fig.~\ref{fig:obj_reachability}) for
our arm-reaching experiments from Sec.~\ref{sec:arm_reaching} to offer more
intuition for the reachability of the different components in the pose mixture
model we defined. As a reminder, reachability of a component is determined by
computing inverse kinematics for 100 samples and counting how many samples were
reachable. A component is deemed reachable if at least 50 of the samples are
reachable.

We found that some components are reachable in certain object poses and not
others. Additionally, for every object pose, there was typically at least one
component that was unreachable. Of the 100 object poses, there were only 5
instances where all components were reachable. 88 of the 100 poses had at least
one component with zero reachable samples.

We visualize the component reachability in
Fig.~\ref{fig:obj_reachability}. Every sub-figure shows the object mesh in the
same 100 uniformly random poses we used in our experiments in
Sec.~\ref{sec:arm_reaching}, where the color of the mesh indicates the
reachability of the associated component. Each sub-figure visualizes
reachablility for a different component. The color gradient shown in the right
side of the figure determines the interpretation of the mesh colors, where
yellow means the component is highly-reachable, purple means the component is
highly-unreachable, and green and blue are on the spectrum in between.

Note that there are far more instances where the components are nearly
completely reachable or completely unreachable in comparison to instances that
fall in the middle of that spectrum. This aligns well with our intuition -- if
one pose is reachable, it's likely that small perturbations of that pose will
also be reachable. However, there are instances where a pose is perhaps
reachable but close to the edge of the reachable workspace, or the robot may be
near a joint limit. In these instances, small perturbations to a reachable pose
may lead to unreachable samples.


\section{Solver Details}
\label{app:solver_details}

We utilized a variety of solvers in our experiments in
Sec.~\ref{sec:applications} including gradient-based and sampling-based
solvers. We will here describe in more details the particular solvers and
hyperparameters we used. We emphasize that we did not expend undue effort
optimizing hyperparameters and it is likely different solvers and
hyperparameters will have better performance on our problems.

\subsection{Dubins Planar Navigation Environment}
\label{app:dubins_solver}

We used a gradient-based solver for the planar navigation environment in
Sec.~\ref{sec:dubins}. We formulated our problem as a constrained optimization
problem in Drake~\cite{drake} and used the SNOPT~\cite{gill2005snopt} solver. We
used the following hyperparameters and parameter settings:

\begin{center}
  \begin{tabular}{||c|c|l||}
    \hline
    \textbf{Parameter} & \textbf{Value} & \textbf{Description} \\
    \hline
    \(I\) & 200 & Maximum number of iterations to run solver \\
    \hline
    \(T\) & 45 & Planning horizon \\
    \hline
    \(\mathrm{d}t\) & 0.3 & Timestep \\
    \hline
    \(\bm{\omega}_t\) & 0.002 & Variance for dynamics noise (isotropic Gaussian)\\
    \hline
    \(\bSi_0\) & 0.02 & Initial state covariance (isotropic Gaussian) \\
    \hline
    \(\beta\) & 2 & Parameter to unscented transform governing sigma point
                    dispersion \\
    \hline
  \end{tabular}
\end{center}

We used the default values provided by the Drake interface to SNOPT for all
SNOPT hyperparameters not mentioned in the table.

\subsection{Ball-Rolling Environment}
\label{app:amplifier_solver}

We used the \textit{cross-entropy method (CEM)}~\cite{kobilarov2012cross} for
our solver for the ball-rolling problem in Sec:~\ref{sec:amplifier}. CEM is a
sampling-based solver that generates solution samples from a Gaussian
distribution, evalutes the cost of each sample, and re-fits the distribution for
the next iteration using the samples with the lowest cost (a.k.a. the
\textit{elite set}). The CEM update rule has been derived from a planning as
inference framework assuming an optimality likelihood with thresholded
utility~\cite{lambert2020stein}.

We note that in principle the gradient solver from
Appendix~\ref{app:dubins_solver} could be used for this problem, but in practice
it was highly sensitive to the initial solution and would frequently get stuck
in local optima. In contrast, CEM found solutions very quickly (within
approximately 5-10 iterations) and was not as susceptible to local optima. We
used the following parameter settings:

\begin{center}
  \begin{tabular}{||c|c|l||}
    \hline
    \textbf{Parameter} & \textbf{Value} & \textbf{Description} \\
    \hline
    \(I\) & 50 & Maximum number of iterations to run CEM \\
    \hline
    \(N\) & 500 & Number of samples generated in each CEM iteration \\
    \hline
    \(K\) & 20 & Number of elite samples to re-fit CEM sampling distribution to
    \\
    \hline
    \(\sigma_0^2\) & 0.8 & Initial covariance for the CEM sampling distribution
                         (isotropic Gaussian) \\
    \hline
    \(H\) & 100 & Planning horizon \\
    \hline
    \(\mathrm{d}t\) & 0.3 & Timestep \\
    \hline
    \(\bm{\omega}_t\) & 0.0001 & Nominal variance for dynamics noise (isotropic Gaussian)\\
    \hline
    \(\bm{\omega}_t^\prime\) & 0.008 & Additional noise from amplifier
                                       (isotropic Gaussian) \\
    \hline
    \(\beta\) & 2 & Parameter to unscented transform governing sigma point
                    dispersion \\
    \hline
  \end{tabular}
\end{center}

Note the planning horizon for this environment is modeling the passive dynamics
of the rolling ball, since controls are only applied at the initial timestep to
initiate the motion of the ball.

\subsection{Target Intercept Environment}
\label{app:moving_target_solver}

We use \textit{model predictive path integral control
(MPPI)}~\cite{williams2017model} for our dynamic goal experiments in
Sec.~\ref{sec:moving_target}. MPPI is a sampling-based solver that iteratively
updates a Gaussian distribution of solutions by generating noisy perturbations
of the current mean solution at each iteration. A weighted average of the
samples is computed to determine the new mean solution, where higher cost
samples contribute less to the update. This procedure is carried out in a model
predictive control (MPC) scheme such that a plan for a small horizon is
generated at every iteration, the first action of that plan is executed, and the
agent re-plans at the next iteration. This procedure is widely used in robotics
due to the its computational simplicity, ease of
parallelization~\cite{bhardwaj2021storm}, and applicability to complex problems
even when the cost is non-differentiable. See~\cite{williams2017model} for a
more detailed and formal presentation of MPPI.

We used the following parameter settings:

\begin{center}
  \begin{tabular}{||c|c|l||}
    \hline
    \textbf{Parameter} & \textbf{Value} & \textbf{Description} \\
    \hline
    \(I\) & 70 & Number of iterations to run MPC \\
    \hline
    \(N\) & 100 & Number of samples generated in each MPPI iteration \\
    \hline
    \(\sigma_0^2\) & 0.02 & Initial variance for the MPPI sampling distribution
                           (isotropic Gaussian) \\
    \hline
    \(\sigma_N^2\) & 0.002 & Terminal variance for the MPPI sampling
                              distribution (isotropic Gaussian) \\
    \hline
    \(H_\mathrm{max}\) & 25 & Maximum lanning horizon \\
    \hline
    \(H_\mathrm{min}\) & 3 & Minimum lanning horizon \\
    \hline
  \end{tabular}
\end{center}

Note we employ a couple of strategies to get better results:
\begin{enumerate}
  \item We reduce the MPPI variance from \(\sigma_0^2\) to \(\sigma_N^2\) in
  even steps over the number of MPPI iterations \(I\). We found this strategy to
  produce better results since it can find more refined solutions as the number
  of MPPI iterations increases.
  \item We dynamically adapt the planning horizon to start at a maximum of
  \(H_\mathrm{max}\) and reduce to a minimum of \(H_\mathrm{min}\) based on how
  far the agent is from the target. In order to adapt this value based only on
  information available to the agent, we use the KL divergence between the
  agent's state distribution and the projected belief distribution as a measure
  of proximity so that as the agent gets closer to the target, it uses a shorter
  planning horizon.
\end{enumerate}

\subsection{Arm-Reaching Environment}
\label{app:arm_reaching_solver}

We use MPPI as our solver for the arm reaching environment from our experiments
in Sec.~\ref{sec:arm_reaching}. We note again that in principle the
gradient-based solver could in principle be used for this environment. However,
the solver required a good initial solution in order to find a valid plan, and
the initial solutions ended up over-biasing the solver to particular components
to reach to. We found the sampling scheme of MPPI to produce the least biased
results. Note that we are not using MPPI in an MPC setting as we did in
Appendix~\ref{app:moving_target_solver}, but instead just using the MPPI
sampling and update scheme to efficiently find solutions for the full planning
horizon. We used the following settings:

\begin{center}
  \begin{tabular}{||c|c|l||}
    \hline
    \textbf{Parameter} & \textbf{Value} & \textbf{Description} \\
    \hline
    \(I\) & 30 & Maximum number of iterations to run MPPI \\
    \hline
    \(N\) & 500 & Number of samples generated in each MPPI iteration \\
    \hline
    \(\sigma_0^2\) & 0.002 & Initial variance for the MPPI sampling distribution
                           (isotropic Gaussian) \\
    \hline
    \(\sigma_N^2\) & 0.0001 & Terminal variance for the MPPI sampling
                              distribution (isotropic Gaussian) \\
    \hline
    \(H\) & 10 & Planning horizon \\
    \hline
  \end{tabular}
\end{center}


\section{Skill Planning -- Additional Details}
\label{app:skill_planning}

In this appendix, we provide additional details on our experiments from
Sec.~\ref{sec:skill_planning} regarding applying our probabilistic planning
framework to the domain of skill planning.

\newcommand\figgoalgenW{0.329}

\begin{figure}[h!]
  \centering
  \begin{subfigure}{\figgoalgenW\textwidth}
    \centering
    \includegraphics[width=\textwidth]{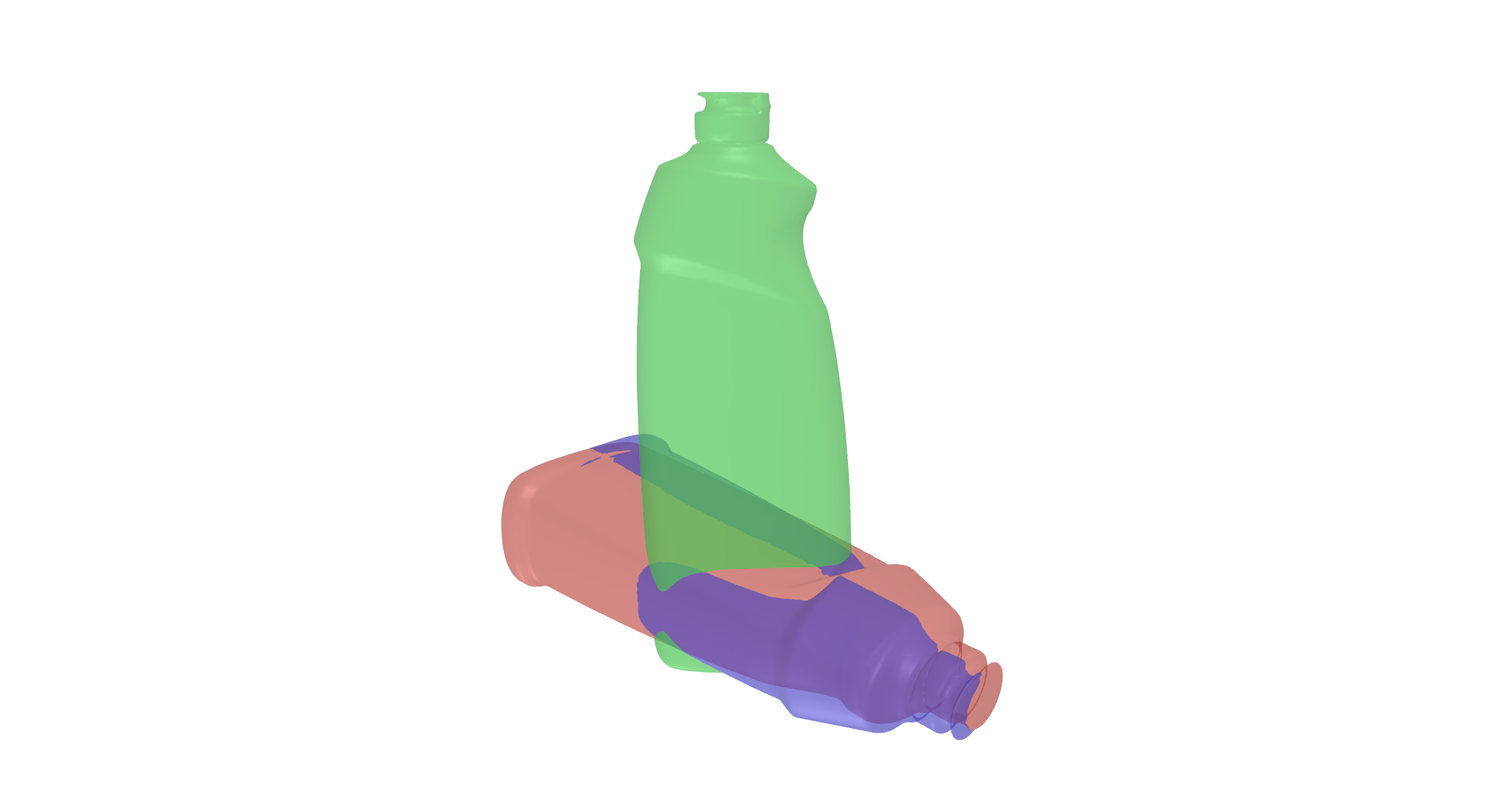}
    \caption{Stable poses}
    \label{fig:goal_gen_stable}
  \end{subfigure}
  \begin{subfigure}{\figgoalgenW\textwidth}
    \centering
    \includegraphics[width=\textwidth]{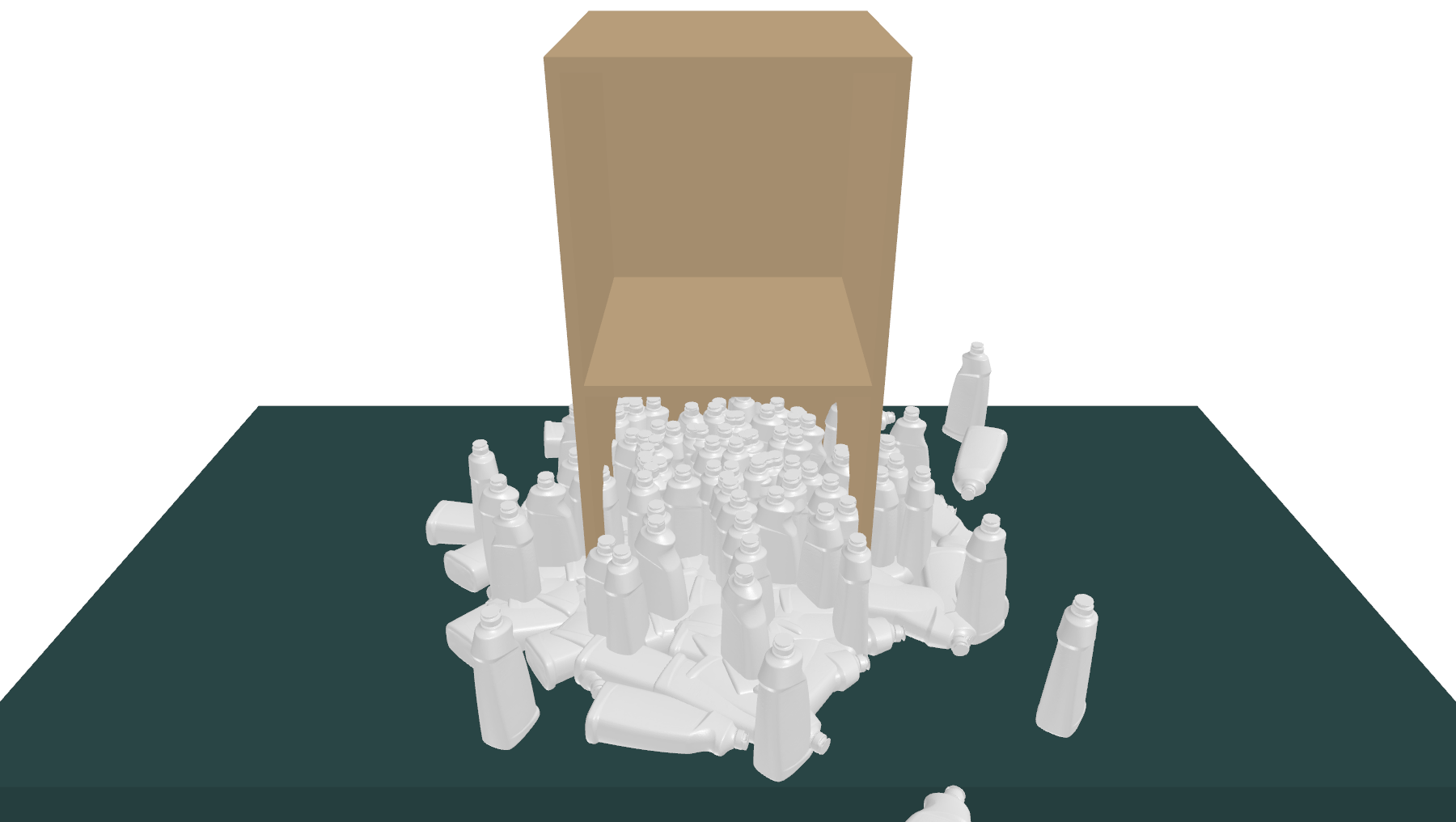}
    \caption{Initial samples}
    \label{fig:goal_gen_all_samples}
  \end{subfigure}
  \begin{subfigure}{\figgoalgenW\textwidth}
    \centering
    \includegraphics[width=\textwidth]{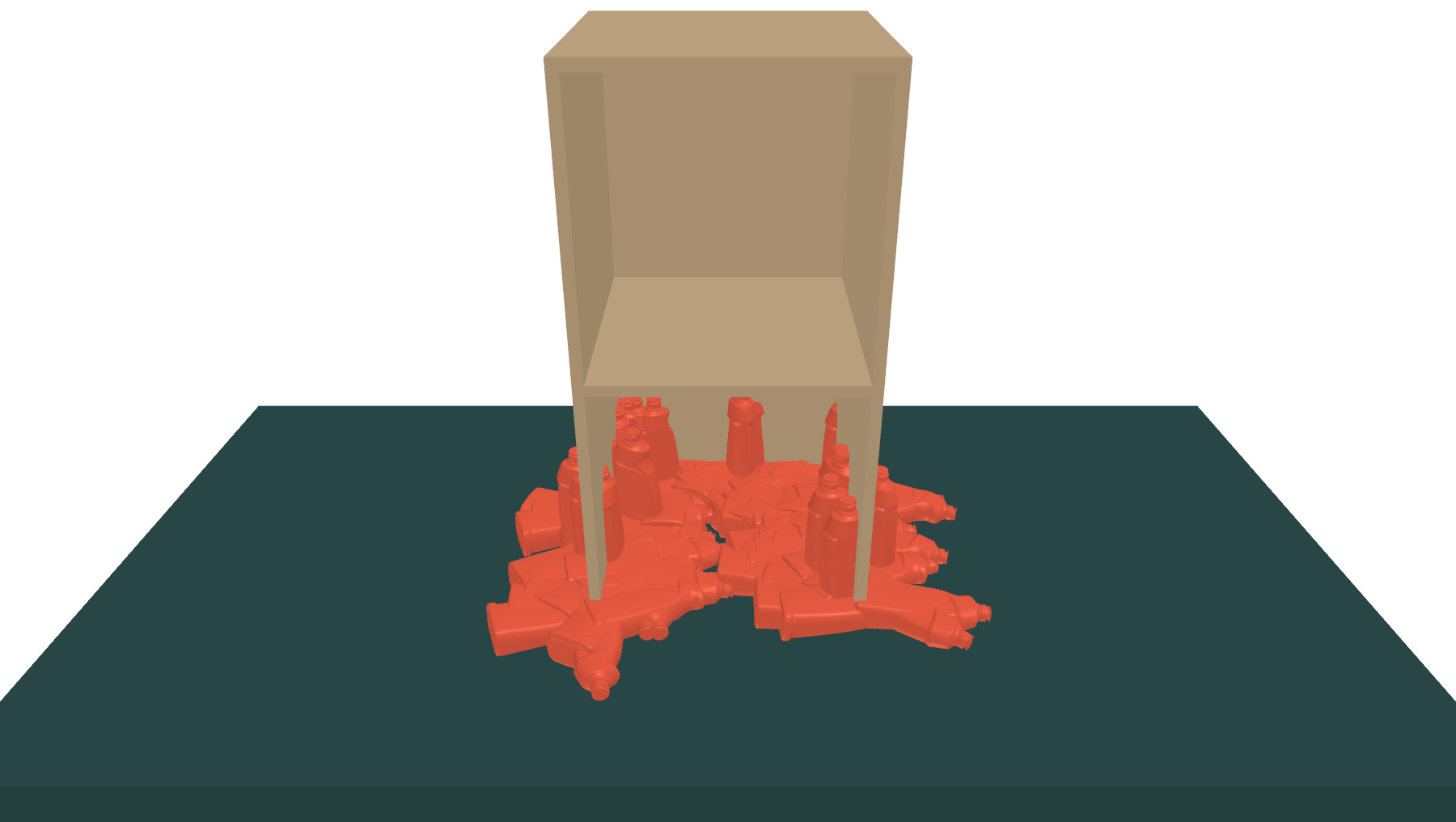}
    \caption{Samples in collision}
    \label{fig:goal_gen_in_collision}
  \end{subfigure}
  \begin{subfigure}{\figgoalgenW\textwidth}
    \centering
    \includegraphics[width=\textwidth]{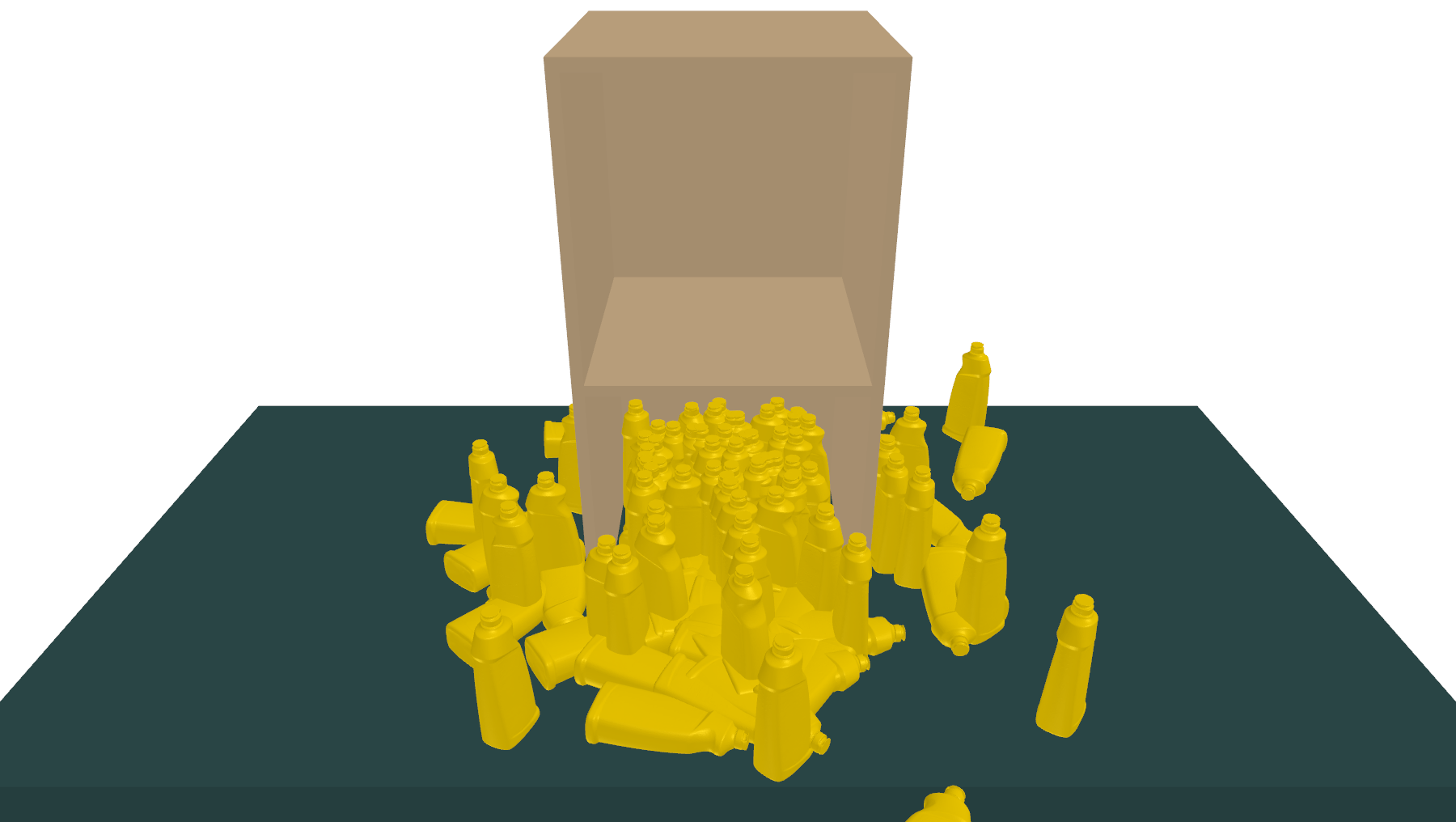}
    \caption{Samples free of collision}
    \label{fig:goal_gen_no_collision}
  \end{subfigure}
  \begin{subfigure}{\figgoalgenW\textwidth}
    \centering
    \includegraphics[width=\textwidth]{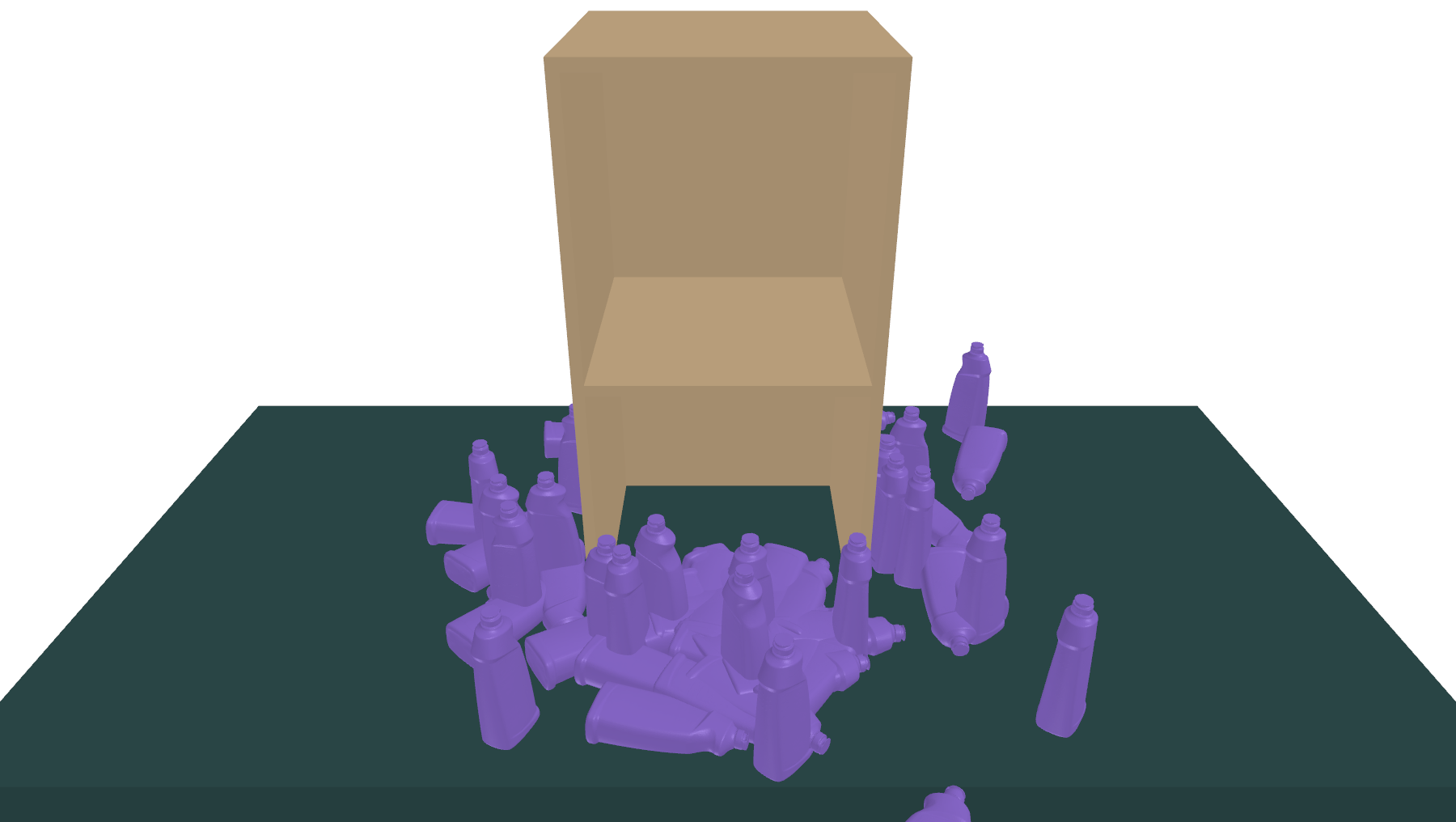}
    \caption{Samples not in shelf}
    \label{fig:goal_gen_no_contained}
  \end{subfigure}
  \begin{subfigure}{\figgoalgenW\textwidth}
    \centering
    \includegraphics[width=\textwidth]{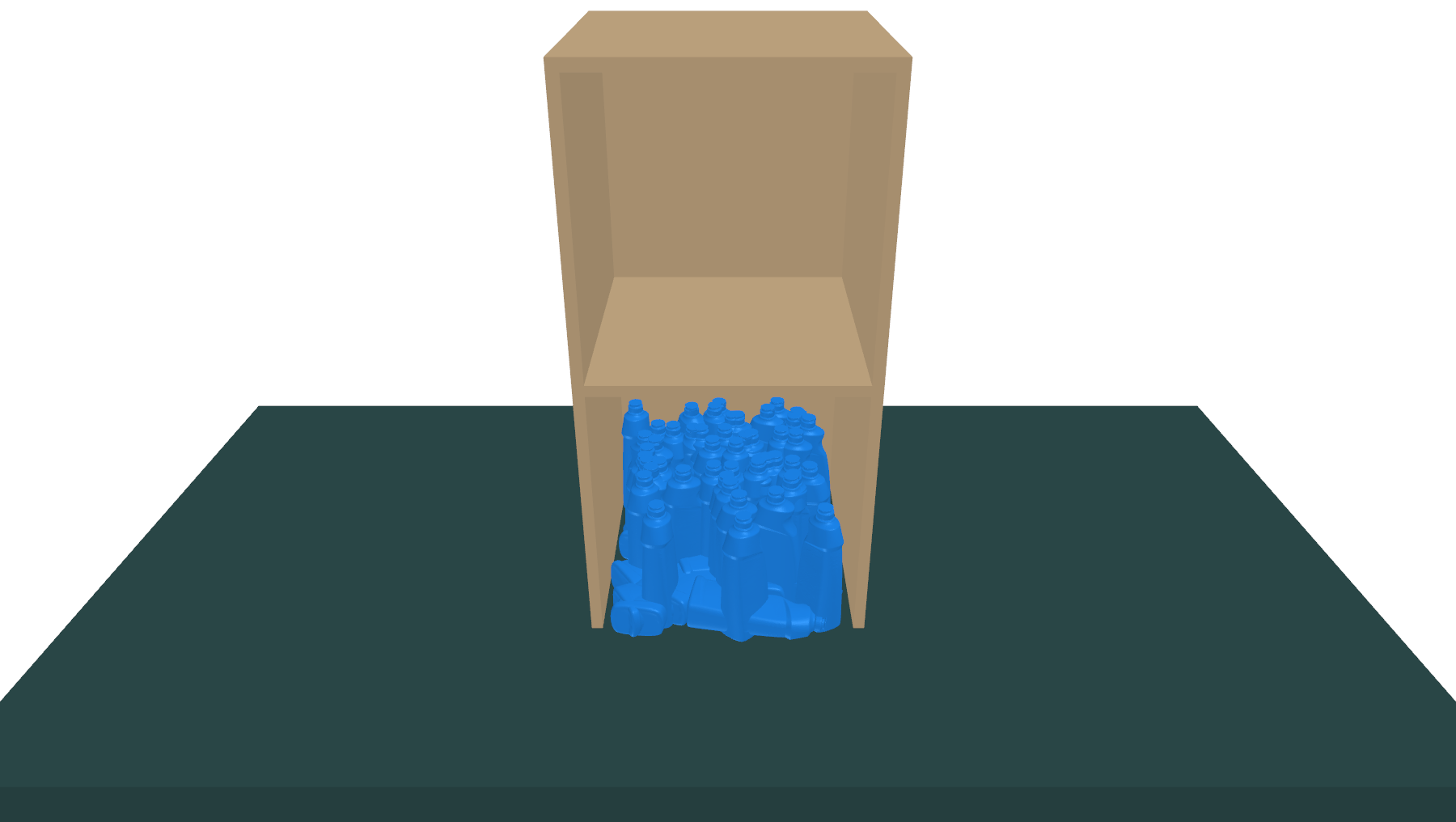}
    \caption{Samples contained in shelf}
    \label{fig:goal_gen_contained}
  \end{subfigure}
  \begin{subfigure}{\figgoalgenW\textwidth}
    \centering
    \includegraphics[width=\textwidth]{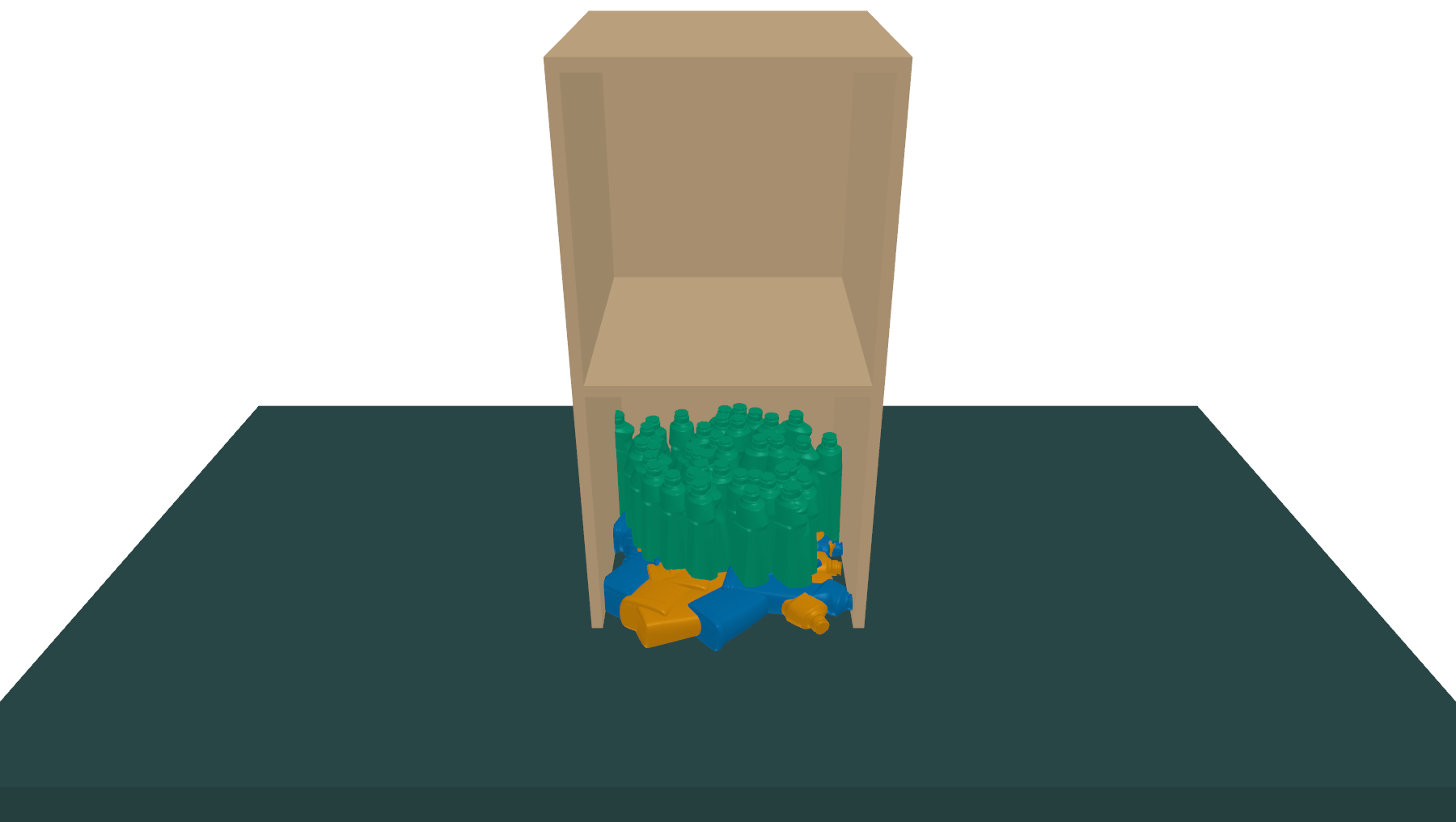}
    \caption{Goal GMM}
    \label{fig:goal_gen_gmm_all}
  \end{subfigure}
  \begin{subfigure}{\figgoalgenW\textwidth}
    \centering
    \includegraphics[width=\textwidth]{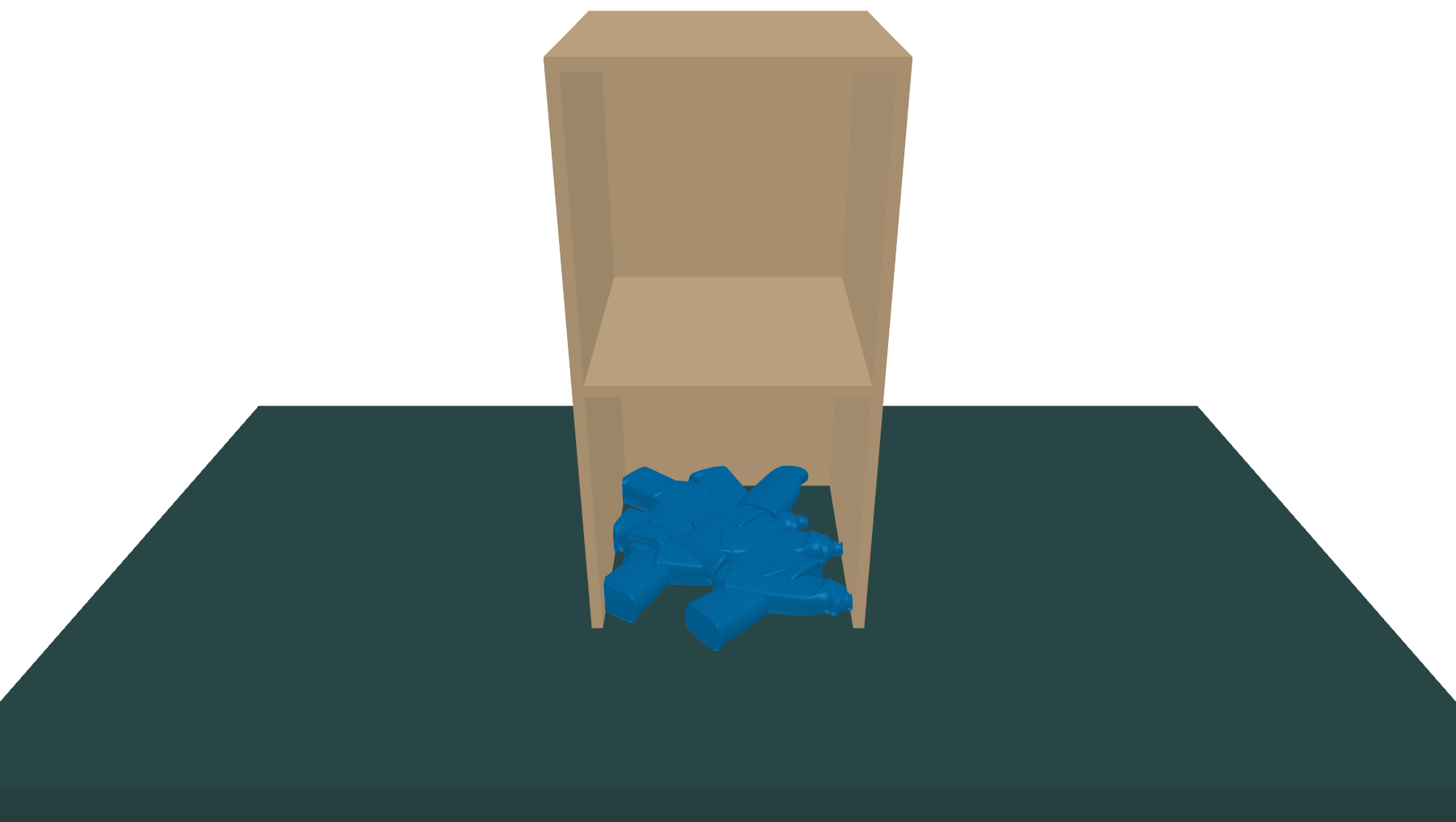}
    \caption{First component}
    \label{fig:goal_gen_gmm_c1}
  \end{subfigure}
  \begin{subfigure}{\figgoalgenW\textwidth}
    \centering
    \includegraphics[width=\textwidth]{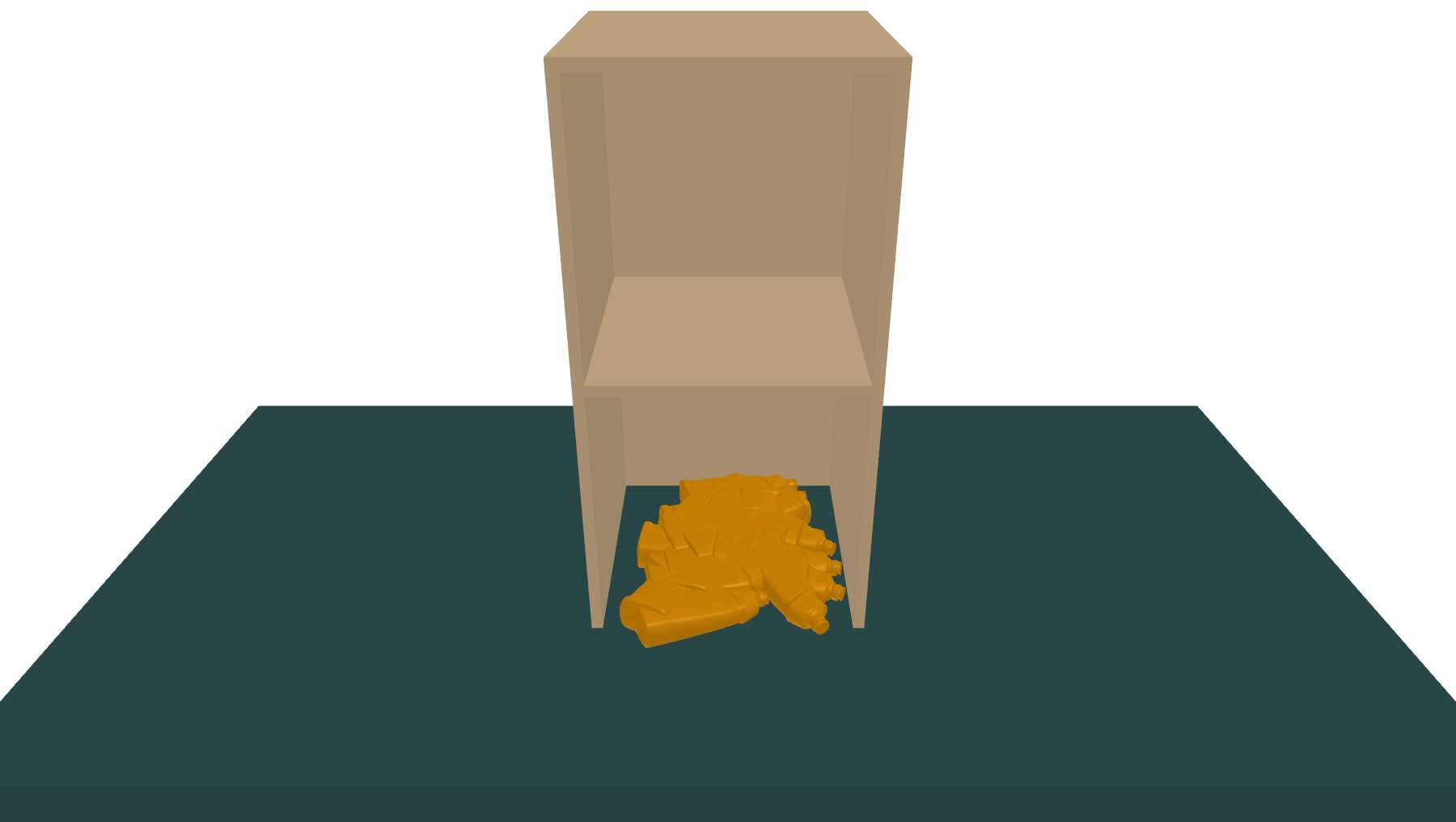}
    \caption{Second component}
    \label{fig:goal_gen_gmm_c2}
  \end{subfigure}
  \begin{subfigure}{\figgoalgenW\textwidth}
    \centering
    \includegraphics[width=\textwidth]{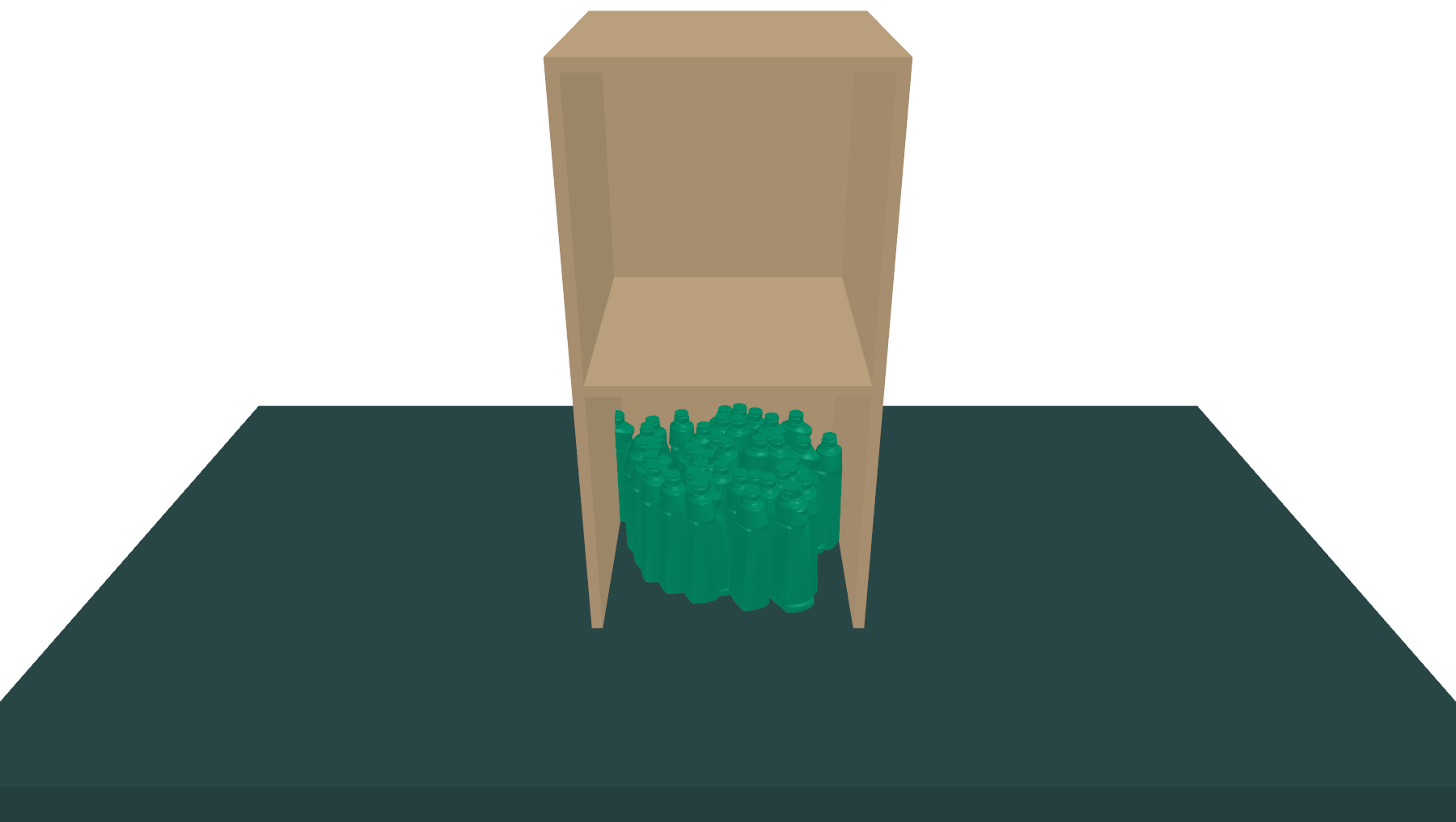}
    \caption{Third component}
    \label{fig:goal_gen_gmm_c3}
  \end{subfigure}
  \begin{subfigure}{\figgoalgenW\textwidth}
    \centering
    \includegraphics[width=\textwidth]{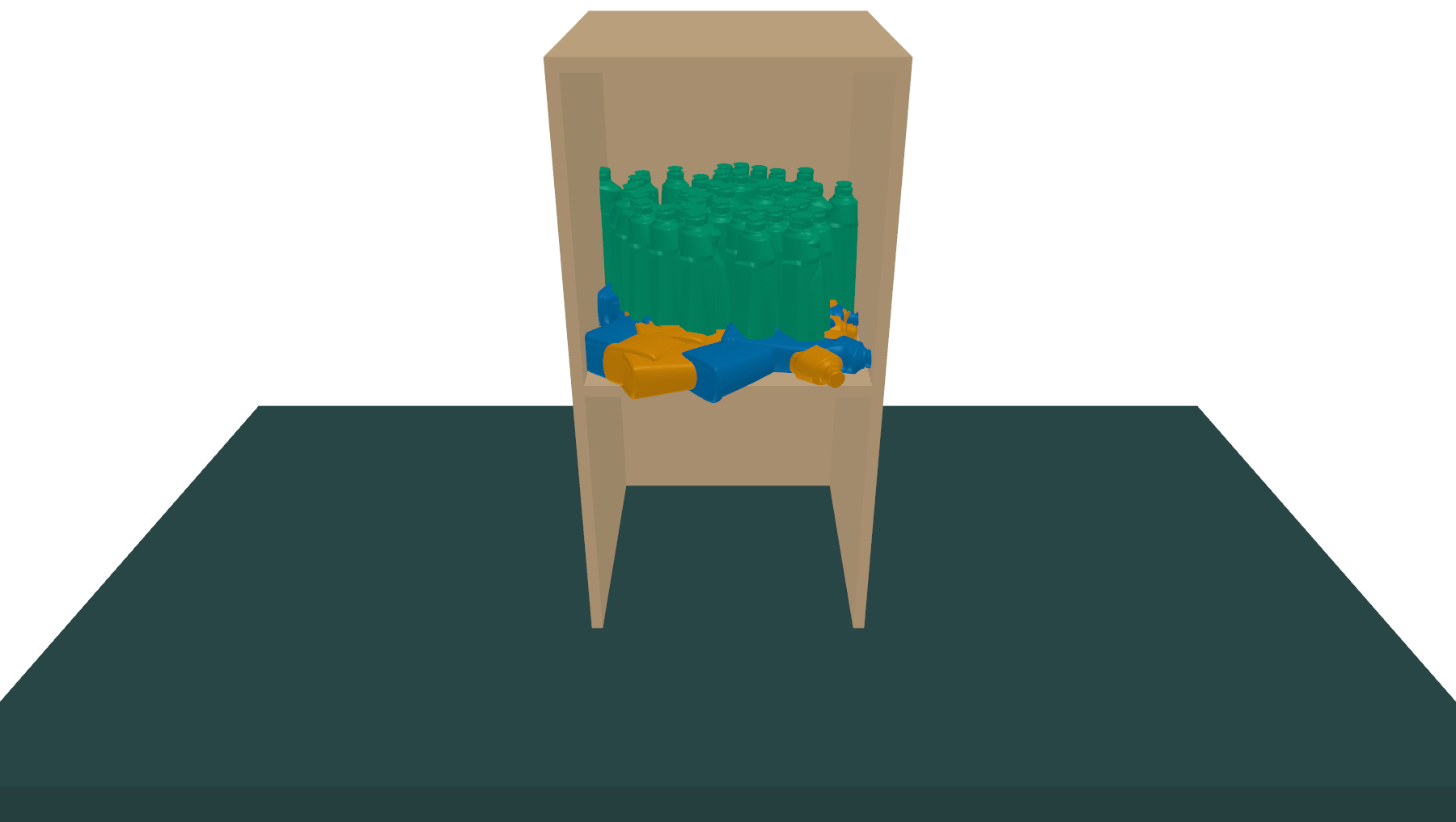}
    \caption{Goal GMM on middle shelf}
    \label{fig:goal_gen_mid}
  \end{subfigure}
  \begin{subfigure}{\figgoalgenW\textwidth}
    \centering
    \includegraphics[width=\textwidth]{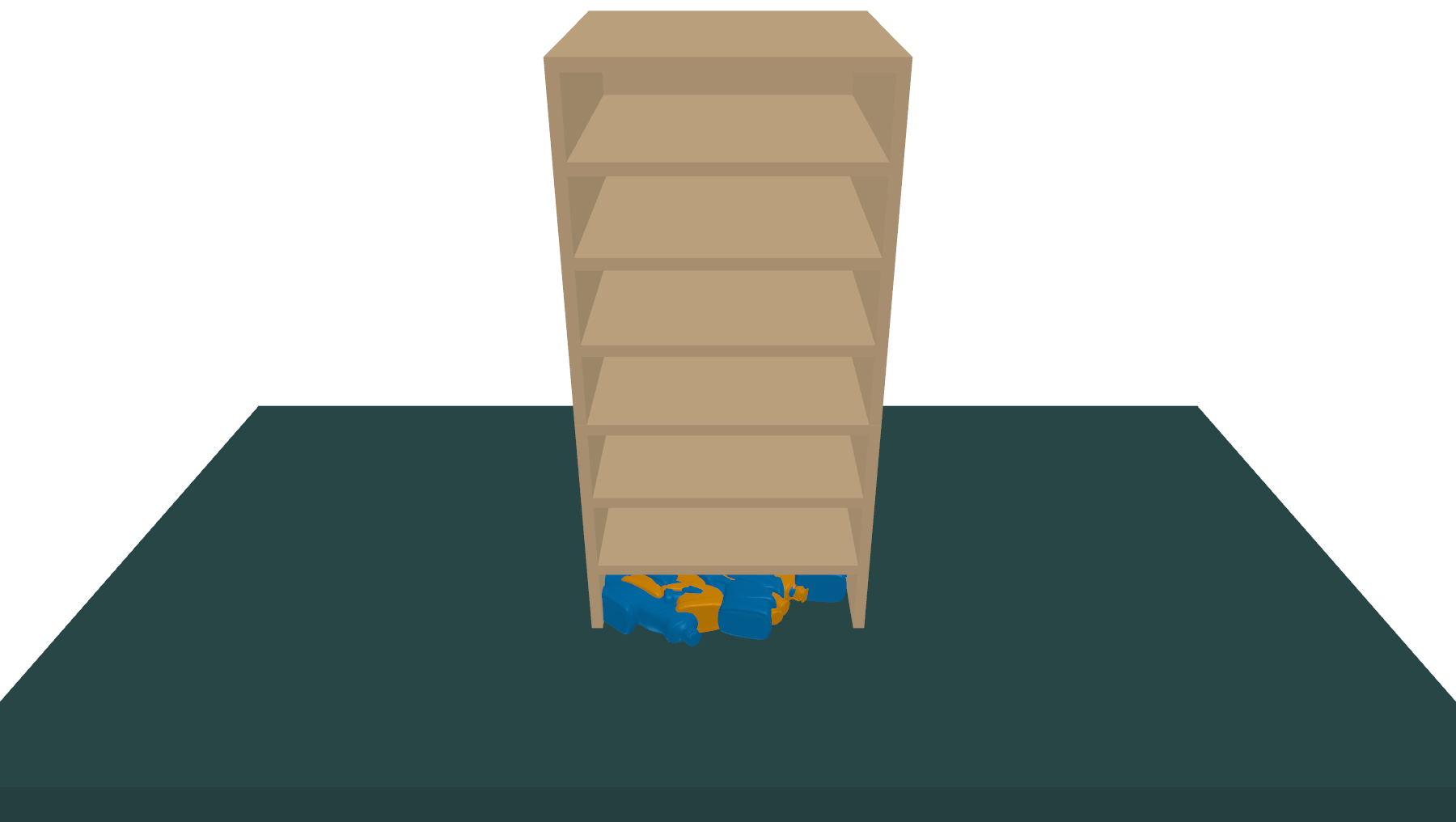}
    \caption{Goal GMM new bookcase}
    \label{fig:goal_gen_low}
  \end{subfigure}
  \caption[Example goal distribution generation process for the goal of
  ``cleaner bottle on shelf'' of a bookcase.]{Example goal distribution
  generation process for the goal of ``cleaner bottle on shelf'' of a
  bookcase. \textbf{(a)} Stable poses of the object on a support
  surface. \textbf{(b)} Pose samples generated from Gaussian-distributed planar
  perturbations of the stable poses. \textbf{(c)} Pose samples that place the
  object in collision with the shelf. \textbf{(d)} Samples free of collision
  with the environment. \textbf{(e)} Samples not contained within the bounding
  box of the shelf. \textbf{(f)} Samples contained within the
  shelf. \textbf{(g)} The resulting goal GMM fit to the collection of valid
  samples. \textbf{(h)} The first GMM component, with the bottle laying on its
  front side. \textbf{(i)} The second GMM component, with the bottle laying on
  its back side. \textbf{(j)} The third GMM component, with the bottle standing
  upright. \textbf{(k)} The goal GMM generated on the middle shelf of the
  bookcase. \textbf{(l)} Goal GMM generated for a different bookcase with lower
  shelves, which disallows samples with the bottle upright.}
  \label{fig:goal_gen}
\end{figure}


\subsection{Generating Goal Distributions}
\label{app:sp_goal_gen}

While goal distributions are a more expressive goal representation than single
points, it may be more of a burden to acquire a meaningful goal distribution in
certain contexts. In this section, we describe how we generated the Gaussian
mixture model goal distribution we used in Sec.~\ref{sec:skill_planning} to
capture the semantic goal of ``object on a shelf of bookcase'', shown in
Fig.~\ref{fig:goal_gen}. We assume object and environment meshes are given.

We perform the following steps to generate a goal GMM:
\begin{enumerate}
  \item \textbf{Compute stable object poses:} We compute stable poses (also
  known as \textit{support surfaces}) for the object to be placed on the
  shelf. We utilize Trimesh~\cite{trimesh} for this purpose, which computes
  stable poses by evaluating the probability of landing in random poses when
  dropped onto a table. The output is a small number of stable poses the object
  can be on a planar surface. In the case of the cleaner bottle we utilize in
  our experiments, it computed three stable poses (shown in
  Fig.~\ref{fig:goal_gen_stable}): one upright, one on the front flat side, and
  one on the rear flat side. We translate these stable poses to the center of
  the shelf surface for which we're generating a goal distribution for.
  \item \textbf{Generate random pertubations of stable poses:} We generate
  random planar pose perturbations from a zero-mean Gaussian distribution with a
  small, manually set variance. We compose these perturbations with each stable
  pose to produce a set of pose samples resting on the shelf surface
  (Fig.~\ref{fig:goal_gen_all_samples}).
  \item \textbf{Filter out samples in collision:} Many of the samples may
  collide with the walls of the shelf if the variance is set high enough. We
  reject these samples using the Trimesh~\cite{trimesh} interface to the
  Flexible Collision Library (FCL)~\cite{pan2012fcl}. Note we take the convex
  hull of the object mesh to speed up collision checking. Samples in collision
  are shown in Fig.~\ref{fig:goal_gen_in_collision}.
  \item \textbf{Filter out samples not contained by the shelf:} Some samples may
  have been generated outside the support surface of the shelf, and even out of
  the boundary of the shelf entirely. We filter these samples by checking
  whether the object mesh is contained within the bounding box of the
  shelf. Fig.~\ref{fig:goal_gen_no_contained} shows the samples not contained
  within the bookcase, and Fig.~\ref{fig:goal_gen_contained} shows contained
  samples. Note this step might be used in lieu of (3) since containment in the
  bounding box will likely filter the samples in collision. However, we include
  both steps for generality since the collision checking may still be needed if,
  e.g., there are other objects on the shelf.
  \item \textbf{Fit a parametric distribution to the samples:} We fit a Gaussian
  mixture model to the samples. We can conveniently set the number of GMM
  components to be the number of stable poses computed in (1). In this case we
  get the resulting 3-component GMM visualized in
  Fig.~\ref{fig:goal_gen_gmm_all} where the individual components are shown in
  Fig.~\ref{fig:goal_gen_gmm_c1}, Fig.~\ref{fig:goal_gen_gmm_c2}, and
  Fig.~\ref{fig:goal_gen_gmm_c3}.
\end{enumerate}

This procedure can be performed for any of the shelves in the bookcase once the
height of the shelf support surface is known. We generate the goal GMM on the
middle shelf of the bookcase in Fig.~\ref{fig:goal_gen_mid}. Due to the
collision-aware sampling, we can also generate the goal distribution when
shelves are more closely spaced apart, as shown in
Fig.~\ref{fig:goal_gen_low}. The shorter shelves require the bottle to lay on
one of its flat sides and prevent the bottle from sitting upright. Our
generation process automatically handles this case when in-collision samples are
filtered out in step (3).

This process is also not limited to this environment, and each phase may be
relaxed to achieve the desired goal distribution. We note that for more complex
scenarios, collision and containment checks may not be sufficient. For example,
if we had the goal of ``object in bin'', the object in that case may not be
limited to resting only on its support surfaces. In this case, we might want to
leverage a physics simulator to spawn the object randomly above the bin and let
the object fall into the bin. Using physics-informed constraints and heuristics
to filter samples enables acquiring goal distributions that are meaningful for
manipulation without the burden of having to manually specify associated
uncertainty parameters for a particular goal.

\subsection{Data Collection}

We use the NVIDIA Isaac Gym simulator for all data collection for our skills
from Sec.~\ref{sec:skill_planning}. We developed engineered, motion-planned
behaviors\footnote{Code for our engineered behaviors:
\texttt{\scriptsize\url{https://bitbucket.org/robot-learning/ll4ma_isaacgym}}}
using MoveIt~\cite{coleman2014reducing} to support autonomous collection. The
Isaac Gym simulator enables several environments (we used 16) to run in parallel
on a single machine. This together with our engineered behaviors enables large
scale data collection. We were able to collect 10,000 instances of a push skill
in approximately 12 hours on a Linux machine with Ubuntu 20.04, an Intel Core i7
processor, and an NVIDIA RTX2070 GPU.

\subsection{Point-Based Versus Unimodal Versus Multimodal Predictions}
\label{app:sp_model_compare}

\begin{figure}[t]
  \centering
  \includegraphics[width=\textwidth]{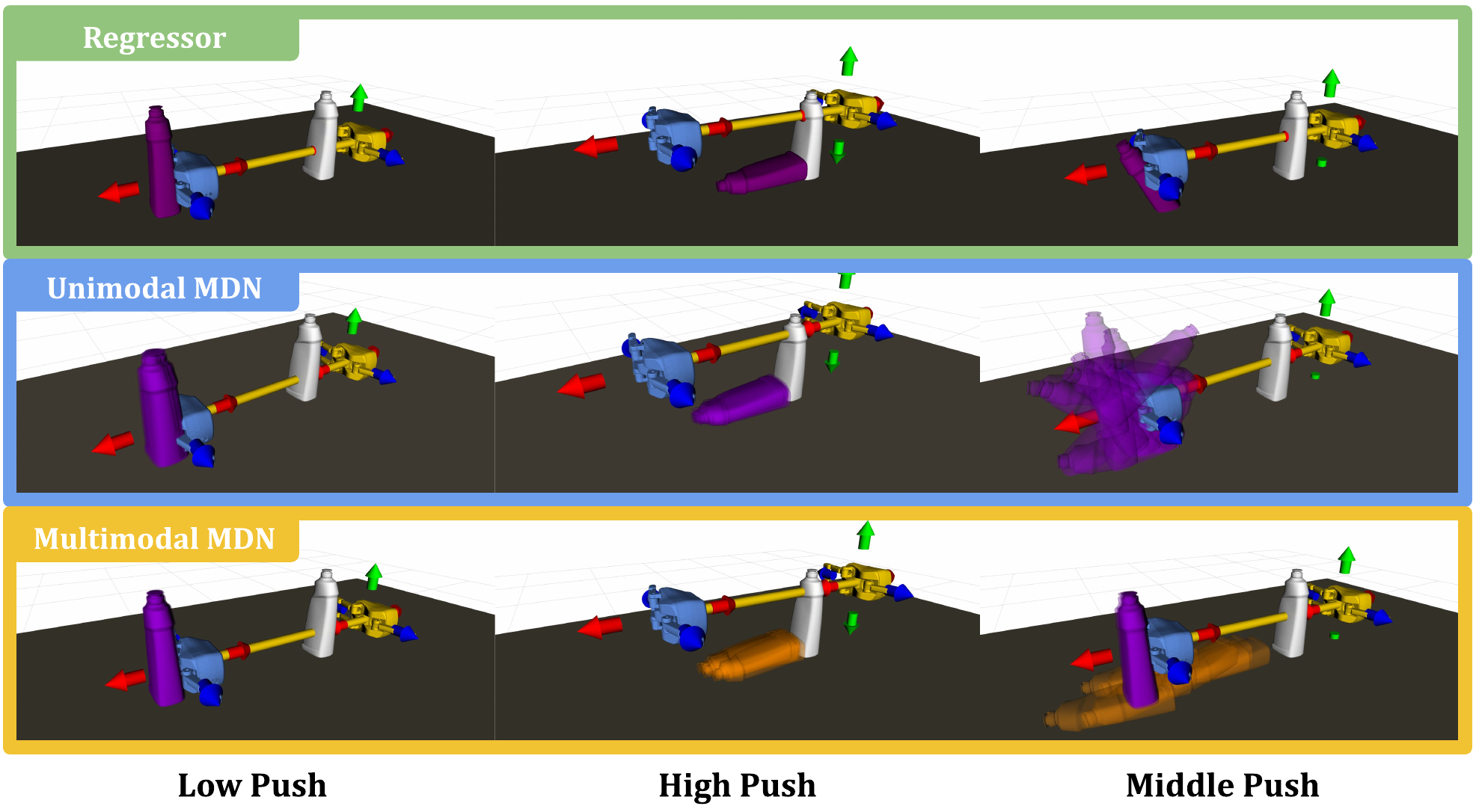}
  \caption[Qualitative comparisons of point-based (regressor), unimodal
  (1-component MDN), and multimodal (2-component MDN) outcome predictions for
  the push skill pushing at low, high, and middle heights.]{Qualitative
  comparisons of point-based (regressor), unimodal (1-component MDN), and
  multimodal (2-component MDN) outcome predictions for the push skill pushing at
  low, high, and middle heights. For low-pushes \textbf{(left column)} and
  high-pushes \textbf{(middle column)} all three models give reasonable
  predictions of the object remaining upright and toppled,
  respectively. However, for middle-pushes \textbf{(right column)}, only the
  multimodal MDN \textbf{(bottom row)} gives plausible predictions with one mode
  for the object remaining upright and a second mode with the object toppled. In
  contrast, the point-based regressor \textbf{(top row)} predicts the object in
  a half-toppled state, and the unimodal predictor \textbf{(middle row)}
  predicts a distribution with samples dispersed in the air and in collision
  with the end-effector.}
  \label{fig:push_model_compare}
\end{figure}


We compare different choices for the skill effect model outputs. In
Sec.~\ref{sec:skill_planning} we indicated we desire multimodal predictions to
capture different manipulation modes the object might end up in. We provide some
qualitative results here to support this idea, shown in
Fig.~\ref{fig:push_model_compare}.

Our learned models\footnote{Code for our learned
models:\texttt{\scriptsize{\url{https://bitbucket.org/robot-learning/multisensory_learning}}}}
were developed in PyTorch. Our MDN model consisted of two hidden layers each of
size 256 with ReLU activations. Since we predict distributions over (delta)
object poses, standard MDNs~\cite{bishop1994mixture} are insufficient since the
inputs and predictions must satisfy the constraints of \(SE(3)\) rigid body
transformations. Common parameterizations of rotations like quaternions and
Euler angles introduce discontinuities that are difficult for neural networks to
learn~\cite{zhou2019continuity}. We address this problem by utilizing the
learned 6D representation described in~\cite{zhou2019continuity} that uses a
Gram-Schmidt process to ensure orthogonalization. This provides us with a
continuous representation for orientations to learn skill effect models over
object poses. The point-based models also have two hidden layers of size 256 with
ReLU activations, and differ only in the output -- instead of predicting the
parameters of a GMM they predict only a single point estimate

We train three different skill effect models for pushing the cleaner bottle from
the YCB dataset~\cite{calli2015ycb}. All three models receive the same input,
which is a single point-based estimate of the initial object pose, as well as
the parameters for the push skill, which consist of the initial and terminal
end-effector poses in the object frame. The models differ only in the training
loss and class of predicted output. The models each output the predicted change
in object pose from its initial pose, where they predict either a point estimate
or a distribution over poses. We then compare model predictions for different
push heights where the object should remain upright for low-pushes, topple for
high-pushes, and there should be some uncertainty about whether it topples or
remains upright for middle pushes.

We train a point-based regressor (top row of Fig.~\ref{fig:push_model_compare})
that predicts the terminal pose of the object upon executing the push
skill. This model is trained with a regression loss using the ground truth
terminal pose as the target, where the loss is computed as the sum of the
position and rotation error between the predicted and ground truth object pose
\begin{equation}
  \mathcal{L}_{\mathrm{reg}} = ||\bm{p} - \hat{\bm{p}}||^2_2 + \theta_R
\end{equation}
where position error is the L2-norm between the predicted position
\(\hat{\bm{p}}\) and ground truth position \(\bm{p}\), and \(\theta_R\) is the
L2-norm of the matrix-logarithm error~\cite{lynch2017modern} between the
predicted orientation \(\hat{\bm{R}}\) and ground truth orientation
\(\bm{R}\). We see that a low push (top row, left column) and a high push (top
row, middle column) produce reasonable predictions of the object upright and
toppled, respectively. However, for a middle push, the model tends to predict
the object is suspended halfway between upright and toppled (top row, right
column). We also train a unimodal predictor, which is a single-component MDN
(middle row, right column) trained with the MDN loss defined as
\begin{equation}
  \label{eq:mdn_loss}
  \mathcal{L}_{\mathrm{MDN}}(\bm{\xi}) = \frac{1}{|\mathcal{D}|}
  \sum_{(\bm{\omega}_t, \bm{x}_{t+1}, \bm{\theta}) \in \mathcal{D}} -\log
  p\left(\bm{x}_{t+1} \middle| \bm{\omega}_t, \bm{\theta}, \{\alpha_{t+1}^i, \bm{\mu}_{t+1}^i,
  \bm{\Sigma}_{t+1}^i\}_{i=1}^M\right)
\end{equation}
where \(\mathcal{D}\) is a training dataset of state transitions and skill
parameters. Similar to the regressor, predictions are reasonable for low and
high pushes having Gaussian predictions with low variance of the object upright
and toppled, respectively. However, the middle pushes where there is ambiguity
about it being toppled or upright result in a Gaussian prediction with samples
dispersed in the air and in collision with the end-effector and the table. In
contrast to the regressor and the unimodal MDN, a 2-component MDN (bottom row,
right column) captures both possibilities of the middle push, one component with
the object upright and another with it toppled.

While it may seem innocuous that the point-based and unimodal models produce bad
predictions on occasion, we find that even seemingly rare bad predictions are
easily exploited by the planner. Our multimodal predictions mitigate some of
these exploits. We discuss additional ways in which poor model predictions were
exploited by the planner, and how we remedied those scenarios in the next
section.

\subsection{Fixing Erroneous Model Predictions Exploited by Planner}
\label{app:sp_exploit}

\begin{figure}[t]
  \centering
  \begin{subfigure}{0.497\textwidth}
    \centering
    \includegraphics[width=\textwidth]{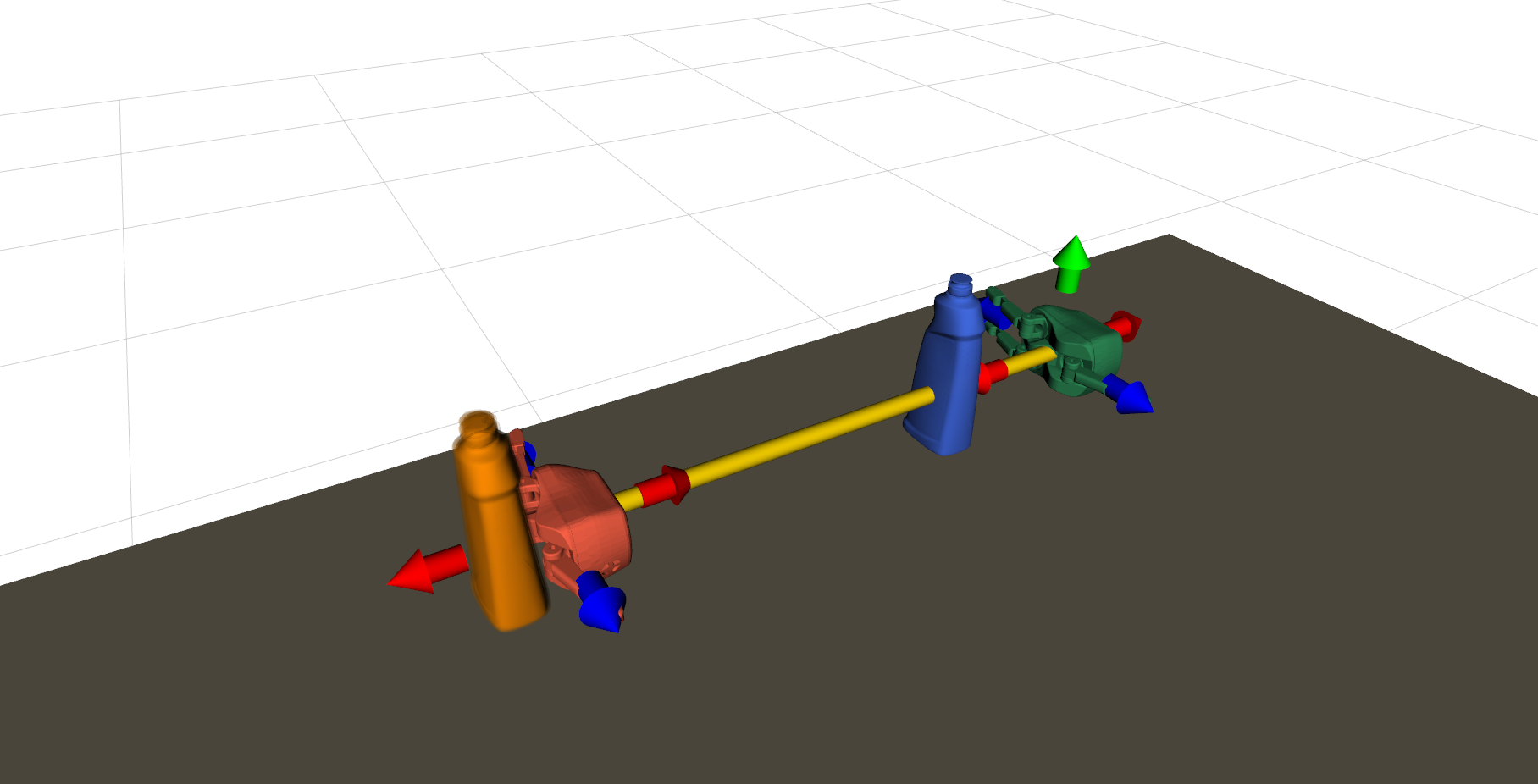}
    \caption{Correct prediction}
    \label{fig:push_pred_good}
  \end{subfigure}
  \begin{subfigure}{0.497\textwidth}
    \centering
    \includegraphics[width=\textwidth]{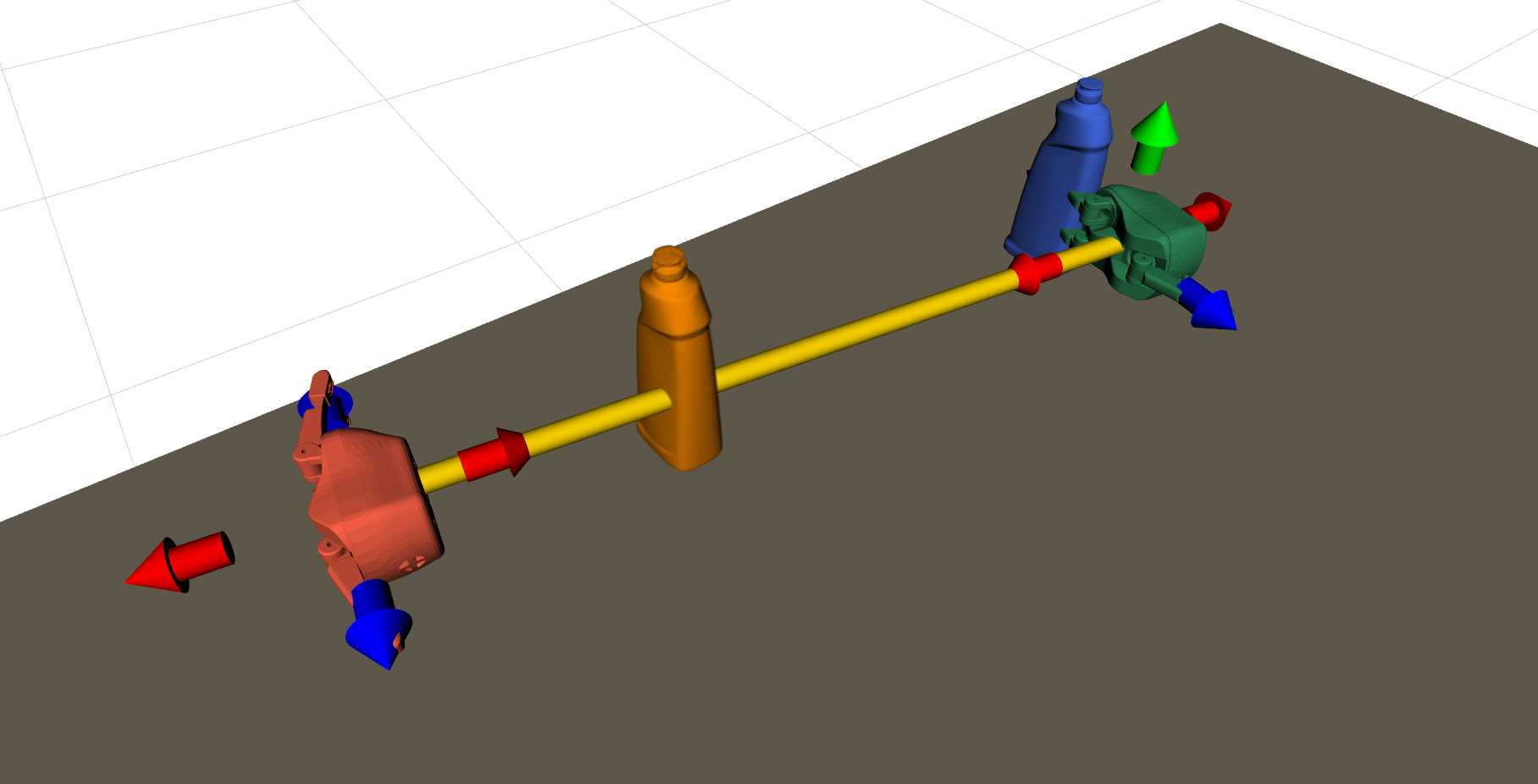}
    \caption{Prediction error from collision}
    \label{fig:push_pred_collision}
  \end{subfigure}

  \begin{subfigure}{0.497\textwidth}
    \centering
    \includegraphics[width=\textwidth]{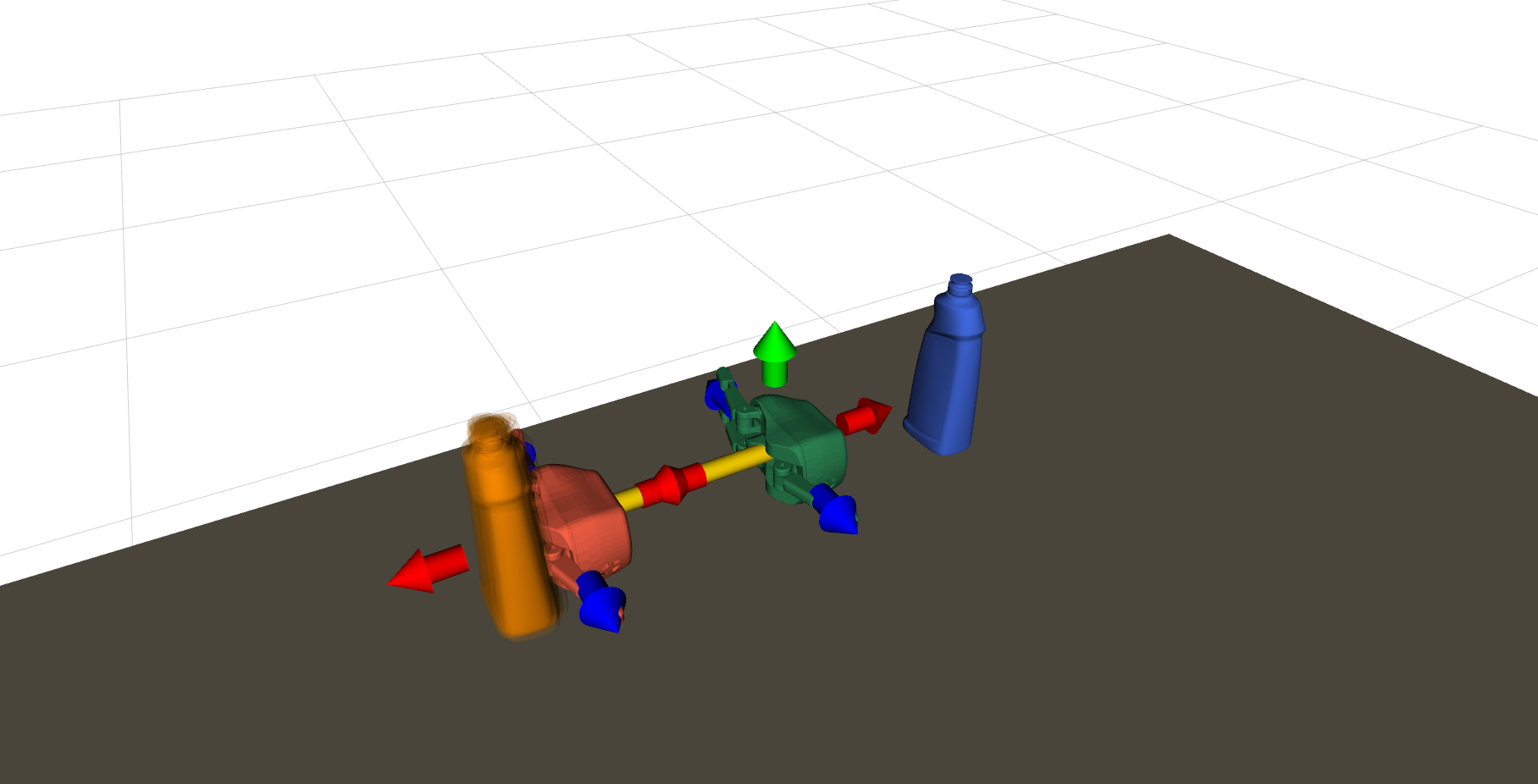}
    \caption{Prediction error no-op action}
    \label{fig:push_pred_noop_bad}
  \end{subfigure}
  \begin{subfigure}{0.497\textwidth}
    \centering
    \includegraphics[width=\textwidth]{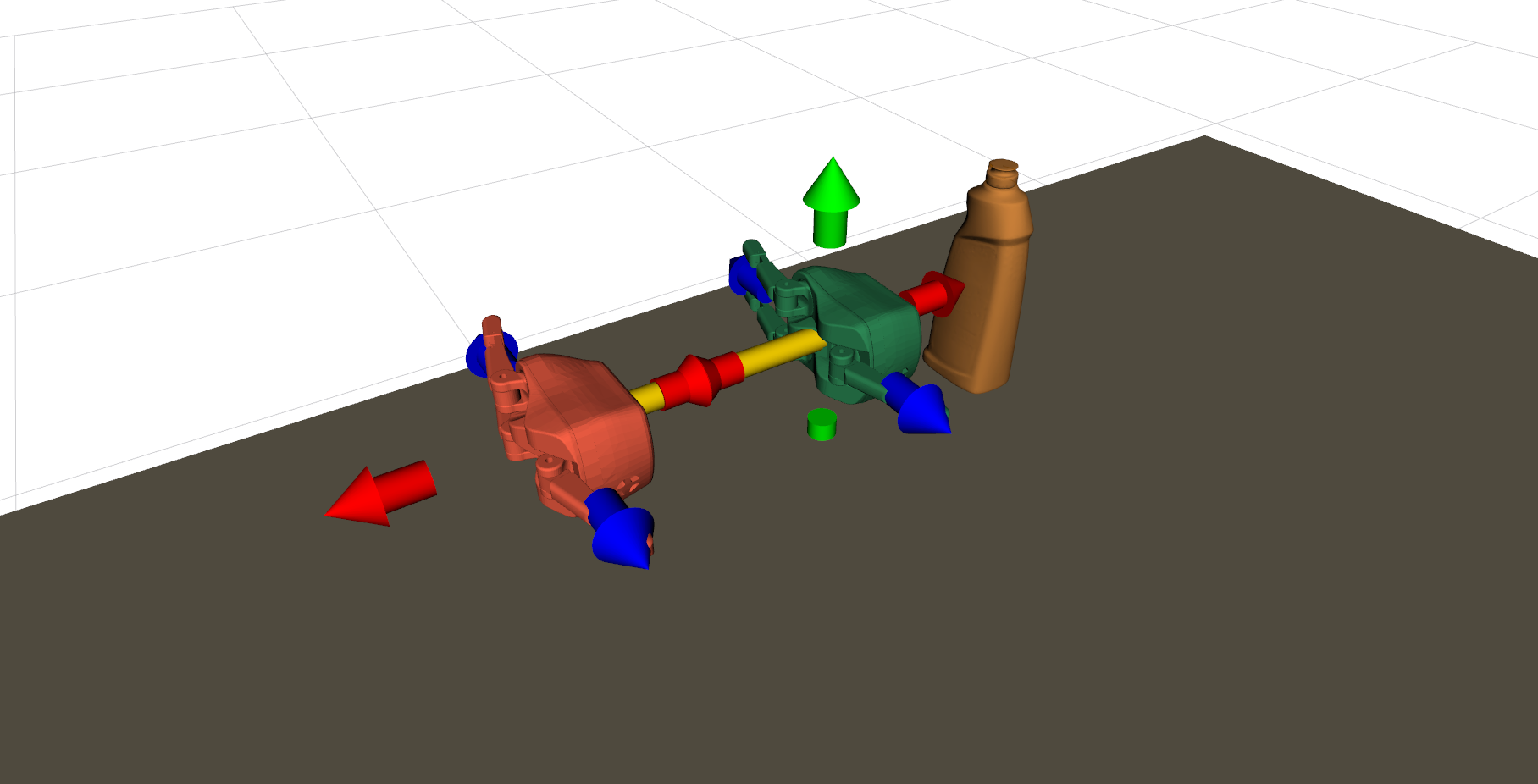}
    \caption{Correct no-op action prediction}
    \label{fig:push_pred_noop_good}
  \end{subfigure}
  \caption[Examples of the planner exploiting conditions outside the training
  data.]{Examples of the planner exploiting conditions outside the training
  data. Shown are initial object pose (blue mesh), predicted object pose
  distribution (orange meshes), start end-effector pose (green hand) and final
  end-effector pose (red mesh) for the push skill. \textbf{(a)} A correct
  outcome prediction for the push skill that we expect the planner to utilize.
  \textbf{(b)} The end-effector start pose is in a penetrating collision with
  the object, erroneously predicting it can relocate the object by executing a
  long push past the terminal object pose.  \textbf{(c)} The end-effector begins
  in front-of the object, erroneously predicting a push.  \textbf{(d)} A correct
  no-op prediction when the end-effector starts in front of the object after
  training data is augmented with no-op data.}
  \label{fig:push_exploits}
\end{figure}


Ideally we can learn models only from instances of the robot executing skills
successfully in order to capture correct outcome predictions for the
object. However, unless the planner is restricted to optimizing actions only
seen in the training data, it is wonderfully proficient at finding actions
outside the training data where model predictions are nonsensical, and
exploiting them. We provide examples we encountered in
Fig.~\ref{fig:push_exploits}. In these examples, the robot was tasked to simply
push the bottle from its initial pose (blue mesh) to a target pose 40cm
away. The orange meshes in Fig.~\ref{fig:push_pred_good} show a correct
prediction and the intended push for the planner to find.

However, because the training data contained no instances of the end-effector in
a penetrating collision with the object (a physical impossibility), the planner
exploits that condition as shown in Fig.~\ref{fig:push_pred_collision} and
believes it can achieve the target pose by penetrating the object and executing
a long push past the target pose. In order to remedy this, we incorporate a
collision cost as an auxiliary cost for planning that penalizes collisions
between the end-effector in its pre-push pose and the object for each push step
in the plan. We note that no amount of training data can remedy this condition
since even in simulation such an event is not possible, and it must instead be
accounted for in the planning cost.

A second exploit occurs for no-op actions, i.e., valid skill executions that do
not interact with the object and should then produce no change in the state of
the object. Fig.~\ref{fig:push_pred_noop_bad} shows a push in which the
end-effector starts in front of the object, and executes a push without ever
touching the object. Because the training data did not contain such an instance,
the model predicts it was an actual push and that the terminal state
distribution will be located near the terminal pose of the end-effector after
the push (orange meshes in Fig.~\ref{fig:push_pred_noop_bad}). We remedy this
condition by augmenting the training data with no-op actions. Instead of
collecting these instances in the simulator (which takes several hours), we take
the nominal training set and generate synthetic no-op instances. For each
initial object pose in the nominal set we generate random pushes and check for
collisions between the end-effector and the object using
Trimesh~\cite{trimesh}. If any collision is detected, that sample is rejected,
otherwise we set the initial and final object poses to be the initial object
pose for the no-op sample. We collect \(N=5\) no-op samples per nominal instance
and incorporate these into our training data. Fig.~\ref{fig:push_pred_noop_good}
shows the model prediction after the model is trained with the augmented
dataset. The model predicts the object distribution remains a tight Gaussian at
the initial object pose, i.e., the action leaves the object undisturbed, as
desired.


\end{document}